\documentclass[10pt,twocolumn,letterpaper]{article}

\usepackage[pagenumbers]{cvpr} 

\usepackage[table, dvipsnames]{xcolor}
\newcommand{\red}[1]{{\color{red}#1}}
\newcommand{\green}[1]{{\color{PineGreen}#1}}

\usepackage{algorithmic}
\usepackage{algorithm2e}
\usepackage{amsmath}
\usepackage{amssymb}
\usepackage{amsfonts}
\usepackage{xspace}
\usepackage{graphicx}
\usepackage{soul} 
\usepackage{physics} 
\usepackage{booktabs}

\usepackage{pifont}
\usepackage{makecell, cellspace}
\usepackage[accsupp]{axessibility} 

\DeclareMathOperator*{\argmin}{\arg\!\min}

\newcommand{\Paragraph}[1] {\leavevmode\newline \noindent \textbf{#1}}

\newcolumntype{C}[1]{>{\centering\let\newline\\\arraybackslash\hspace{0pt}}m{#1}}
\newcolumntype{L}[1]{>{\raggedright\let\newline\\\arraybackslash\hspace{0pt}}m{#1}}
\newcolumntype{R}[1]{>{\raggedleft\let\newline\\\arraybackslash\hspace{0pt}}m{#1}}

\definecolor{cvprblue}{rgb}{0.21,0.49,0.74}
\usepackage[pagebackref,breaklinks,colorlinks,citecolor=cvprblue]{hyperref}

\def\paperID{715} 
\def\confName{CVPR}
\def\confYear{2024}

\title{Specularity Factorization for Low-Light Enhancement}

\author{Saurabh Saini\\
CVIT, KCIS\\
IIIT-Hyderabad, India\\
{\tt\small saurabh.saini@research.iiit.ac.in}
\and
P J Narayanan\\
CVIT, KCIS\\
IIIT-Hyderabad, India\\
{\tt\small pjn@iiit.ac.in}
}

\begin{document}
\twocolumn[{
\renewcommand\twocolumn[1][]{#1}
\maketitle
\begin{center}
    \centering
    \vspace*{-1\baselineskip}
    \captionsetup{type=figure}
    \includegraphics[width=\linewidth,height=5cm]{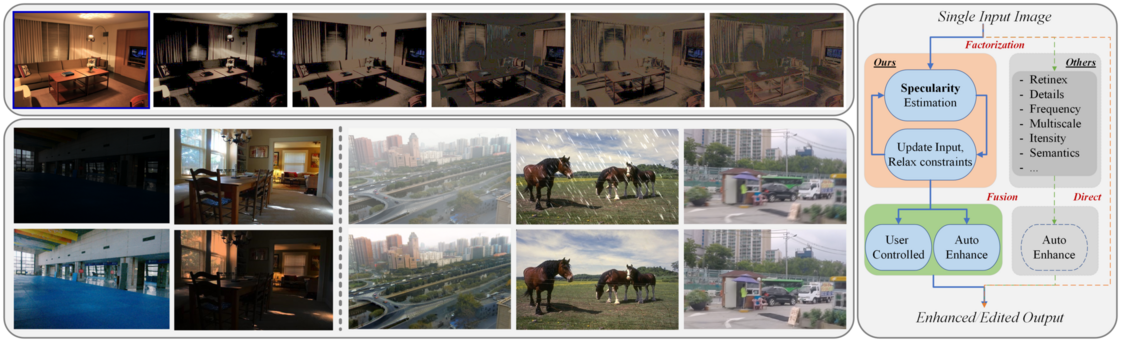}
    \vspace*{-1.5\baselineskip}
    \captionof{figure}
    {\textbf{Sepcularity Factorization}: We factorize a single input image (blue box, top row) into multiple \textit{soft} specular factors (rescaled for visualization) based on their similar illumination characteristics (note table shadow and lamp reflection). Our factors directly enable zero-reference low-light enhancement and user controlled image relighting (bottom left). Additionally, they can also be used as a plug-and-play prior for various supervised image enhancement tasks like dehazing, deraining and deblurring. On right, our conceptual block diagram.}
    \vspace*{0.5\baselineskip}
    \label{fig:teaser}
\end{center}

}]
\begin{abstract}
We present a new additive image factorization technique that treats images to be composed of multiple latent specular components which can be simply estimated recursively by modulating the sparsity during decomposition.
Our model-driven {\em RSFNet} estimates these factors by unrolling the optimization into network layers requiring only a few scalars to be learned.
The resultant factors are interpretable by design and can be fused for different image enhancement tasks via a network or combined directly by the user in a controllable fashion.
Based on RSFNet, we detail a zero-reference Low Light Enhancement (LLE) application trained without paired or unpaired supervision. 
Our system improves the state-of-the-art performance on standard benchmarks and achieves better generalization on multiple other datasets.
We also integrate our factors with other task specific fusion networks for applications like deraining, deblurring and dehazing with negligible overhead thereby highlighting the multi-domain and multi-task generalizability of our proposed RSFNet.
The code and data is released for reproducibility on the project homepage \footnote{ \url{https://sophont01.github.io/data/projects/RSFNet/}}.
\vspace*{-1\baselineskip}
\end{abstract}    
\section{Introduction}
\label{sec:intro}
A low-light image has most regions too dark for comprehension due to low  exposure setting or insufficient scene lighting which makes images highly challenging for computer processing and aesthetically unpleasant. Low-Light Enhancement (LLE) aims to recovers a well-exposed image from a low-light input \cite{LLESurvey}.
LLE can be a critical pre-processing step before the downstream applications \cite{Exdark,SCI}.
Core LLE challenge lies in modeling the degradation function which is spatially varying and has complex dependence on multiple variables like color, camera sensitivity, illuminant spectra, scene geometry, \etc. 

Most LLE solutions decompose the image into meaningful latent factors based on a relevant optical property (\cref{tab:facRW}). This allows individual manipulation of each factor which simplifies the enhancement operation. A common factorization is based on the Retinex approximation \cite{retinexTheory_77, retinex50}, which assumes a multiplicative disentanglement of image $I$ into two intrinsic factors: illumination-invariant, piecewise constant {\em reflectance} $R$ and color-invariant, smooth {\em shading} $S$ as $I=R \cdot S$.
Other factorization criteria include frequency \cite{freqLLE,FEC}, spatial scale \cite{afifiEC,DSLR}, spatio-frequency representation \cite{halfWave,WDRN}, intensity \cite{sef}, reflectance rank \cite{LR3M,QSEF}, \etc.
Fixed number of factors \cite{retinexNet2018,FEC,LR3M} and variable number that allow better representation \cite{sef,afifiEC,DSLR} have been used. Some decompose image multiplicatively like Retinex \cite{retinexNet2018,FEC}, while others split into additive factors which are numerically more stable \cite{sefIPOL,WDRN,specCGF}.
Pixel segmentation could be soft or hard based on the membership across factors, with the former introducing fewer artifacts \cite{aksoySoftSeg}. LLE solutions can be \textit{global} or \textit{local}. Global methods use whole image level statistics like gamma \cite{zeroDCE}, histograms \cite{HEP}, \etc., to enhance the images. Local methods employ spatially varying features like illumination maps \cite{retinexNet2018}, intensity/segmentation masks \cite{sef,PSENet}, \etc., for the same. Global methods are simpler but local ones can capture scene semantics better. 

Traditional LLE methods used manually-designed model-based optimisation by deriving specific priors from the image itself \cite{LIME,SRIE,DUAL}, needing no training. Data-driven, machine learning based solutions have done better recently. They use training datasets to tune the model which generalizes to other images \cite{SNR,afifiEC,retinexNet2018}. {\em Supervised} methods require annotated input-output pairs of images \cite{kind,UretinexNet,SNR}. {\em Unsupervised} methods require annotated training data but not necessarily paired \cite{jiang2021enlightengan,UNIE}. {\em Zero-reference} methods do not need annotated data and approach the problem by explicitly encoding the domain knowledge from training images \cite{zeroDCE,RUAS,SCI}. They generalize better and are simpler, lighter, and more interpretable by design.

In this paper, we present a zero-reference LLE method that outperforms prior methods on the average. At core is a novel {\em Recursive Specularity Factorization (RSF)} of the image factorization based on image specularity. We decompose an image into additive specular factors by thresholding the amount of sparsity of each pixel recursively. Successive factor differences mark out newly discovered image regions which are then individually targeted for enhancement. Our {\em RSFNet} that computes the factorization is model-driven, task-agnostic, and light-weight, needing about $200$ trainable parameters. The image factors are fused using a task-specific UNet-based module to enhance each region appropriately. RSF is useful to other applications when combined with other task-specific fusion modules. Our main contributions are:
\begin{itemize}
    \item A novel image factorization criterion and optimization formulation based on recursive specularity estimation.
    \item A model-driven RSFNet to learn factorization thresholds in a data-driven fashion using algorithm unrolling.
    \item A simple and flexible zero-reference LLE solution that surpasses the state of the art on multiple benchmarks and in the average generalization performance.
    \item Demonstration of RSF's usability to tasks like dehazing, deraining, and deblurring. RSF has high potential as a structural prior for several image understanding and enhancement tasks.
\end{itemize}

\begin{table}[t]
\scriptsize
\centering
    \begin{tabular}{l|c|c|c|c|r}
        \toprule
        \textbf{Criteria} & \textbf{No.} & \textbf{Type} & \textbf{Map} & \textbf{Seg} & \textbf{Example}\\
        \midrule
        Retinex & $2$ & $*$ & global & soft & \cite{UretinexNet, RUAS}  \\
        Frequency & $2$ & $+$, low/high & global & hard & \cite{freqLLE} \\
        Spectral & $2$ & $*$, fourier & global & soft & \cite{FEC} \\
        Low Rank & $2$ & $+$ & global & soft & \cite{LR3M,QSEF}\\        
        Wavelets & $2^n$ & $+$, pyramid & global & soft & \cite{WDRN, halfWave}\\
        Multiscale & $2^n$ & $+$, pyramid & global & soft & \cite{afifiEC,DSLR} \\
        Glare/Shadow & $3,4$ & $*,+$ & local & hard & \cite{shadingNet,Sharma2021NighttimeVE} \\
        Intensity & var. & $+$, bands & local & hard & \cite{sef,sefIPOL} \\
        \midrule
        Specularity & var. & $+$ & local & soft & RSFNet \\
        \bottomrule
    \end{tabular}
    \vspace{-0.5\baselineskip}
    \caption{\small Various LLE factorization criteria, with number of components (var. implies variable), type of factorization ($+$ additive/$*$ multiplicative), types of output maps (local/global), pixel segmentation across maps (soft/hard) and corresponding examplar methods. Our RSFNet proposes a novel specularity based factorization which allows a variable number of local soft-segmented factors.}
    \vspace{-1.5\baselineskip}
    \label{tab:facRW}
\end{table}
\begin{figure*}[t]
    \centering
    \includegraphics[width=\linewidth, height=45mm]{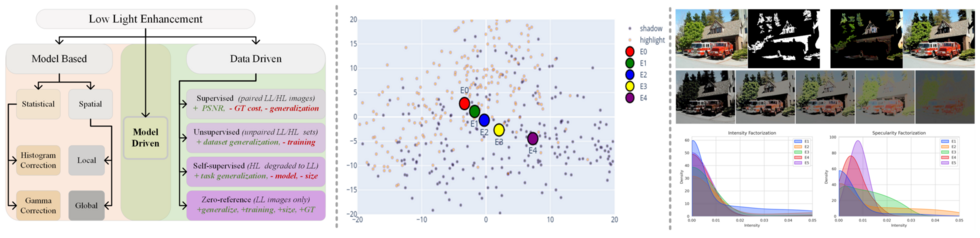}
    \vspace{-2\baselineskip}
    \caption{\textbf{Categorization and Motivation:} Left shows categorization of various LLE solution types (\cref{sec:relatedwork}). Middle plot shows the relationship between five factor cluster centers w.r.t each other and the background comprising of shadow/non-shadow regions estimated using PCA dimensionality reduced DINO features \cite{DINO}.  Gradual progression of feature cluster centers from highlight region to shadow region indicates their capability to capture various illumination regions in an image. Top right shows one data point from CHUK dataset \cite{CHUK} with mask, processed shadow/highlight regions and extracted factors. Bottom right plots distinguish our specular fuzzy factors from intensity thresholding based binary division, with ours allowing more diverse distributions and richer representation.}
    \label{fig:factExp}
\end{figure*}

\section{Background}
\label{sec:relatedwork}
\paragraph{Model-based LLE:} Early LLE solutions used traditional optimization models using either global statistics \cite{pizer1987adaptive,CLAHE,celik2011contextual,lee2013contrast} or spatially varying illumination maps for local editing \cite{LIME,NPE,fu2016fusion,zhang2018high}. 
They were more interpretable but required hand-crafted algorithms and heuristics.\vspace*{-5mm}

\paragraph{Data-driven LLE:} Modern solutions take inspiration from traditional techniques and induce domain knowledge via loss terms or designed within the network architecture which are learned from large datasets in a data-driven fashion.
They belong to one of the five training paradigms \citep{LLESurvey}. 
\textit{Supervised} LLE methods require both low-light and well-lit \textit{paired images} like \citet{retinexNet2018,kindPP,LOLv2,Sharma2021NighttimeVE,SNR}. 
On the other hand, \textit{unsupervised} methods like \citet{jiang2021enlightengan,UDEGan,HEP}, require only unpaired low-light and well-lit \textit{image sets}. 
\textit{Semi-supervised} methods combine the previous two techniques and use both paired and unpaired annotations \cite{BandNet,HybridNet}.
\textit{Self-supervised} solutions \citep{SSL_LLE,PSENet} generate their own annotations using pseudo-labels or synthetic degradations.
Contrary to all of these, \textit{zero-reference} methods do not use ground truth reconstruction losses and assess the quality of output based upon encoded prior terms \cite{zeroDCE,zeroDCEPP,RUAS,SCI,ExcNet,RRDNet}. 
These methods, like ours, possess improved generalizability due to explicit induction of domain knowledge and reduced chances of overfitting \cite{zeroDCE}. 
Zero-reference insights also provide direct valuable additions to the subsequent solutions in other paradigms. \vspace*{-5mm}

\paragraph{Model-Driven Networks:}
Data-driven solutions have good performance but lack interpretability, whereas model-based methods are explainable by design but often compromise with lower performance. 
Model-driven networks \cite{mongaUnrollingSurvey} are hybrids which bring the best of both together.
Such networks \textit{unroll} optimization steps as differentiable layers with learnable parameters, inducing data-driven priors in place of hand-crafted heuristics.
Although data-driven solutions are plenty, only a few model-driven solutions exist for low-level vision tasks like image restoration \cite{lecouat2020designing,lecouat2020fully}, shadow removal \cite{zhu2022efficient}, dehazing \cite{modelDehazing}, deraining \cite{modelRainRemoval}, denoising \cite{modelDenoise} and super-resolution \cite{deepRep,Bhat2021DeepBS}. 
Such solutions are concise and efficient due to underlying task specific formulation.

Model-driven LLE solutions are very recent.
UretinexNet \citep{UretinexNet} and UTVNet \citep{adptativeUretinex} are both supervised methods which respectively unroll the Retinex and total variational LLE formulations.
RUAS \cite{RUAS} and SCI \cite{SCI} are closest to our approach as they both propose model-driven zero-reference LLE solutions.
RUAS \cite{RUAS} unrolls illumination estimation and noise removal steps in their optimization and compliment it with learnable architecture search, towards a dynamic LLE framework. 
SCI \cite{SCI} on the other hand propose a residual framework wherein reflectance estimation is done by a self-calibration module which is then used to iteratively refine illumination maps.
In contrast, our method is inspired directly by image formation fundamentals and presents a novel factorization criterion which provides better interpretability, performance and flexibility.\vspace*{-5mm}
\begin{figure*}[th!]
    \centering
    \includegraphics[width=\linewidth,height=5.4cm] {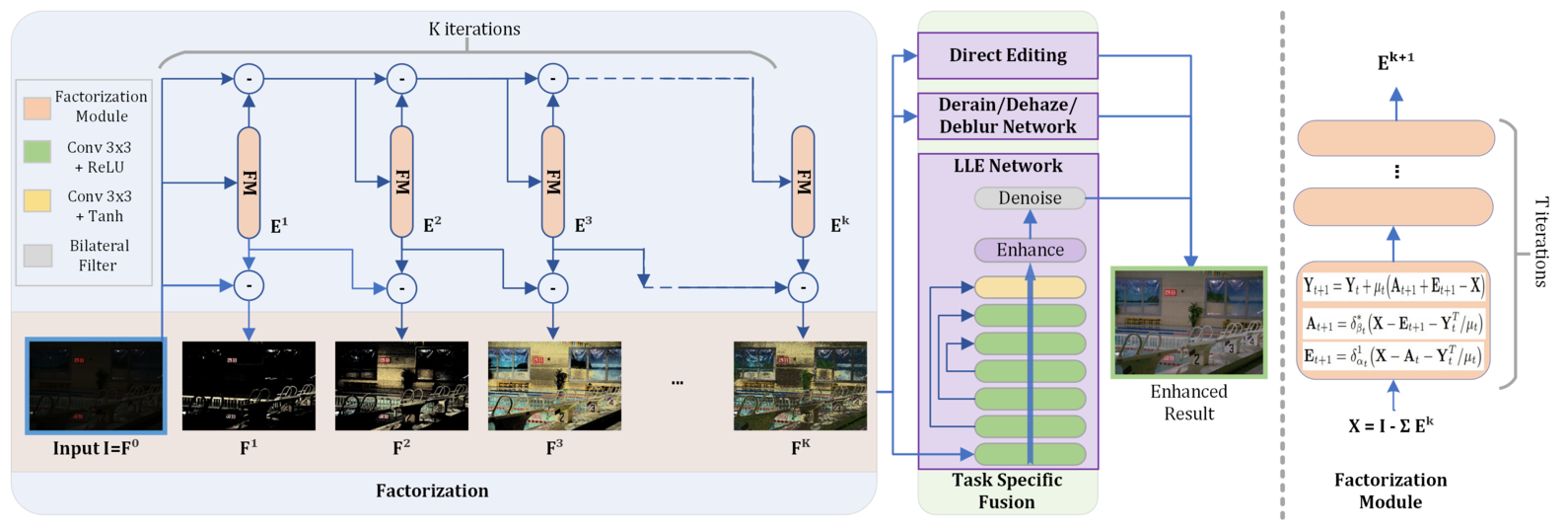}
    \vspace*{-6mm}
    \caption{\textbf{Block Diagram:} Our factorization module (RSFNet) splits a given image into multiple specular components using model-driven unrolled optimization. Then fusion module combines all the factors to generate the enhanced output.}
    \vspace{-1\baselineskip}
    \label{fig:blockDiagram}
\end{figure*}

\paragraph{Retinex Factorization for LLE:} 
Retinex \cite{retinexTheory_77,retinex50} is the most widely used factorization strategy for LLE \cite{LLESurvey, retinexNet2018, UretinexNet, RUAS, sainiIJCV} and beyond \cite{editingIID, garcesIIDsurvey, sainiBMVC, sainiFSIID}. 
One major Retinex limitation is due to the Lambertian reflection \cite{lambertianReflection_1760} assumption which approximates all surfaces as diffuse, thereby ignoring prevalent non-Lambertian effects in a real scene like specularity, translucency, caustics \etc.
Another issue is that pixel-wise multiplicative nature of Retinex factors is cumbersome to handle numerically (especially in LLE with near zero pixel values) and the obtained illumination maps require further semantic analysis for downstream applications.
Extensions of Retinex like dichromatic model \cite{dichromaticRefModel_1994} and shadow segmentation \cite{shadingNet}, separate one extra component each in addition to diffuse $R$ and $S$ \eg
\citet{Sharma2021NighttimeVE} and \citet{shadingNet} used glare and shadow image decomposition respectively. 
From  this perspective our recursive specular factorization can be understood as an extension of the same idea with continuously varying illumination characteristics starting from bright glares and ending with dark shadows (see \cref{fig:factExp} and \cref{sec:method} for details). \vspace*{-5mm}
\paragraph{Others Factorization Strategies:} Apart from Retinex, other factorization techniques are listed in Tab. \ref{tab:facRW}.
\citet{mertensExpFusion2,freqLLE,afifiEC,DSLR} employ spatial or frequency based image decomposition.
Recently, \citet{BandNet} used recursively concatenated features from a supervised encoder and \citet{FEC} proposed a Fourier disentanglement based solution. 
Apart from these supervised factorizations, \citet{SGZ} proposed semantic classification based ROI identification using a pretrained segmentation network.
\cite{zeroDCE,PSENet} predict multiple gamma correction maps for enhancement. 
\cite{sef,sefIPOL} simulate single image exposure burst using piece-wise thresholded intensity functions whereas \cite{LR3M} uses low-rank decomposition for reflectance. 
Each factorization strategy harnesses crucial underlying optical observations and adds valuable insights to the low-level vision research.
To the best of our knowledge, our proposed method here is the first to use recursive specularity estimation as a factorization strategy for LLE and other enhancement tasks. 

\section{Approach}
\label{sec:method}
\paragraph{Outline:} 
Our entire pipeline consists of two parts. 
We first decompose the image into $K$ factors using our {\em Recursive Specularity Factorization Network} (RSFNet),
which consists of multiple factorization modules (FM) with each optimization step encoded as a differentiable network layer.
Then we fuse, enhance, and denoise the factors using a {\em fusion network}, which is built using task dependent pre-existing architectures.
This modular design allows easy adoption of our technique in several other tasks and learning paradigms (\cref{sec:ExpNres}).
\subsection{Factorization Network: RSFNet}
\paragraph{Specularity Estimation:} 
Specularity removal is a well studied problem.
Most specularity removal methods \cite{eladSpec,specECCV,sumitSpec} exploit the relative sparsity of specular highlights and use pre-defined fixed sparsity thresholds to isolate the specular component.
According to dichromatic reflection model \cite{dichromaticRefModel_1994} image consists of a diffuse $\mathbf{A}$ and a specular $\mathbf{E}$ term: $\mathbf{X} = \mathbf{A} + \mathbf{E}$  for input $X$ where specular component can be estimated by minimizing the $L_0$ norm approximated as:
\begin{equation}
\label{eqn:sparsity}
\small
    \argmin_{\mathbf{E,A}} \norm{\mathbf{A}}_* +\lambda \norm{\mathbf{E}}_1  
    \quad  \text{s.t.} \;\; \mathbf{X}=\mathbf{A}+\mathbf{E},
\end{equation}
where $L_1$ is relaxation of $L_0$, $*$ is Frobenius norm regularizer and $\lambda$ is the sparsity parameter with higher values encouraging sparser results. 
\cref{eqn:sparsity} can be restated as augmented Lagrangian \citep{boydConvex} using dual form and auxiliary parameters ($\mathbf{Y},\mu$), which are then solvable using iterative ADMM updates ($t\in [0,T]$) \citep{boydADMM} as given below:
\begin{equation}
\label{eqn:ADMM}
\small
\begin{aligned}
    \mathbf{E}_{t+1}\! &= \delta^{1}_{\alpha_{t}} \big(\mathbf{X}-\mathbf{A}_{t}-\mathbf{Y}^T_{t}/\mu_t \big) 
    & \text{where}~ \alpha:\mathcal{F}(\lambda,\mu),\\
    \mathbf{A}_{t+1}\! &= \delta^{*}_{\beta_t} \big( \mathbf{X}-\mathbf{E}_{t+1}-\mathbf{Y}^T_{t}/\mu_t \big)
    & \text{where}~ \beta:\mathcal{F}(\mu),\\
    \mathbf{Y}_{t+1}\! &= \mathbf{Y}_{t} + \mu_t \big( \mathbf{A}_{t+1} + \mathbf{E}_{t+1} - \mathbf{X} \big)
    & \text{where}~ \mu:\mathcal{F}(\mathbf{X}).\\
\end{aligned}
\end{equation}
Here $\delta^{p}_\alpha$ is element-wise soft-thresholding operator \citep{boydProximal}:
\begin{equation*}
\label{eqn:shrinkage}
\small
     \delta^{p}_{\alpha}(x) = max( 1-{\alpha}/{\abs{x}_p}, ~ 0) \cdot x.
\end{equation*}
We can back-propagate through updates in \cref{eqn:ADMM} \cite{UretinexNet,adptativeUretinex} and hence can unroll them as neural network layers with learnable parameters $\boldsymbol{\alpha}:\{\alpha\}_0^T$,
$\boldsymbol{\beta}:\{\beta\}_0^T$ and $\boldsymbol{\mu}:\{\mu\}_0^T$.

\Paragraph{Relation with ISTA:} Analyzing the structure of \cref{eqn:ADMM}, we can draw parallels with the ISTA problem \cite{ISTA}, which seeks a sparse solution to $\mathbf{E}$ for the condition $\mathbf{X}=\mathcal{G}\mathbf{E}+\epsilon$, with $\mathcal{G}$ as a learnable dictionary and negligible $\epsilon$.
In contrast, we have a non-negligible residue and identity dictionary. 
LISTA by \citet{LISTA} showed how $\mathbf{E}$ update step can be represented as a weighted function which can then be approximated as finite network layers \ie:
\begin{equation}
\label{eqn:LISTA}
\small
    \mathbf{E}_{t+1} = \delta_{\alpha_t}(\mathbf{w}^1_t \mathbf{E}_t+\mathbf{w}^2_t \mathbf{X}),
\end{equation}
with learnable parameters $(\alpha_t,\mathbf{w}^1_t,\mathbf{w}^2_t)$ for each iteration $t \in [0,T]$.
Based on the weight coupling between $\mathbf{w}^1$ and $\mathbf{w}^2$, \citet{chenLISTA} simplified \cref{eqn:LISTA} by deriving both $\mathbf{w}^1$ and $\mathbf{w}^2$ from a single weight term, thereby halving the computation cost. 
A major simplification was further proposed by \citet{ALISTA} as ALISTA, who proved how all weight terms could be analytically obtained for a known dictionary, thereby leaving only step sizes and thresholds \ie $\mu$ and $\alpha_t$ to be estimated.
Later on this idea was extended to other similar optimization formulations and improved upon by additional simplifications and guarantees \eg \citet{LRPCA} unrolled their ADMM updates into a network for robust principal component analysis.

\Paragraph{Recursive Factorization:} Drawing parallels from ALISTA \cite{ALISTA} and its applications \cite{LRPCA}, we propose to learn the analytically reduced sparsity thresholds and step sizes via unrolled network layers.
After optimizing the above mentioned objective \cref{eqn:sparsity} we obtain one specular factor $\mathbf{E}^k$ where index $k \in [1,K]$ indicates the factor number.
For multiple factors, we recursively solve \cref{eqn:sparsity} by resetting the input $X$ after removing the previous specular output and relaxing the initial sparsity weight. 
We initialize variables for each factor at $t=0$ as: 
\begin{equation}
\label{eqn:newFactIn}
\begin{gathered}
\small
    \mathbf{X}^{k+1} = \mathbf{X}^k - \mathbf{E}^k,  ~~\quad~~ 
    \mathbf{Y}^{k} = \mathbf{X}^{k}/\norm{\mathbf{X}^k}_2  \\
    \alpha^k=(1-\nu^k)\hat{X}^k,  ~~\;~~  \beta^k=\nu^k\hat{X}^k,  ~~\;~~  \nu^k = k/K,\\
\end{gathered}
\end{equation}
where $\hat{X}$ indicates input mean and $\mathbf{X}^0 \textstyle{=} \mathbf{I}$. 
Intuitively, this can be understood as progressively removing specularity ($E^k$) from the original image by gradually relaxing the sparsity weight ($\alpha^{k+1} < \alpha^k$).
This lets us split the original image into multiple additive factors as:
\begin{equation}
\label{eqn:facts}
\small
    \mathbf{I} = \mathbf{E}^1 + \mathbf{E}^2 + \ldots + \mathbf{E}^K  = \textstyle{\sum}_{k=1 }^K\mathbf{E}^k
\end{equation}
\Paragraph{Unrolling:} Based upon above discussion, we propose an unrolled network collecting all parameters in a single vector $\boldsymbol{\theta}$. 
In each iteration $t$, we estimate three scalars: thresholds for both components ($\alpha_t$, $\beta_t$) and the step size ($\mu_t$). Hence for a factor $k$, we have $3T$ parameters $\boldsymbol{\theta}^k:=(\alpha^k,\beta^k,\mu^k)$ and overall we have only $3KT$ parameters
$\boldsymbol{\theta} := \{\boldsymbol{\theta}^k\}_1^K$.
Hence our model-driven factorization module is extremely light-weight compared to other decompositions (\cref{tab:quant}). 
We propose the following novel factorization loss:
\begin{equation}
\label{eqn:L_fact}
\small
    L_f = \lambda_f \sum_{k=1}^{K} L^k_f  
    ~~\quad~~ \text{where} ~~\quad~~
    L^k_f = \abs{\hat{E}^k / \hat{X}^k -\nu^k}.
\end{equation}
This constraints the ratio of signal energy in the $k^{th}$ factor compared to the input, to $\nu^k$.
As $\nu^k$ increases for higher factors, our factorization loss relaxes the sparsity constraint, thereby gradually increasing the number of pixels in the specular component.
After training, we are left with $K$ specular factors which sum to $I$.
As shown in \cref{fig:teaser} and \cref{fig:factExp}, each one of these factors highlights specific image regions with similar illumination characteristics which can be individually targeted for enhancement.

\Paragraph{Motivation/Validation:}
The core assumption behind our factorization is that an image can be split into multiple specular factors with each representing specific illumination characteristic. 
Although such factorization quality assessment is difficult to estimate \cite{editingIID, avaniICVGIP, garcesIIDsurvey}, we performed a toy experiment to validate our hypothesis using shadow detection dataset \cite{CHUK} which contains binary shadow masks in complex real world images (\cref{fig:factExp}).
We extract semantics-rich DINO image features \cite{DINO} after masking shadow and non-shadow image regions and visualize them in 2D using PCA. This marks separation of feature space between shadowed and highlighted regions in the background. The regions with progressively degrading illumination characteristics (glare, direct light, indirect light, soft shadow, dark shadow, \etc) are  expected to gradually lie between the two extremes.
Next we factorize each image into five factors using our approach and plot the cluster mean for each factor feature distribution on the same graph. We can observe in \cref{fig:factExp} that successive factors gradually shift from the non-shadow towards the shadowed feature space region mirroring the expected illumination order. This confirms that our factorization decomposes the pixel values across fundamental illumination types like glare, direct light, indirect light, shadow, \etc. 

We also plot the respective factor distribution densities of intensity factorization \cite{sef,sefIPOL} and our specularity factorization (\cref{fig:factExp}, bottom right). Intensity factorization allows little variation in the underlying factor distributions and imposes hard segmentation constraints with binary pixel masks. Our specular factors, on the other hand, permit higher variability and soft masks, with each pixel value spread across multiple factors. This provides more flexible representation and better optical approximation.
\newlength{\textfloatsepsave} 
\setlength{\textfloatsepsave}{\textfloatsep}
\setlength{\textfloatsep}{0pt}
\begin{center}
    \RestyleAlgo{ruled}
    \begin{algorithm}[t]
    \scriptsize
    \SetCommentSty{textit}
    \KwIn{Lowlight:$I$;  Hyperparams: $\lambda_{c|e|s}, K,T$}
    \KwOut{Enhanced: $O$;  Params: $\boldsymbol{\theta}$=$\{\boldsymbol{\alpha}\}^K_0,
    \{\boldsymbol{\beta}\}^K_0, \{\boldsymbol{\mu}\}^K_0 $ }
    \For{$e \gets 0$ \KwTo \text{num of epochs}}
    {
        \tcp{Train Factorization Module}
        \For{$k \gets 0$ \KwTo $K$}               
        {
            \For{$t \gets 0$ \KwTo $T$}
            {
                Initialize $E^k_0, A^k_0, Y^k_0$        \tcp*[r]{Eqn. \ref{eqn:newFactIn}}
                $E_t, A_t, Y_t \gets$ ADMM updates   \tcp*[r]{Eqn. \ref{eqn:ADMM}}
            }
            $F^k \gets E^{k}-E^{k-1}$           \tcp*[r]{Eqn. \ref{eqn:factDiff}}
        }
        Compute $L_f$                      \tcp*[r]{Eqn. \ref{eqn:L_fact}}
        \texttt{\\}
        \tcp{Train Fusion Module}
        \If{$e > \textit{freeze epoch}$}
        {
            Freeze all $\boldsymbol{\alpha}, \boldsymbol{\beta}, \boldsymbol{\mu}$ \;
            $L_f \gets 0$ \;
        }
        $I_{fuse} \gets \textit{Concatenate } [I, F^1, ..., F^K]$ \;
        $O \gets \textit{Forward } (I_{fuse})$ \;
        Compute $L$                             \tcp*[r]{Eqn. \ref{eqn:Loss}}
        Backpropagate $L$\;
    }
    \caption{\footnotesize LLE Training}
    \label{alg:RSFF}
    \end{algorithm}
    \vspace{-2\baselineskip}
\end{center}
\setlength{\textfloatsep}{\textfloatsepsave}
\begin{table*}[ht]
\centering
\resizebox{\textwidth}{!}{%
\scriptsize
\begin{tabular}{@{}p{12mm} || p{10mm}p{10mm}p{10mm} | p{10mm}p{10mm}p{10mm}p{10mm}p{10mm}p{10mm}p{10mm} | p{12mm}@{}}
\toprule
\toprule
Paradigm & \multicolumn{3}{c|}{\textbf{Traditional Model Based}} & \multicolumn{8}{c}{\textbf{Zero-reference}} \\
\midrule
Method & LIME \cite{LIME} & DUAL \cite{DUAL} & SDD \;\cite{SDD} & ECNet \cite{ExcNet} & ZDCE \cite{zeroDCE} & ZD++ \cite{zeroDCEPP} & RUAS \cite{RUAS} & SCI \quad \cite{SCI} & PNet \; \cite{PSENet} & GDP \;\cite{GDP} & \textbf{RSFNet} \;(Ours)\\
$\text{Params x}10^3$ & - & - & - & $16.5 \text{x} 10^3$ & $79.42$ & $10.56$ & $3.43$ & $\textbf{0.26}$ & $15.25$ & $552 \text{x} 10^3$ & $\underline{2.11}$ \\
\midrule
\multicolumn{11}{c}{\textbf{Lolv1} \cite{retinexNet2018} \qquad (dataset split: 485/15, mean$\approx 0.05$, resolution: $400 \times 600$)}\\
\midrule
$\text{PSNR}_y$ $\uparrow$  & $16.20$ & $15.97$ & $15.14$ & $18.01$ & $16.76$ & $16.38$ & $18.45$ & $16.45$ & $\underline{19.85}$ & $17.68$ & $\textbf{22.17}$\\
$\text{SSIM}_y$ $\uparrow$  & $0.695$ & $0.692$ & $0.754$ & $0.644$ & $0.734$ & $0.645$ & $\underline{0.766}$ & $0.709$ & $0.718$ & $0.678$ &  $\textbf{0.860}$\\
$\text{PSNR}_c$ $\uparrow$  & $14.22$ & $14.02$ & $13.34$ & $15.81$ & $14.86$ & $14.74$ & $16.40$ & $14.78$ & $\underline{17.50}$ & $15.80$ &  $\textbf{19.39}$\\
$\text{SSIM}_c$ $\uparrow$  & $0.521$ & $0.519$ & $\underline{0.634}$ & $0.469$ & $0.562$ & $0.496$ & $0.503$ & $0.525$ & $0.550$ & $0.539$ &  $\textbf{0.755}$\\
$\text{NIQE}$ $\downarrow$  & $8.583$ & $8.611$ & $\underline{3.706}$ & $8.844$ & $8.223$ & $8.195$ & $5.927$ & $8.374$ & $8.629$ & $6.437$ &  $\textbf{3.129}$\\
$\text{LPIPS}$$\downarrow$  & $0.344$ & $0.346$ & $\underline{0.278}$ & $0.358$ & $0.331$ & $0.346$ & $0.303$ & $0.327$ & $0.340$ & $0.375$ &  $\textbf{0.265}$\\
\midrule
\multicolumn{11}{c}{\textbf{Lolv2-real} \cite{LOLv2} \qquad (dataset split: 689/100, mean$\approx 0.05$, resolution: $400 \times 600$)} \\
\midrule
$\text{PSNR}_y$ $\uparrow$  & $19.31$ & $19.10$ & $18.47$ & $18.86$ & $\underline{20.31}$ & $19.36$ & $17.49$ & $19.37$ & $20.08$ & $15.83$ &  $\textbf{21.46}$\\
$\text{SSIM}_y$ $\uparrow$  & $0.705$ & $0.704$ & $\underline{0.792}$ & $0.613$ & $0.745$ & $0.585$ & $0.742$ & $0.722$ & $0.691$ & $0.627$ &  $\textbf{0.836}$\\
$\text{PSNR}_c$ $\uparrow$  & $17.14$ & $16.95$ & $16.64$ & $16.27$ & $\underline{18.06}$ & $17.36$ & $15.33$ & $17.30$ & $17.63$ & $14.05$ &  $\textbf{19.27}$\\
$\text{SSIM}_c$ $\uparrow$  & $0.537$ & $0.535$ & $\underline{0.678}$ & $0.459$ & $0.580$ & $0.442$ & $0.493$ & $0.540$ & $0.539$ & $0.502$ &  $\textbf{0.738}$\\
$\text{NIQE}$ $\downarrow$  & $9.076$ & $9.083$ & $\underline{4.191}$ & $9.475$ & $\underline{4.191}$ & $8.709$ & $6.172$ & $8.739$ & $9.152$ & $6.867$ &  $\textbf{3.769}$\\
$\text{LPIPS}$$\downarrow$  & $0.322$ & $0.324$ & $\textbf{0.280}$ & $0.360$ & $0.310$ & $0.340$ & $0.325$ & $\underline{0.294}$ & $0.340$ & $0.390$ &  $\textbf{0.280}$\\
\midrule
\multicolumn{11}{c}{GENERALIZED PERFORMANCE \quad \underline{\textbf{Mean Scores}} \qquad (\textbf{Lolv1} \cite{retinexNet2018}, \textbf{Lolv2-real} \cite{LOLv2}, \textbf{Lolv2-syn} \cite{LOLv2} and \textbf{VE-Lol} \cite{VELOL}) } \\
\midrule
$\text{PSNR}_y$ $\uparrow$ & $18.50$  &  $17.83$  &  $17.50$  &  $18.45$  &  $19.26$  &  $18.73$  &  $17.09$  &  $18.07$  &  $\underline{19.65}$  &  $15.88$  &  $\textbf{21.16}$ \\
$\text{SSIM}_y$ $\uparrow$ & $0.737$  &  $0.728$  &  $\underline{0.781}$  &  $0.677$  &  $0.777$  &  $0.674$  &  $0.743$  &  $0.745$  &  $0.743$  &  $0.634$  &  $\textbf{0.854}$ \\
$\text{PSNR}_c$ $\uparrow$ & $16.53$  &  $15.88$  &  $15.77$  &  $16.25$  &  $17.19$  &  $16.76$  &  $15.12$  &  $16.20$  &  $\underline{17.35}$  &  $14.15$  &  $\textbf{18.45}$ \\
$\text{SSIM}_c$ $\uparrow$ & $0.596$  &  $0.583$  &  $\underline{0.679}$  &  $0.538$  &  $0.634$  &  $0.548$  &  $0.532$  &  $0.587$  &  $0.605$  &  $0.504$  &  $\textbf{0.758}$ \\
$\text{NIQE}$ $\downarrow$ & $7.855$  &  $7.478$  &  $\underline{4.077}$  &  $7.543$  &  $4.270$  &  $7.468$  &  $5.841$  &  $7.626$  &  $7.791$  &  $6.726$  &  $\textbf{3.763}$ \\
$\text{LPIPS}$$\downarrow$ & $0.291$  &  $0.297$  &  $\textbf{0.266}$  &  $0.329$  &  $\underline{0.273}$  &  $0.296$  &  $0.346$  &  $0.295$  &  $0.302$  &  $0.379$  &  $0.276$ \\
\bottomrule
\bottomrule
\end{tabular}
}
\vspace{-0.5\baselineskip}
\caption{Quantitative comparison of our method RSFNet with other traditional and zero-reference solutions on multiple lowlight benchmarks and six evaluation metrics. Shown here are scores for two datasets LOLv1 and LOLv2-real with mean value across all datasets in the last sub-table (key: $\uparrow$ higher better; $\downarrow$ lower better; \textbf{bold}: best; \underline{underline}: second best).}
\vspace{-0.5\baselineskip}
\label{tab:quant}
\end{table*}
\subsection{Fusion Network}
In order to adhere to the zero-reference paradigm, we choose our fusion module to be a small fully-convolutional UNet like architecture with symmetric skip connections similar to other zero-reference methods \citep{zeroDCE,PSENet,SGZ}.
One fundamental difference is that we modify the architecture to harness multiple factors and simultaneously perform fusion, enhancement and denoising.
Specifically, it comprises of seven $3 \times 3$ convolutional layers with symmetric skip connections. 
We first pre-process all of our factors by subtracting the adjacent factors to discover the additional pixel values allowed in the current factor compared to the previous one as a soft mask:
\begin{equation}
\label{eqn:factDiff}
\small
    \mathbf{F}^k = \mathbf{E}^k - \mathbf{E}^{k-1} ~~\text{where}~~ \mathbf{F}^1 = \mathbf{E}^1.
\end{equation}
These factors are weighted if required using fixed scalar values and are then passed as a concatenated tensor into the fusion network. 
The output gamma maps $\mathbf{R}^k$ rescale different image regions differently and are applied directly on the original image inside the curve adjustment equation \cite{zeroDCE} for the fused result:
\begin{equation}
\label{eqn:dce}
\small
    O =  \Phi(\sum_{k=0}^K I + R^k . \big((I)^2-I \big)).
\end{equation}
The fused output is finally passed through a differentiable bilateral filtering layer $\Phi$ \cite{kornia} for the final enhanced result $O$.
Note that all the parameters from both factorization and fusion networks are trained together in end-to-end manner.
\Paragraph{Loss Terms:} We use two widely employed zero-reference losses for enhancement \cite{zeroDCE, PSENet, HEP} and one image smoothing loss for denoising.
First \textit{color loss} $L_c$ \cite{zeroDCE,HEP} is based on the gray-world assumption which tries to minimize the mean value difference between each color channel pair:
\begin{equation*}
\label{eqn:Lcolor}
\small
    L_c = \sum_{(i,j)\in C} \big( \hat{O}^i -\hat{O}^j \big)^2, ~~~~ C \in \{(r,g), (g,b), (b,r)\}. 
\end{equation*}
Second is the \textit{exposure loss} $L_e$ \cite{mertensExpFusion2,zeroDCE,sef}, which penalizes grayscale intensity deviation from the mid-tone value:
\begin{equation*}
\label{eqn:Lexp}
\small
    L_e = \frac{1}{|\Omega|} \sum_{\Omega} ~~ \big(\phi(O) - 0.6 \big)^2 ~~\text{where}~~ \Omega\in \{c\times h \times w\},
\end{equation*}
where $\phi$ represents the average value over a $16\times 16$ window.
Our third loss is the pixel-wise \textit{smoothing loss} which controls the local gradients $\nabla_{x|y}$ in the final output:
\begin{equation*}
\label{eqn:Ltv}
\small
    L_{s} = \frac{1}{|\Omega|} \sum_{\Omega} \big( (\nabla_x O)^2 + (\nabla_y O)^2 \big),
\end{equation*}
Note that this differs from the previous works who focus on total variational loss of the gamma maps instead. 
Our final training loss with $\lambda$'s as respective loss weights, is given as:
\begin{equation}
\label{eqn:Loss}
    L = \lambda_f L_f + \lambda_c L_c + \lambda_e L_e + \lambda_s L_s.
\end{equation}

\begin{table*}[h!]
    \begin{minipage}[b]{0.24\linewidth}
    \footnotesize
        \begin{tabular}{ @{}p{14mm} || p{8mm} p{8mm}}
        \toprule
        \toprule
            Variants & \scriptsize $\text{PSNR}_y\uparrow$ & \scriptsize $\text{SSIM}_y\uparrow$ \\
        \midrule
            $w/o$ $L_{e}$ & $8.12$ & $0.238$\\
            $w/o$ $L_{c}$ & $16.05$ & $0.724$\\
            $w/o$ $L_{s}$ & $20.13$ & $0.846$\\
            $w/o$ $\text{\scriptsize Denoise}$ & $19.51$ & $0.756$\\
            $w/o$ $\text{\scriptsize \underline{Fusion}}$ & $\underline{19.32}$ & $\underline{0.830}$\\
            \midrule
            $\text{\textbf{Full}}$ & $\textbf{22.17}$ & $\textbf{0.860}$\\
        \bottomrule
        \bottomrule
       \end{tabular}
       \vspace{-0.5\baselineskip}
        \caption{Ablation analysis on five variants of our RSFNet (\cref{sec:ExpNres}).}
        \vspace{-0.5\baselineskip}
        \label{tab:abla}
    \end{minipage}\hfill
    \begin{minipage}[b]{0.72\linewidth}
    \centering
    \scriptsize
    \begin{tabular}{@{}p{12mm} || p{12mm}p{11mm}p{11mm}p{11mm}p{11mm}p{12mm} | p{14mm}@{}}
    \toprule
    \toprule
        \footnotesize NIQE$\downarrow$ \quad \& \textit{LOE}$\downarrow$ & \footnotesize ECNet \cite{ExcNet} & \footnotesize  ZDCE \;\cite{zeroDCE}  & \footnotesize  ZD++ \;\cite{zeroDCEPP}  & \footnotesize  RUAS \;\cite{RUAS}  & \footnotesize  PNet \qquad\cite{PSENet}  & \footnotesize  SCI \qquad\cite{SCI}  & \footnotesize  \textbf{RSFNet} (Ours) \\
    \midrule
        DICM \cite{DICM}    &  3.37|\textit{676.7} & 3.10|\textit{340.8} & \textbf{2.94}|\textit{511.9} & 4.89|\textit{1421} & 3.00|\textit{590.3} & 3.61|\textit{321.9} & 3.23|\textbf{\textit{303.1}}  \\
        LIME \cite{LIME}    &  \textbf{3.75}|\textit{685.1} & 3.79|\textit{135.0} & 3.89|\textit{332.2} & 4.26|\textit{719.9} & 3.84|\textit{223.2} & 4.14|\textit{75.5} & 3.80|\textbf{\textit{68.3}}  \\
        MEF  \cite{MEF}     &  3.30|\textit{863.3} & 3.31|\textit{164.3} & 3.18|\textit{458.5} & 4.08|\textit{784.2} & 3.25|\textit{363.0} & 3.43|\textbf{\textit{95.0}} & \textbf{3.00}|\textit{100.7}  \\
        NPE  \cite{NPE}     &  \textbf{3.24}|\textit{936.1} & 3.52|\textit{312.9} & 3.27|\textit{532.2} & 5.75|\textit{1399} & 3.29|\textit{601.1} & 3.89|\textit{239.8} & 3.31|\textbf{\textit{221.5}}  \\
        VV   \cite{VV}      &  2.15|\textit{292.}4 & 2.75|\textit{145.4} & 2.53|\textit{222.9} & 3.82|\textit{583.7} & 2.56|\textit{260.2} & 2.30|\textbf{\textit{109.0}} & \textbf{1.96}|\textbf{\textit{109.0}}  \\
    \midrule
        \footnotesize \textbf{Mean} & 3.16|\textit{690.7} & 3.29|\textit{219.7} & 3.16|\textit{411.5} & 4.56|\textit{981.7} & 3.19|\textit{407.5} & 3.47|\textit{168.2} & \textbf{3.06}|\textbf{\textit{160.5}}
        \\
    \bottomrule
    \bottomrule
    \end{tabular}
    \vspace{-0.5\baselineskip}
    \caption{Qualitative comparison using naturalness preserving metrics (NIQE $\downarrow$ | LOE $\downarrow$) on five no-reference benchmarks: DICM, LIME, MEF, NPE and VV (\textbf{best} scores in bold, lower is better).}
    \vspace{-0.5\baselineskip}
    \label{tab:qual_quant}
    \end{minipage}
\end{table*}


\section{Experiments and Results}
\label{sec:ExpNres}
We now report our implementation details, results and extensions. Please see the supplementary document for additional details and results.
\Paragraph{Setup:}
We implement our combined network end-to-end on a single Nvidia 1080Ti GPU in PyTorch.  
We directly use low-light RGB images as inputs without any additional pre-processing. 
We first train factorization module for $25$ epochs which we freeze and then optimize the fusion module for next $25$ epochs.
We use stochastic gradient descent for optimization with batch size of $10$ and $0.01$ as learning rate. Model hyper-parameters are fixed using grid search and the entire training take less than 30 minutes.
\begin{figure*}[h]
   \centering
   \begin{minipage}{0.48\linewidth}
       \centering
      \includegraphics[width=\linewidth,height=5cm]{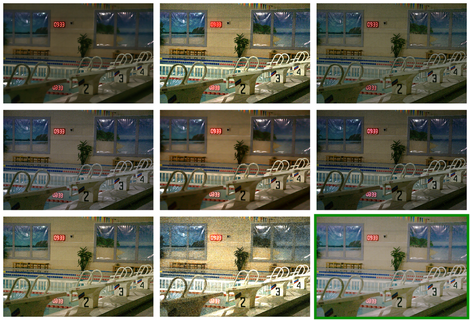}
   \end{minipage}
   \hspace{5mm}
  \begin{minipage}{0.48\linewidth}
       \centering
       \includegraphics[width=\linewidth,height=5cm]{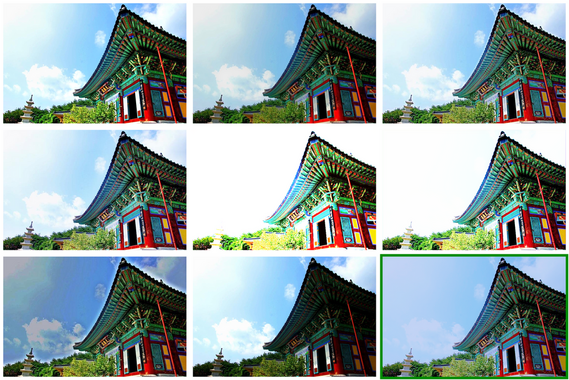}
   \end{minipage}
\vspace*{-0.5\baselineskip}
\caption{\textbf{Results:} Qualitative comparison of our method (green box) with other solutions (from top left 3 per row: SDD \cite{SDD}, ECNet \cite{ExcNet}, ZDCE \cite{zeroDCE}; ZD++ \cite{zeroDCEPP}, RUAS \cite{RUAS}, SCI \cite{SCI}; PNet \cite{PSENet}, GDP \cite{GDP} and  our RSFNet respectively). Our method generate natural looking images by handling noisy over and under exposed regions equally well, without over-saturating color or losing geometric details.}
\vspace*{-0.5\baselineskip}
\label{fig:qual}
\end{figure*}

\begin{figure}[b]
    \centering
    \begin{minipage}{0.60\columnwidth}
    \includegraphics[width=\linewidth, height=4.0cm]{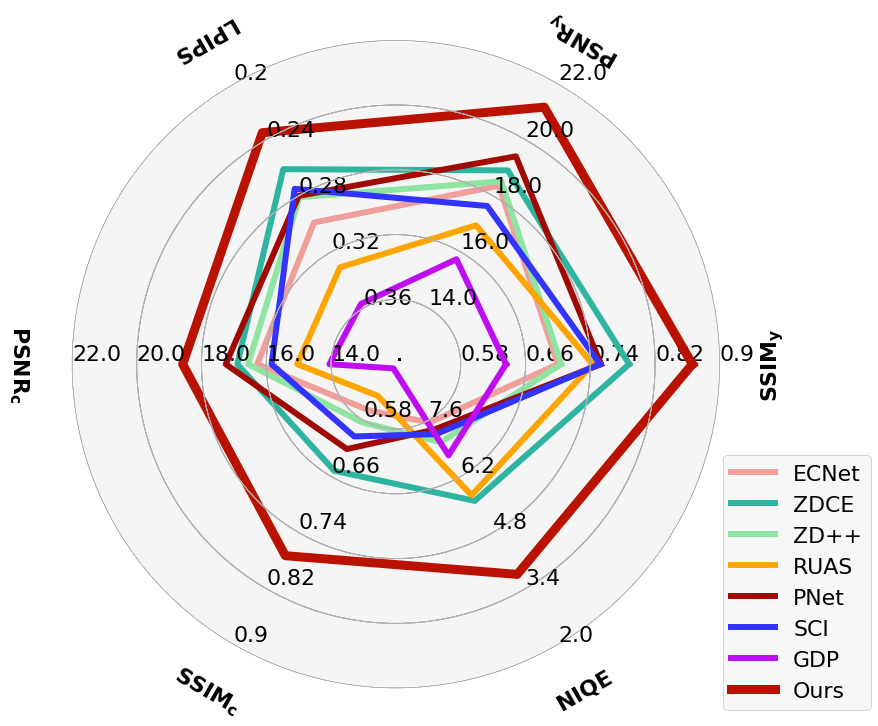}
    \end{minipage}\hfill
    \begin{minipage}{0.35\columnwidth}
    \includegraphics[width=\linewidth, height=4.6cm]{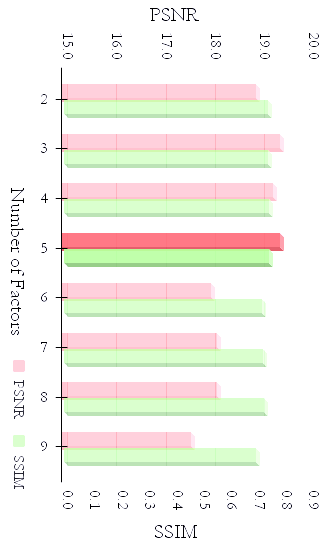}
    \end{minipage}
    \vspace{-0.5\baselineskip}
    \caption{\textbf{Analysis:} On left, our average score on all datasets \vs other methods (more area implies better). On right, ablation analysis with varying number of factors.}
    \vspace{-0.5\baselineskip}
    \label{fig:radar}
\end{figure}

\Paragraph{Datasets:}
We evaluate our method using multiple LLE benchmark datasets (Lolv1 \cite{retinexNet2018}, Lolv2-real \cite{LOLv2}, Lolv2-synthetic \cite{LOLv2} and VE-Lol \cite{VELOL}) with standard train/test splits (\cref{tab:quant}). 
These datasets comprise of several underexposed small-aperture inputs and corresponding well-exposed ground-truth pairs.
Here we report results on two datasets: Lolv1 and Lolv2-real and finally show the mean scores on all four datasets combined in the last sub-table \cref{tab:quant} and in \cref{fig:radar}.
Furthermore, we report generalization results (\cref{tab:qual_quant}) on five additional no-reference datasets which have significant domain shift: DICM \cite{DICM}, LIME \cite{LIME}, MEF \cite{MEF}, NPE \cite{NPE} and VV \cite{VV}.
\begin{figure*}[th!]
    \centering
   \includegraphics[width=\linewidth,height=4.4cm]{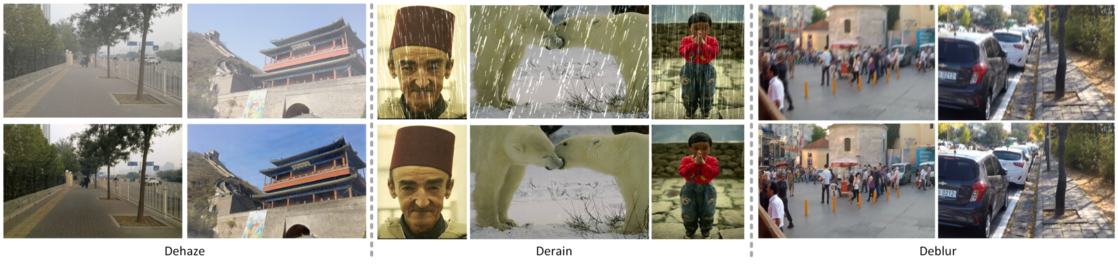}
    \vspace*{-2\baselineskip}
    \caption{Image enhancement applications using our specular factors as inputs on the AirNet \cite{AirNet} base model. Shown here left-to-right are our results for Dehazing \cite{dehazeDataset}, Deraining \cite{derainDataset} and Deblurring \cite{deblurDataset} tasks respectively using AirNet \cite{AirNet} as base model.}
    \vspace*{-0.5\baselineskip}
    \label{fig:apps}
\end{figure*}

\Paragraph{Metrics:} We report both single channel (Y from YCbCr) and multichannel (RGB) performance scores. 
As full-reference metrics (which require ground truth), we use Peak Signal to Noise-Ratio (PSNR), Structural Similarity Index Metric (SSIM) \cite{SSIM} and Learned Perceptual Image Patch Similarity (LPIPS) \cite{LPIPS}.
For no-reference assessment (without ground truth), we report Naturalness Image Quality Evaluator (NIQE) \cite{NIQE} and Lightness Order Error \cite{LOE}. 
Note while PSNR and SSIM gauge performance quantitatively, other three metrics estimate perceptual quality. 

\Paragraph{Comparisons:}
We compare against three model-based traditional optimization methods: LIME \cite{LIME}, DUAL \cite{DUAL} and SDD \cite{SDD} (others ignored due to low performance).
For data-driven methods we use seven recent zero-reference methods
(chronologically ordered): ECNet \cite{ExcNet}, zeroDCE \cite{zeroDCE}, zeroDCE++ \cite{zeroDCEPP}, RUAS \cite{RUAS}, SCI \cite{SCI}, PNet \cite{PSENet} and GDP \cite{GDP}.
We use the official code releases with pretrained weights and default parameters for results generation.
Quantitative and qualitative performance comparison is shown in \cref{tab:quant} and \cref{fig:qual} respectively.
Qualitatively, our method is cleaner with fewer artifacts and natural illumination (\cref{fig:qual}). 
This is validated by perceptual metrics like NIQE, LPIPS and LOE scores (\cref{tab:quant,tab:qual_quant}).
Our method outperforms other similar category contemporary solutions on multiple metrics and achieves the best generalization performance across datasets. For a generalized performance, we take mean of all the scores across benchmarks and graphically show them in the polar plot in \cref{fig:radar}. Each polygon represents a separate LLE method with higher area inside indicating better performance.

\Paragraph{Ablation:}
To validate our design choices, we conduct ablation study on several variants of our methods using Lolv1 dataset.
The effect of different number of factors $K$ on the final PSNR and SSIM scores are shown on right in \cref{fig:radar}.
We choose the best observed hyper-parameter settings $K\textstyle{=}5$ for all our experiments.
The effect of various loss terms after removing them one at a time (\ie $w/o ~ L_{e|c|s}$) and the effect of the final denoising step are shown in \cref{tab:abla}.
The last variant ({\it w/o} Fusion) represents an especially interesting setting where the fusion network is totally removed and inference uses only 
$3KT$ (=3*5*3=45) parameters. Fusion now reduces to a running average of the current image and the next factor, weighted by the normalized mean:
\begin{equation}
\label{eqn:factOnly}
\small
    O^{k+1} = (1-w^k) O^{k} + w^k F^k,  ~~\text{where}~~ w^k=\hat{F}^k/\textstyle{\sum}_k{\hat{F}^k}.  
\end{equation}
Even without any other zero-reference losses and using only a simple linear fusion, this method performs well, which demonstrates the effectiveness of our factors.
Note here we have an order of magnitude smaller network size than SCI (0.045 \vs 0.26 thousand parameters in \cref{tab:quant}).
\begin{figure}[b]
    \centering
   \includegraphics[width=\linewidth,height=3.4cm]{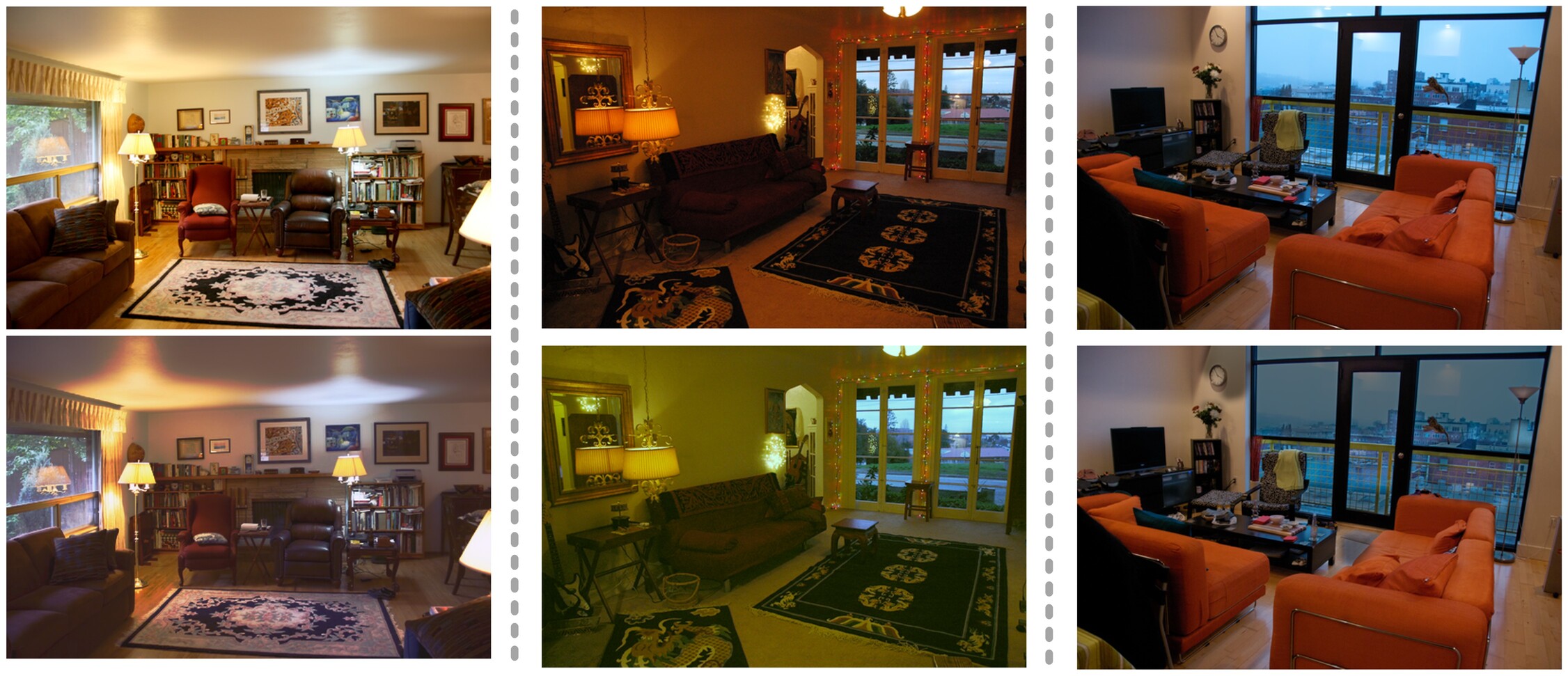}
    \vspace{-1.5\baselineskip}
    \caption{Controlable relighting applications using our factors as layers \cite{GIMP} (top:inputs; bottom:results; from left:edited light specularity, indoor color and outdoor intensity respectively).}
    \vspace{-0.5\baselineskip}
    \label{fig:userApps}
\end{figure}


\Paragraph{Extensions:}
Our specular factors are easily interpretable and can be used directly for image manipulation as image layers in standard image editing tools like GIMP \cite{GIMP}, Photoshop \cite{Photoshop}, \etc. We show an image relighting example by varying the color and blending modes of factors in (\cref{fig:teaser} bottom left, \cref{fig:userApps}). This indicates the potential of our factorization to complex downstream applications. We explore three diverse image enhancement tasks: dehazing, deraining and deblurring.
Here our goal is to evaluate the use of specularity factorization as a pre-processing step on an existing base model.
We chose the recent AirNet \cite{AirNet} as it allows experimentation on multiple image enhancement tasks with minor backbone modification. 
To induce our factors as prior information, we concatenate them along with the original input and alter the first convolutional layer input channels.
Note that we do not introduce any new loss or layers and directly train the model for three tasks one by one: \textit{(i)} Dehazing on RESIDE dataset \cite{dehazeDataset} \textit{(ii)} Deraining on Rain100L dataset \cite{derainDataset} and \textit{(iii)} Debluring on GoPro dataset \cite{deblurDataset}. 
As seen in \cref{fig:apps} and \cref{tab:app}, our results are perceptually more pleasing and improve the previously reported scores from multi-task methods consistently \cite{IDR,AirNet}.
We believe this is due to the induction of structural prior in the form of illumination based region categorization as the intensity and order of illumination at a pixel depends on the scene structure.
See the supplementary for more results.\vspace*{-2mm}
\begin{table}[bt]
    \scriptsize
    \resizebox{\columnwidth}{!}{%
    \begin{tabular}{l||cc|cc|cc}
    \toprule
    \toprule
        TASK $\xrightarrow{}$ & \multicolumn{2}{c|}{DEHAZE \cite{dehazeDataset}} & \multicolumn{2}{c|}{DERAIN \cite{derainDataset}} & \multicolumn{2}{c}{DEBLUR \cite{deblurDataset}} \\
    \midrule
        Method & $\text{PSNR}$  & $\text{SSIM}$  & $\text{PSNR}$  & $\text{SSIM}$  & $\text{PSNR}$  & $\text{SSIM}$  \\
    \midrule
        AirNet (multi-task) & 21.04 & 0.884 & 32.98 & 0.951 & 24.35 & 0.781 \\
        AirNet (uni-task) & 23.18 & 0.900 & 34.90 & 0.9657 & 26.42 & 0.801  \\
        \textbf{AirNet + Ours} & \textbf{24.96} & \textbf{0.9292} & \textbf{36.19} & \textbf{0.9718} & \textbf{27.29} & \textbf{0.827} \\
    \bottomrule
    \bottomrule
    \end{tabular}
    }
    \vspace*{-0.5\baselineskip}
    \caption{Our factors can induce structure prior in an existing base model and improve performance for multiple enhancement tasks.}
    \vspace*{-0.5\baselineskip}
    \label{tab:app}
\end{table}\vspace*{-2mm}

\Paragraph{Limitations:} 
Our method is sensitive to initialization conditions like the underlying algorithms \cite{ALISTA, LRPCA}.
As a heuristic we use dataset mean for initialization.
Another idea, to be explored in future, is to dynamically adapt to each input which is expected to further increase the performance.\vspace*{-2mm}

\Paragraph{Acknowledgements:} We acknowledge the support of TCS Foundation and the Kohli Centre on Intelligent Systems for this research.
\vspace*{-2mm}

\section{Conclusions}
\label{sec:conc}
In this paper, we presented a recursive specularity factorization (RSF) and its application to zero-reference LLE. We learn optimization hyperparameters in a data-driven fashion by unrolling the stages into a small neural network. The factors are fused using a network to yield the final result. We also demonstrate the utility of RSFs for image relighting as well as for image enhancement tasks like dehazing, deraining and deblurring. We are exploring the extension of RSFs to applications like image harmonization, foreground matting, white-balancing, depth estimation, \etc, and extend the technique to other signals beyond the visible spectrum.\vspace{-3mm}
\paragraph{Ethical Concerns:} This work enhances captured images and poses no special ethical issues we are aware of.


{
    \bibliographystyle{ieeenat_fullname}
    \bibliography{main}

\begin{thebibliography}{108}
\providecommand{\natexlab}[1]{#1}
\providecommand{\url}[1]{\texttt{#1}}
\expandafter\ifx\csname urlstyle\endcsname\relax
  \providecommand{\doi}[1]{doi: #1}\else
  \providecommand{\doi}{doi: \begingroup \urlstyle{rm}\Url}\fi

\bibitem[Adler et~al.(2013)Adler, Elad, Hel-Or, and Rivlin]{eladSpec}
Amir Adler, Michael Elad, Yacov Hel-Or, and Ehud Rivlin.
\newblock Sparse coding with anomaly detection.
\newblock In \emph{2013 IEEE MLSP}, 2013.

\bibitem[{Adobe Inc.}(2023)]{Photoshop}
{Adobe Inc.}
\newblock Adobe photoshop, 2023.

\bibitem[Afifi et~al.(2021)Afifi, Derpanis, Ommer, and Brown]{afifiEC}
Mahmoud Afifi, Konstantinos Derpanis, Bjorn Ommer, and Michael Brown.
\newblock Learning multi-scale photo exposure correction.
\newblock In \emph{CVPR}, 2021.

\bibitem[Aksoy et~al.(2018)Aksoy, Oh, Paris, Pollefeys, and
  Matusik]{aksoySoftSeg}
Ya\u{g}{\i}z Aksoy, Tae-Hyun Oh, Sylvain Paris, Marc Pollefeys, and Wojciech
  Matusik.
\newblock Semantic soft segmentation.
\newblock \emph{ACM ToG (SIGGRAPH)}, 37\penalty0 (4), 2018.

\bibitem[Baslamisli et~al.(2021)Baslamisli, Das, Le, Karaoglu, and
  Gevers]{shadingNet}
Anil~S. Baslamisli, Partha Das, Hoang-An Le, Sezer Karaoglu, and Theo Gevers.
\newblock Shadingnet: Image intrinsics by fine-grained shading decomposition.
\newblock \emph{IJCV}, 129\penalty0 (8), 2021.

\bibitem[Bell et~al.(2014)Bell, Bala, and Snavely]{IIW}
Sean Bell, Kavita Bala, and Noah Snavely.
\newblock Intrinsic images in the wild.
\newblock \emph{ACM Trans. on Graphics (SIGGRAPH)}, 33\penalty0 (4), 2014.

\bibitem[Bhat et~al.(2021{\natexlab{a}})Bhat, Danelljan, Gool, and
  Timofte]{Bhat2021DeepBS}
Goutam Bhat, Martin Danelljan, Luc~Van Gool, and Radu Timofte.
\newblock Deep burst super-resolution.
\newblock \emph{CVPR}, 2021{\natexlab{a}}.

\bibitem[Bhat et~al.(2021{\natexlab{b}})Bhat, Danelljan, Yu, Gool, and
  Timofte]{deepRep}
Goutam Bhat, Martin Danelljan, Fisher Yu, Luc~Van Gool, and Radu Timofte.
\newblock Deep reparametrization of multi-frame super-resolution and denoising.
\newblock \emph{ICCV}, 2021{\natexlab{b}}.

\bibitem[Bonneel et~al.(2017)Bonneel, Kovacs, Paris, and Bala]{editingIID}
Nicolas Bonneel, Balazs Kovacs, Sylvain Paris, and Kavita Bala.
\newblock Intrinsic decompositions for image editing.
\newblock \emph{Computer Graphics Forum (Eurographics State of the Art
  Reports)}, 36\penalty0 (2), 2017.

\bibitem[Boyd and Vandenberghe(2004)]{boydConvex}
Stephen Boyd and Lieven Vandenberghe.
\newblock \emph{Convex Optimization}.
\newblock Cambridge University Press, 2004.

\bibitem[Boyd et~al.(2011)Boyd, Parikh, Chu, Peleato, and Eckstein]{boydADMM}
Stephen Boyd, Neal Parikh, Eric Chu, Borja Peleato, and Jonathan Eckstein.
\newblock Distributed optimization and statistical learning via the alternating
  direction method of multipliers.
\newblock \emph{Foundations and Trends in Machine Learning}, 3\penalty0 (1),
  2011.

\bibitem[Cai et~al.(2021)Cai, Liu, and Yin]{LRPCA}
HanQin Cai, Jialin Liu, and Wotao Yin.
\newblock Learned robust pca: A scalable deep unfolding approach for
  high-dimensional outlier detection.
\newblock \emph{NeurIPS}, 34, 2021.

\bibitem[Cai et~al.(2023)Cai, Bian, Lin, Wang, Timofte, and Zhang]{RFormer}
Yuanhao Cai, Hao Bian, Jing Lin, Haoqian Wang, Radu Timofte, and Yulun Zhang.
\newblock Retinexformer: One-stage retinex-based transformer for low-light
  image enhancement.
\newblock In \emph{ICCV}, 2023.

\bibitem[Caron et~al.(2021)Caron, Touvron, Misra, J\'egou, Mairal, Bojanowski,
  and Joulin]{DINO}
Mathilde Caron, Hugo Touvron, Ishan Misra, Herv\'e J\'egou, Julien Mairal,
  Piotr Bojanowski, and Armand Joulin.
\newblock Emerging properties in self-supervised vision transformers.
\newblock In \emph{Proceedings of the International Conference on Computer
  Vision (ICCV)}, 2021.

\bibitem[Celik and Tjahjadi(2011)]{celik2011contextual}
Turgay Celik and Tardi Tjahjadi.
\newblock Contextual and variational contrast enhancement.
\newblock \emph{IEEE TIP}, 20\penalty0 (12), 2011.

\bibitem[Chen et~al.(2018)Chen, Liu, Wang, and Yin]{chenLISTA}
Xiaohan Chen, Jialin Liu, Zhangyang Wang, and Wotao Yin.
\newblock Theoretical linear convergence of unfolded ista and its practical
  weights and thresholds.
\newblock In \emph{NeurIPS}, 2018.

\bibitem[Daubechies et~al.(2003)Daubechies, Defrise, and Mol]{ISTA}
Ingrid Daubechies, Michel Defrise, and Christine~De Mol.
\newblock An iterative thresholding algorithm for linear inverse problems with
  a sparsity constraint.
\newblock \emph{Communications on Pure and Applied Mathematics}, 57, 2003.

\bibitem[Fan et~al.(2022)Fan, Liu, and Liu]{halfWave}
Chi-Mao Fan, Tsung-Jung Liu, and Kuan-Hsien Liu.
\newblock Half wavelet attention on m-net+ for low-light image enhancement.
\newblock In \emph{IEEE ICIP}, 2022.

\bibitem[Fan et~al.(2019)Fan, Chen, Yuan, Hua, Yu, and Chen]{multitaskDL}
Qingnan Fan, Dongdong Chen, Lu Yuan, Gang Hua, Nenghai Yu, and Baoquan Chen.
\newblock A general decoupled learning framework for parameterized image
  operators.
\newblock 2019.

\bibitem[Fei et~al.(2023)Fei, Lyu, Pan, Zhang, Yang, Luo, Zhang, and Dai]{GDP}
Ben Fei, Zhaoyang Lyu, Liang Pan, Junzhe Zhang, Weidong Yang, Tianyue Luo, Bo
  Zhang, and Bo Dai.
\newblock Generative diffusion prior for unified image restoration and
  enhancement.
\newblock In \emph{CVPR}, 2023.

\bibitem[Fu et~al.(2019)Fu, Zhang, Song, Lin, and Xiao]{specCGF}
Gang Fu, Qing Zhang, Chengfang Song, Qifeng Lin, and Chunxia Xiao.
\newblock Specular highlight removal for real-world images.
\newblock \emph{Computer Graphics Forum}, 38\penalty0 (7), 2019.

\bibitem[Fu et~al.(2016{\natexlab{a}})Fu, Zeng, Huang, Liao, Ding, and
  Paisley]{fu2016fusion}
Xueyang Fu, Delu Zeng, Yue Huang, Yinghao Liao, Xinghao Ding, and John Paisley.
\newblock A fusion-based enhancing method for weakly illuminated images.
\newblock \emph{Signal Processing}, 129, 2016{\natexlab{a}}.

\bibitem[Fu et~al.(2016{\natexlab{b}})Fu, Zeng, Huang, Zhang, and Ding]{SRIE}
Xueyang Fu, Delu Zeng, Yue Huang, Xiao-Ping Zhang, and Xinghao Ding.
\newblock A weighted variational model for simultaneous reflectance and
  illumination estimation.
\newblock In \emph{CVPR}, 2016{\natexlab{b}}.

\bibitem[Fu et~al.(2023)Fu, Yang, Tu, Huang, Ding, and Ma]{PairLLE}
Zhenqi Fu, Yan Yang, Xiaotong Tu, Yue Huang, Xinghao Ding, and Kai-Kuang Ma.
\newblock Learning a simple low-light image enhancer from paired low-light
  instances.
\newblock In \emph{CVPR}, 2023.

\bibitem[Garces et~al.(2022)Garces, Rodriguez-Pardo, Casas, and
  Lopez-Moreno]{garcesIIDsurvey}
Elena Garces, Carlos Rodriguez-Pardo, Dan Casas, and Jorge Lopez-Moreno.
\newblock A survey on intrinsic images: Delving deep into lambert and beyond.
\newblock \emph{IJCV}, 2022.

\bibitem[Gregor and LeCun(2010)]{LISTA}
Karol Gregor and Yann LeCun.
\newblock Learning fast approximations of sparse coding.
\newblock In \emph{ICML}, 2010.

\bibitem[Guo et~al.(2020)Guo, Li, Guo, Loy, Hou, Kwong, and Runmin]{zeroDCE}
Chunle Guo, Chongyi Li, Jichang Guo, Chen~Change Loy, Junhui Hou, Sam Kwong,
  and Cong Runmin.
\newblock Zero-reference deep curve estimation for low-light image enhancement.
\newblock \emph{CVPR}, 2020.

\bibitem[Guo et~al.(2018)Guo, Zhou, and Wang]{specECCV}
Jie Guo, Zuojian Zhou, and Limin Wang.
\newblock Single image highlight removal with a sparse and low-rank reflection
  model.
\newblock In \emph{ECCV}, 2018.

\bibitem[Guo et~al.(2016)Guo, Li, and Ling]{LIME}
Xiaojie Guo, Yu Li, and Haibin Ling.
\newblock Lime: Low-light image enhancement via illumination map estimation.
\newblock \emph{IEEE TIP}, 26\penalty0 (2), 2016.

\bibitem[Gupta et~al.(2022)Gupta, Saini, and Narayanan]{avaniICVGIP}
Avani Gupta, Saurabh Saini, and P.~J. Narayanan.
\newblock Interpreting intrinsic image decomposition using concept activations.
\newblock In \emph{ACM ICVGIP}, 2022.

\bibitem[Hao et~al.(2020)Hao, Han, Guo, Xu, and Wang]{SDD}
Shijie Hao, Xu Han, Yanrong Guo, Xin Xu, and Meng Wang.
\newblock Low-light image enhancement with semi-decoupled decomposition.
\newblock \emph{IEEE TMM}, 22\penalty0 (12), 2020.

\bibitem[Hessel(2019)]{sefIPOL}
Charles Hessel.
\newblock {Simulated Exposure Fusion}.
\newblock \emph{{Image Processing On Line}}, 9, 2019.

\bibitem[Hessel and Morel(2020)]{sef}
Charles Hessel and Jean-Michel Morel.
\newblock An extended exposure fusion and its application to single image
  contrast enhancement.
\newblock In \emph{WACV}, 2020.

\bibitem[Hu et~al.(2021)Hu, Wang, Fu, Jiang, Wang, and Heng]{CHUK}
Xiaowei Hu, Tianyu Wang, Chi-Wing Fu, Yitong Jiang, Qiong Wang, and Pheng-Ann
  Heng.
\newblock Revisiting shadow detection: A new benchmark dataset for complex
  world.
\newblock \emph{IEEE TIP}, 30, 2021.

\bibitem[Huang et~al.(2022)Huang, Liu, Zhao, Yan, Zhang, Huang, Zhou, and
  Xiong]{FEC}
Jie Huang, Yajing Liu, Feng Zhao, Keyu Yan, Jinghao Zhang, Yukun Huang, Man
  Zhou, and Zhiwei Xiong.
\newblock Deep fourier-based exposure correction network with spatial-frequency
  interaction.
\newblock In \emph{ECCV}, 2022.

\bibitem[Jiang et~al.(2021)Jiang, Gong, Liu, Cheng, Fang, Shen, Yang, Zhou, and
  Wang]{jiang2021enlightengan}
Yifan Jiang, Xinyu Gong, Ding Liu, Yu Cheng, Chen Fang, Xiaohui Shen, Jianchao
  Yang, Pan Zhou, and Zhangyang Wang.
\newblock Enlightengan: Deep light enhancement without paired supervision.
\newblock \emph{IEEE TIP}, 30, 2021.

\bibitem[Jose~Valanarasu et~al.(2022)Jose~Valanarasu, Yasarla, and
  Patel]{TransWeather}
Jeya~Maria Jose~Valanarasu, Rajeev Yasarla, and Vishal~M. Patel.
\newblock Transweather: Transformer-based restoration of images degraded by
  adverse weather conditions.
\newblock In \emph{CVPR}, 2022.

\bibitem[Lambert(1760)]{lambertianReflection_1760}
Johann~Heinrich Lambert.
\newblock \emph{Photometria sive de mensura et gradibus luminis, colorum et
  umbrae}.
\newblock Klett, 1760.

\bibitem[Land(1977)]{retinexTheory_77}
Edwin~Herbert Land.
\newblock The retinex theory of color vision.
\newblock \emph{Scientific American}, 237 6, 1977.

\bibitem[Lecouat et~al.(2020{\natexlab{a}})Lecouat, Ponce, and
  Mairal]{lecouat2020designing}
Bruno Lecouat, Jean Ponce, and Julien Mairal.
\newblock Designing and learning trainable priors with non-cooperative games.
\newblock \emph{NeurIPS}, 2020{\natexlab{a}}.

\bibitem[Lecouat et~al.(2020{\natexlab{b}})Lecouat, Ponce, and
  Mairal]{lecouat2020fully}
Bruno Lecouat, Jean Ponce, and Julien Mairal.
\newblock Fully trainable and interpretable non-local sparse models for image
  restoration.
\newblock \emph{ECCV}, 2020{\natexlab{b}}.

\bibitem[Lee et~al.(2013{\natexlab{a}})Lee, Lee, and Kim]{DICM}
Chulwoo Lee, Chul Lee, and Chang-Su Kim.
\newblock Contrast enhancement based on layered difference representation of 2d
  histograms.
\newblock \emph{IEEE TIP}, 22\penalty0 (12), 2013{\natexlab{a}}.

\bibitem[Lee et~al.(2013{\natexlab{b}})Lee, Lee, and Kim]{lee2013contrast}
Chulwoo Lee, Chul Lee, and Chang-Su Kim.
\newblock Contrast enhancement based on layered difference representation of 2d
  histograms.
\newblock \emph{IEEE TIP}, 22\penalty0 (12), 2013{\natexlab{b}}.

\bibitem[Li et~al.(2019)Li, Ren, Fu, Tao, Feng, Zeng, and Wang]{dehazeDataset}
Boyi Li, Wenqi Ren, Dengpan Fu, Dacheng Tao, Dan Feng, Wenjun Zeng, and
  Zhangyang Wang.
\newblock Benchmarking single-image dehazing and beyond.
\newblock \emph{IEEE TIP}, 28\penalty0 (1), 2019.

\bibitem[Li et~al.(2022)Li, Liu, Hu, Wu, Lv, and Peng]{AirNet}
Boyun Li, Xiao Liu, Peng Hu, Zhongqin Wu, Jiancheng Lv, and Xi Peng.
\newblock {All-In-One Image Restoration for Unknown Corruption}.
\newblock In \emph{CVPR}, 2022.

\bibitem[Li et~al.(2021{\natexlab{a}})Li, Guo, Han, Jiang, Cheng, Gu, and
  Loy]{LLESurvey}
Chongyi Li, Chunle Guo, Linghao Han, Jun Jiang, Ming-Ming Cheng, Jinwei Gu, and
  Chen~Change Loy.
\newblock Low-light image and video enhancement using deep learning: A survey.
\newblock \emph{IEEE TPAMI}, 2021{\natexlab{a}}.

\bibitem[Li et~al.(2021{\natexlab{b}})Li, Guo, and Loy]{zeroDCEPP}
Chongyi Li, Chunle Guo, and Chen~Change Loy.
\newblock Learning to enhance low-light image via zero-reference deep curve
  estimation.
\newblock \emph{IEEE TPAMI}, 2021{\natexlab{b}}.

\bibitem[Liang et~al.(2022)Liang, Xu, Quan, Shi, and Ji]{SSL_LLE}
Jinxiu Liang, Yong Xu, Yuhui Quan, Boxin Shi, and Hui Ji.
\newblock Self-supervised low-light image enhancement using discrepant
  untrained network priors.
\newblock \emph{IEEE TCSVT}, 32\penalty0 (11), 2022.

\bibitem[Liang et~al.(2023)Liang, Li, Zhou, Feng, and Loy]{CLIP-LIT}
Zhexin Liang, Chongyi Li, Shangchen Zhou, Ruicheng Feng, and Chen~Change Loy.
\newblock Iterative prompt learning for unsupervised backlit image enhancement.
\newblock In \emph{ICCV}, 2023.

\bibitem[Lim and Kim(2021)]{DSLR}
Seokjae Lim and Wonjun Kim.
\newblock Dslr: Deep stacked laplacian restorer for low-light image
  enhancement.
\newblock \emph{IEEE TMM}, 23, 2021.

\bibitem[Liu et~al.(2019{\natexlab{a}})Liu, Chen, Wang, and Yin]{ALISTA}
Jialin Liu, Xiaohan Chen, Zhangyang Wang, and Wotao Yin.
\newblock {ALISTA}: Analytic weights are as good as learned weights in {LISTA}.
\newblock In \emph{ICLR}, 2019{\natexlab{a}}.

\bibitem[Liu et~al.(2021)Liu, Dejia, Yang, Fan, and Huang]{VELOL}
Jiaying Liu, Xu Dejia, Wenhan Yang, Minhao Fan, and Haofeng Huang.
\newblock Benchmarking low-light image enhancement and beyond.
\newblock \emph{IJCV}, 129, 2021.

\bibitem[Liu et~al.(2022)Liu, Xie, Zhang, Yuan, Chen, Zhou, Li, and Tian]{TAPE}
Lin Liu, Lingxi Xie, Xiaopeng Zhang, Shanxin Yuan, Xiangyu Chen, Wengang Zhou,
  Houqiang Li, and Qi Tian.
\newblock Tape: Task-agnostic prior embedding for image restoration.
\newblock In \emph{ECCV}, 2022.

\bibitem[Liu et~al.(2019{\natexlab{b}})Liu, Pan, Ren, and Su]{modelDehazing}
Yang Liu, Jinshan Pan, Jimmy Ren, and Zhixun Su.
\newblock Learning deep priors for image dehazing.
\newblock In \emph{ICCV}, 2019{\natexlab{b}}.

\bibitem[Loh and Chan(2019)]{Exdark}
Yuen~Peng Loh and Chee~Seng Chan.
\newblock Getting to know low-light images with the exclusively dark dataset.
\newblock \emph{CVIU}, 178, 2019.

\bibitem[Ma et~al.(2015)Ma, Zeng, and Wang]{MEF}
Kede Ma, Kai Zeng, and Zhou Wang.
\newblock Perceptual quality assessment for multi-exposure image fusion.
\newblock \emph{IEEE TIP}, 24\penalty0 (11), 2015.

\bibitem[Ma et~al.(2022)Ma, Ma, Liu, Fan, and Luo]{SCI}
Long Ma, Tengyu Ma, Risheng Liu, Xin Fan, and Zhongxuan Luo.
\newblock Toward fast, flexible, and robust low-light image enhancement.
\newblock In \emph{CVPR}, 2022.

\bibitem[{McCann}(2017)]{retinex50}
John~J. {McCann}.
\newblock {Retinex at 50: color theory and spatial algorithms, a review}.
\newblock \emph{Journal of Electronic Imaging}, 26, 2017.

\bibitem[Mertens et~al.(2009)Mertens, Kautz, and Reeth]{mertensExpFusion2}
Tom Mertens, Jan Kautz, and Frank~Van Reeth.
\newblock Exposure fusion: A simple and practical alternative to high dynamic
  range photography.
\newblock \emph{Computer Graphics Forum}, 28, 2009.

\bibitem[Mittal et~al.(2013)Mittal, Soundararajan, and Bovik]{NIQE}
Anish Mittal, Rajiv Soundararajan, and Alan~C. Bovik.
\newblock Making a “completely blind” image quality analyzer.
\newblock \emph{IEEE Signal Processing Letters}, 20\penalty0 (3), 2013.

\bibitem[Monga et~al.(2021)Monga, Li, and Eldar]{mongaUnrollingSurvey}
Vishal Monga, Yuelong Li, and Yonina~C. Eldar.
\newblock Algorithm unrolling: Interpretable, efficient deep learning for
  signal and image processing.
\newblock \emph{IEEE Signal Processing Magazine}, 38\penalty0 (2), 2021.

\bibitem[Nah et~al.(2017)Nah, Kim, and Lee]{deblurDataset}
Seungjun Nah, Tae~Hyun Kim, and Kyoung~Mu Lee.
\newblock Deep multi-scale convolutional neural network for dynamic scene
  deblurring.
\newblock In \emph{CVPR}, 2017.

\bibitem[Nguyen et~al.(2023)Nguyen, Tran, Nguyen, and Nguyen]{PSENet}
Hue Nguyen, Diep Tran, Khoi Nguyen, and Rang Nguyen.
\newblock Psenet: Progressive self-enhancement network for unsupervised
  extreme-light image enhancement.
\newblock In \emph{WACV}, 2023.

\bibitem[Ni et~al.(2020)Ni, Yang, Wang, Ma, and Kwong]{UDEGan}
Zhangkai Ni, Wenhan Yang, Shiqi Wang, Lin Ma, and Sam Kwong.
\newblock Towards unsupervised deep image enhancement with generative
  adversarial network.
\newblock \emph{IEEE TIP}, 29, 2020.

\bibitem[Parikh and Boyd(2014)]{boydProximal}
Neal Parikh and Stephen Boyd.
\newblock Proximal algorithms.
\newblock \emph{Foundations and Trends in Optimization}, 1\penalty0 (3), 2014.

\bibitem[Pizer et~al.(1987)Pizer, Amburn, Austin, Cromartie, Geselowitz, Greer,
  ter Haar~Romeny, Zimmerman, and Zuiderveld]{pizer1987adaptive}
Stephen~M Pizer, E~Philip Amburn, John~D Austin, Robert Cromartie, Ari
  Geselowitz, Trey Greer, Bart ter Haar~Romeny, John~B Zimmerman, and Karel
  Zuiderveld.
\newblock Adaptive histogram equalization and its variations.
\newblock \emph{CVGIP}, 39\penalty0 (3), 1987.

\bibitem[Puthussery et~al.(2020)Puthussery, Panikkasseril~Sethumadhavan,
  Kuriakose, and Charangatt~Victor]{WDRN}
Densen Puthussery, Hrishikesh Panikkasseril~Sethumadhavan, Melvin Kuriakose,
  and Jiji Charangatt~Victor.
\newblock Wdrn: A wavelet decomposed relightnet for image relighting.
\newblock In \emph{ECCV workshop}, 2020.

\bibitem[Ren et~al.(2022)Ren, Pan, and Huang]{modelDenoise}
Chao Ren, Yizhong Pan, and Jie Huang.
\newblock Enhanced latent space blind model for real image denoising via
  alternative optimization.
\newblock In \emph{NeurIPS}, 2022.

\bibitem[Ren et~al.(2020)Ren, Yang, Cheng, and Liu]{LR3M}
Xutong Ren, Wenhan Yang, Wen-Huang Cheng, and Jiaying Liu.
\newblock Lr3m: Robust low-light enhancement via low-rank regularized retinex
  model.
\newblock \emph{IEEE TIP}, 29, 2020.

\bibitem[Reza(2004)]{CLAHE}
Ali~M. Reza.
\newblock Realization of the contrast limited adaptive histogram equalization
  (clahe) for real-time image enhancement.
\newblock \emph{J. VLSI Signal Process. Syst.}, 38\penalty0 (1), 2004.

\bibitem[Riba et~al.(2020)Riba, Mishkin, Ponsa, Rublee, and Bradski]{kornia}
E. Riba, D. Mishkin, D. Ponsa, E. Rublee, and G. Bradski.
\newblock Kornia: an open source differentiable computer vision library for
  pytorch.
\newblock In \emph{WACV}, 2020.

\bibitem[Risheng et~al.(2021)Risheng, Long, Jiaao, Xin, and Zhongxuan]{RUAS}
Liu Risheng, Ma Long, Zhang Jiaao, Fan Xin, and Luo Zhongxuan.
\newblock Retinex-inspired unrolling with cooperative prior architecture search
  for low-light image enhancement.
\newblock In \emph{CVPR}, 2021.

\bibitem[Robert et~al.(2018)Robert, Thome, and Cord]{HybridNet}
Thomas Robert, Nicolas Thome, and Matthieu Cord.
\newblock Hybridnet: Classification and reconstruction cooperation for
  semi-supervised learning.
\newblock In \emph{ECCV}, 2018.

\bibitem[Saini and Narayanan(2018)]{sainiBMVC}
Saurabh Saini and P.~J. Narayanan.
\newblock Semantic priors for intrinsic image decomposition.
\newblock In \emph{BMVC}, 2018.

\bibitem[Saini and Narayanan(2019)]{sainiIJCV}
Saurabh Saini and P.~J. Narayanan.
\newblock Semantic hierarchical priors for intrinsic image decomposition.
\newblock \emph{ArXiv}, abs/1902.03830, 2019.

\bibitem[Saini and Narayanan(2023)]{QSEF}
Saurabh Saini and P.~J. Narayanan.
\newblock Quaternion factorized simulated exposure fusion.
\newblock In \emph{ACM ICVGIP}, 2023.

\bibitem[Saini et~al.(2016)Saini, Sakurikar, and Narayanan]{sainiFSIID}
Saurabh Saini, Parikshit Sakurikar, and P.~J. Narayanan.
\newblock Intrinsic image decomposition using focal stacks.
\newblock In \emph{ACM ICVGIP}, 2016.

\bibitem[Sharma and Tan(2021)]{Sharma2021NighttimeVE}
Aashish Sharma and Robby~T. Tan.
\newblock Nighttime visibility enhancement by increasing the dynamic range and
  suppression of light effects.
\newblock \emph{CVPR}, 2021.

\bibitem[Shekhar et~al.(2021)Shekhar, Reimann, Mayer, Semmo, Pasewaldt,
  Döllner, and Trapp]{sumitSpec}
Sumit Shekhar, Max Reimann, Maximilian Mayer, Amir Semmo, Sebastian Pasewaldt,
  Jürgen Döllner, and Matthias Trapp.
\newblock Interactive photo editing on smartphones via intrinsic decomposition.
\newblock \emph{Computer Graphics Forum}, 40\penalty0 (2), 2021.

\bibitem[{The GIMP Development Team}(2023)]{GIMP}
{The GIMP Development Team}.
\newblock Gimp, 2023.

\bibitem[Tominaga(1994)]{dichromaticRefModel_1994}
Shoji Tominaga.
\newblock Dichromatic reflection models for a variety of materials.
\newblock \emph{Color Research and Application}, 19, 1994.

\bibitem[Vonikakis(2007)]{VV}
Vassilios Vonikakis.
\newblock {Busting image enhancement and tone-mapping algorithms}.
\newblock \url{https://sites.google.com/site/vonikakis/datasets/}, 2007.
\newblock [Online; accessed 26-Oct-2023].

\bibitem[Wang et~al.(2020)Wang, Xie, Zhao, and Meng]{modelRainRemoval}
Hong Wang, Qi Xie, Qian Zhao, and Deyu Meng.
\newblock A model-driven deep neural network for single image rain removal.
\newblock 2020.

\bibitem[Wang et~al.(2013{\natexlab{a}})Wang, Zheng, Hu, and Li]{LOE}
Shuhang Wang, Jin Zheng, Hai-Miao Hu, and Bo Li.
\newblock Naturalness preserved enhancement algorithm for non-uniform
  illumination images.
\newblock \emph{IEEE TIP}, 22\penalty0 (9), 2013{\natexlab{a}}.

\bibitem[Wang et~al.(2013{\natexlab{b}})Wang, Zheng, Hu, and Li]{NPE}
Shuhang Wang, Jin Zheng, Hai-Miao Hu, and Bo Li.
\newblock Naturalness preserved enhancement algorithm for non-uniform
  illumination images.
\newblock \emph{IEEE TIP}, 22\penalty0 (9), 2013{\natexlab{b}}.

\bibitem[Wang et~al.(2004)Wang, Bovik, Sheikh, and Simoncelli]{SSIM}
Zhou Wang, A.~C. Bovik, H.~R. Sheikh, and E.~P. Simoncelli.
\newblock Image quality assessment: From error visibility to structural
  similarity.
\newblock \emph{IEEE TIP}, 13\penalty0 (4), 2004.

\bibitem[Wei et~al.(2018)Wei, Wang, Wenhan, and Liu]{retinexNet2018}
Chen Wei, Wenjing Wang, Yang Wenhan, and Jiaying Liu.
\newblock Deep retinex decomposition for low-light enhancement.
\newblock In \emph{BMVC}, 2018.

\bibitem[Wu et~al.(2022)Wu, Weng, Zhang, Wang, Yang, and Jiang]{UretinexNet}
Wenhui Wu, Jian Weng, Pingping Zhang, Xu Wang, Wenhan Yang, and Jianmin Jiang.
\newblock Uretinex-net: Retinex-based deep unfolding network for low-light
  image enhancement.
\newblock In \emph{CVPR}, 2022.

\bibitem[Xu et~al.(2020)Xu, Yang, Yin, and Lau]{freqLLE}
Ke Xu, Xin Yang, Baocai Yin, and Rynson~W.H. Lau.
\newblock Learning to restore low-light images via
  decomposition-and-enhancement.
\newblock In \emph{CVPR}, 2020.

\bibitem[Xu et~al.(2022)Xu, Wang, Fu, and Jia]{SNR}
Xiaogang Xu, Ruixing Wang, Chi-Wing Fu, and Jiaya Jia.
\newblock Snr-aware low-light image enhancement.
\newblock In \emph{CVPR}, 2022.

\bibitem[Yan et~al.(2020)Yan, Tan, and Dai]{UNIE}
Wending Yan, Robby~T Tan, and Dengxin Dai.
\newblock Nighttime defogging using high-low frequency decomposition and
  grayscale-color networks.
\newblock In \emph{ECCV}, 2020.

\bibitem[Yang et~al.(2023)Yang, Ding, Wu, Li, and Zhang]{NERCO}
Shuzhou Yang, Moxuan Ding, Yanmin Wu, Zihan Li, and Jian Zhang.
\newblock Implicit neural representation for cooperative low-light image
  enhancement.
\newblock In \emph{ICCV}, 2023.

\bibitem[Yang et~al.(2017)Yang, Tan, Feng, Liu, Guo, and Yan]{derainDataset}
Wenhan Yang, Robby~T. Tan, Jiashi Feng, Jiaying Liu, Zongming Guo, and
  Shuicheng Yan.
\newblock Deep joint rain detection and removal from a single image.
\newblock In \emph{2017 IEEE Conference on Computer Vision and Pattern
  Recognition (CVPR)}, pages 1685--1694, 2017.

\bibitem[Yang et~al.(2021{\natexlab{a}})Yang, Wang, Fang, Wang, and
  Liu]{BandNet}
Wenhan Yang, Shiqi Wang, Yapplicationsuming Fang, Yue Wang, and Jiaying Liu.
\newblock Band representation-based semi-supervised low-light image
  enhancement: Bridging the gap between signal fidelity and perceptual quality.
\newblock \emph{IEEE TIP}, 30, 2021{\natexlab{a}}.

\bibitem[Yang et~al.(2021{\natexlab{b}})Yang, Wang, Huang, Wang, and
  Liu]{LOLv2}
Wenhan Yang, Wenjing Wang, Haofeng Huang, Shiqi Wang, and Jiaying Liu.
\newblock Sparse gradient regularized deep retinex network for robust low-light
  image enhancement.
\newblock \emph{IEEE TIP}, 30, 2021{\natexlab{b}}.

\bibitem[Zhang et~al.(2021{\natexlab{a}})Zhang, Shao, Sun, Zhu, Gao, and
  Sang]{HEP}
Feng Zhang, Yuanjie Shao, Yishi Sun, Kai Zhu, Changxin Gao, and Nong Sang.
\newblock Unsupervised low-light image enhancement via histogram equalization
  prior.
\newblock \emph{arXiv:2112.01766}, 2021{\natexlab{a}}.

\bibitem[Zhang et~al.(2023)Zhang, Huang, Yao, Yang, Yu, Zhou, and Zhao]{IDR}
Jinghao Zhang, Jie Huang, Mingde Yao, Zizheng Yang, Huikang Yu, Man Zhou, and
  Fengmei Zhao.
\newblock Ingredient-oriented multi-degradation learning for image restoration.
\newblock \emph{CVPR}, 2023.

\bibitem[Zhang et~al.(2019{\natexlab{a}})Zhang, Zhang, Liu, Shen, Zhang, and
  Zhao]{ExcNet}
Lin Zhang, Lijun Zhang, Xinyu Liu, Ying Shen, Shaoming Zhang, and Shengjie
  Zhao.
\newblock Zero-shot restoration of back-lit images using deep internal
  learning.
\newblock \emph{ACM MM}, 2019{\natexlab{a}}.

\bibitem[Zhang et~al.(2018{\natexlab{a}})Zhang, Yuan, Xiao, Zhu, and
  Zheng]{zhang2018high}
Qing Zhang, Ganzhao Yuan, Chunxia Xiao, Lei Zhu, and Wei-Shi Zheng.
\newblock High-quality exposure correction of underexposed photos.
\newblock In \emph{ACM MM}, 2018{\natexlab{a}}.

\bibitem[Zhang et~al.(2019{\natexlab{b}})Zhang, Nie, and Zheng]{DUAL}
Qing Zhang, Yongwei Nie, and Weishi Zheng.
\newblock Dual illumination estimation for robust exposure correction.
\newblock \emph{Computer Graphics Forum}, 38, 2019{\natexlab{b}}.

\bibitem[Zhang et~al.(2018{\natexlab{b}})Zhang, Isola, Efros, Shechtman, and
  Wang]{LPIPS}
Richard Zhang, Phillip Isola, Alexei~A Efros, Eli Shechtman, and Oliver Wang.
\newblock The unreasonable effectiveness of deep features as a perceptual
  metric.
\newblock In \emph{CVPR}, 2018{\natexlab{b}}.

\bibitem[Zhang et~al.(2019{\natexlab{c}})Zhang, Zhang, and Guo]{kind}
Yonghua Zhang, Jiawan Zhang, and Xiaojie Guo.
\newblock Kindling the darkness: A practical low-light image enhancer.
\newblock In \emph{ACM MM}, 2019{\natexlab{c}}.

\bibitem[Zhang et~al.(2021{\natexlab{b}})Zhang, Guo, Ma, Liu, and
  Zhang]{kindPP}
Yonghua Zhang, Xiaojie Guo, Jiayi Ma, Wei Liu, and Jiawan Zhang.
\newblock Beyond brightening low-light images.
\newblock \emph{IJCV}, 129, 2021{\natexlab{b}}.

\bibitem[Zheng et~al.(2021)Zheng, Shi, and Shi]{adptativeUretinex}
Chuanjun Zheng, Daming Shi, and Wentian Shi.
\newblock Adaptive unfolding total variation network for low-light image
  enhancement.
\newblock \emph{ICCV}, 2021.

\bibitem[Zheng et~al.(2023)Zheng, Zhou, Dong, Rui, Huang, Li, and Zhao]{CUE}
Naishan Zheng, Man Zhou, Yanmeng Dong, Xiangyu Rui, Jie Huang, Chongyi Li, and
  Fengmei Zhao.
\newblock Empowering low-light image enhancer through customized learnable
  priors.
\newblock 2023.

\bibitem[Zheng and Gupta(2022)]{SGZ}
Shen Zheng and Gaurav Gupta.
\newblock Semantic-guided zero-shot learning for low-light image/video
  enhancement.
\newblock In \emph{WACV}, 2022.

\bibitem[Zhu et~al.(2020)Zhu, Zhang, Shen, Ma, Zhao, and Zhou]{RRDNet}
Anqi Zhu, Lin Zhang, Ying Shen, Yong Ma, Shengjie Zhao, and Yicong Zhou.
\newblock Zero-shot restoration of underexposed images via robust retinex
  decomposition.
\newblock \emph{ICME}, 2020.

\bibitem[Zhu et~al.(2022)Zhu, Xiao, Fang, Fu, Xiong, and Zha]{zhu2022efficient}
Yurui Zhu, Zeyu Xiao, Yanchi Fang, Xueyang Fu, Zhiwei Xiong, and Zheng-Jun Zha.
\newblock Efficient model-driven network for shadow removal.
\newblock \emph{AAAI}, 2022.

\end{thebibliography}
}
\clearpage
{\appendix
    \clearpage
\setcounter{page}{1}
\maketitlesupplementary
\label{sec:discussion}
\maketitle
Here we provide an elaborate discussion and additional evidences on various points mentioned in the main paper.

\Paragraph{Mathematical Derivation:}
The equations in the main paper (\cref{eqn:ADMM}) are derived directly following relevant sections from \citet{boydADMM,boydConvex,boydProximal}.
    
Initial sparsity objective \cref{eqn:sparsity} can be re-written in augmented Lagrangian form \cite{boydADMM} with $Y,\mu$ as auxiliary variables and $'$ implying matrix transpose as:
{\small 
\begin{equation*}
    \operatorname*{argmin}_{\mathbf{E,A}} \norm{\mathbf{A}}_* +\lambda \norm{\mathbf{E}}_1 + 
    \mathbf{Y}^{'} (\mathbf{A+E-X}) + \frac{\mu}{2} \norm{A+E-X}_2^2.
\end{equation*}
}
After using dual function and variable separation \cite{boydADMM,boydConvex}, we get the ADMM updates for $\mathbf{E, A}$ and $\mathbf{Y}$:
{\small
\begin{equation*}
\begin{aligned}
    \mathbf{E} &\leftarrow  \argmin_{\mathbf{E}} \big(\lambda\norm{E}_1 + \mathbf{Y}^{'} (\mathbf{A}+\mathbf{E}-\mathbf{X}) + \frac{\mu}{2} \norm{\mathbf{A} + \mathbf{E} - \mathbf{X} }^2_2 \big), \\
    \mathbf{A} &\leftarrow  \argmin_{\mathbf{A}} \big(\norm{A}_* + \mathbf{Y}^{'} (\mathbf{A}+\mathbf{E}-\mathbf{X}) + \frac{\mu}{2} \norm{\mathbf{A} + \mathbf{E} - \mathbf{X} }^2_2 \big), \\
    \mathbf{Y} &\leftarrow  \mathbf{Y} + \mu ( \mathbf{A} + \mathbf{E} - \mathbf{X}).
\end{aligned}
\end{equation*}
}
We can eliminate the second term by collecting $\mathbf{Y}$ inside the third square term:
{\small
\begin{equation*}
\begin{aligned}
    \mathbf{E} &\leftarrow \argmin_{\mathbf{E}} \big(\lambda\norm{E}_1 + \frac{\mu}{2} \norm{\mathbf{A} + \mathbf{E} - \mathbf{X} + Y/\mu}^2_2 \big), \\
    \mathbf{A} &\leftarrow \argmin_{\mathbf{A}} \big(\norm{A}_* + \frac{\mu}{2} \norm{\mathbf{A} + \mathbf{E} - \mathbf{X} + Y/\mu}^2_2 \big), \\
    \mathbf{Y} &\leftarrow \mathbf{Y} + \mu(\mathbf{A} + \mathbf{E} - \mathbf{X})
\end{aligned}
\end{equation*}
}
Note that additional $\mathbf{Y}$ terms in each update have no effect on the respective $\argmin$ solution. 
Now using the definition of the $p$-norm proximal operator \cite{boydProximal} given by:
{\small
\begin{equation*}
    \delta^p_{\mu}(v) := \argmin_x ( \norm{x}_p + \frac{\mu}{2} \norm{x-v}_2^2 ),
\end{equation*}
}
we can obtain the \cref{eqn:ADMM} in the main paper using $*$ and $L_1$ soft-thresholding.

\Paragraph{Unsupervised \vs Zero-reference LLE:}
Although similar, there is a crucial difference between the unsupervised and zero-reference LLE paradigms \cite{LLESurvey}.
As mentioned previously, unsupervised LLE solutions like \cite{jiang2021enlightengan,HEP,UNIE,CLIP-LIT,NERCO,PairLLE} require both poorly lit and well illuminated image sets for supervision though they need not be paired. 
On the other hand, zero-reference LLE solutions \cite{ExcNet,zeroDCE,zeroDCEPP,RUAS,SCI,PSENet,GDP} do not need any well-lit examples for training and purely use domain/task dependent loss terms and models for enhancement. In addition to making the methods more inexpensive, this also allows for better generalizability due to low domain dependence. Furthermore, due to explicitly encoded expert knowledge as domain priors, zero-reference solutions are smaller in size with simpler architectures and training curriculums than their unsupervised counterparts. This enables easy adoption of such techniques to other tasks as shown in the main paper. 
Although fair comparison is possible only between the methods of the same paradigm \cite{GDP}, still we report our comparison with various unsupervised solutions in \cref{tab:unsupQuant_supp}.
Note that our method beats several unsupervised LLE solutions and is competitive against the best two unsupervised solutions \cite{NERCO} and \cite{HEP}. \cite{NERCO} uses a complicated architecture comprising of pretrained multi-modal Large Language Models, multiple generator-discriminator pairs, implicit neural representation, collaborative mask attention modules \etc.
Relative to ours, this is significantly complex training process without direct interpretability/utility of intermediate results or possible extension to other enhancement tasks.
In our method, we have focused on encoding the fundamental aspects of the image formation process and represented it as in a recursive specularity factorization model.
Still our method surpasses \cite{NERCO} on 4 out of 6 and \cite{HEP} on 5 out of the 6 reported metrics individually.

\Paragraph{Interpretability:}
Being a model-driven unrolled network, our entire framework is easily interpretable as each optimization step is clearly represented. 
This allows direct user intervention and better analysis of the intermediate latent factors as done in \cref{fig:factExp}.
Here we repeat the same analysis with other parts of the shadow dataset \cite{CHUK}.
\cite{CHUK} dataset consists of manually marked dense shadow regions in images taken from several standard datasets. Specifically there are five categories of such images with test split size mentioned in the parenthesis: shadow\_ADE (226), shadow\_KITTI (555), shadow\_MAP (319), shadow\_USR (489) and shadow\_WEB (511).
The analysis using shadow\_ADE testset images was shown in the main paper.
Here we similarly plot the factor features over the background of shadow and non-shadow PCA reduced feature space, for other sets. 
For feature extraction we use pretrained DINOv2 vits\_14 backbone \cite{DINO} and factors were computed using direct optimization using \cref{eqn:sparsity}, \cref{eqn:ADMM} and \cref{eqn:shrinkage}. These plots are shown in \cref{fig:facAnalysis_supp}. Note how in each case, the extracted features from the factors lie sequentially over the background of shadow and highlight image regions starting from highlight regions for the first factor (indicating glares and specular regions) to complete shadow regions for the last factor (indicating complete dark pixels). The other illumination types are expected to lie in between the two extremes and can be observed from the graph to follow the same. This helps us interpret the extracted factors as approximations of illumination types at each pixel into glare, direct light, indirect light, soft shadow, hard shadows \etc.
\begin{figure}[t]
    \centering
    \includegraphics[width=\linewidth, height=6cm]{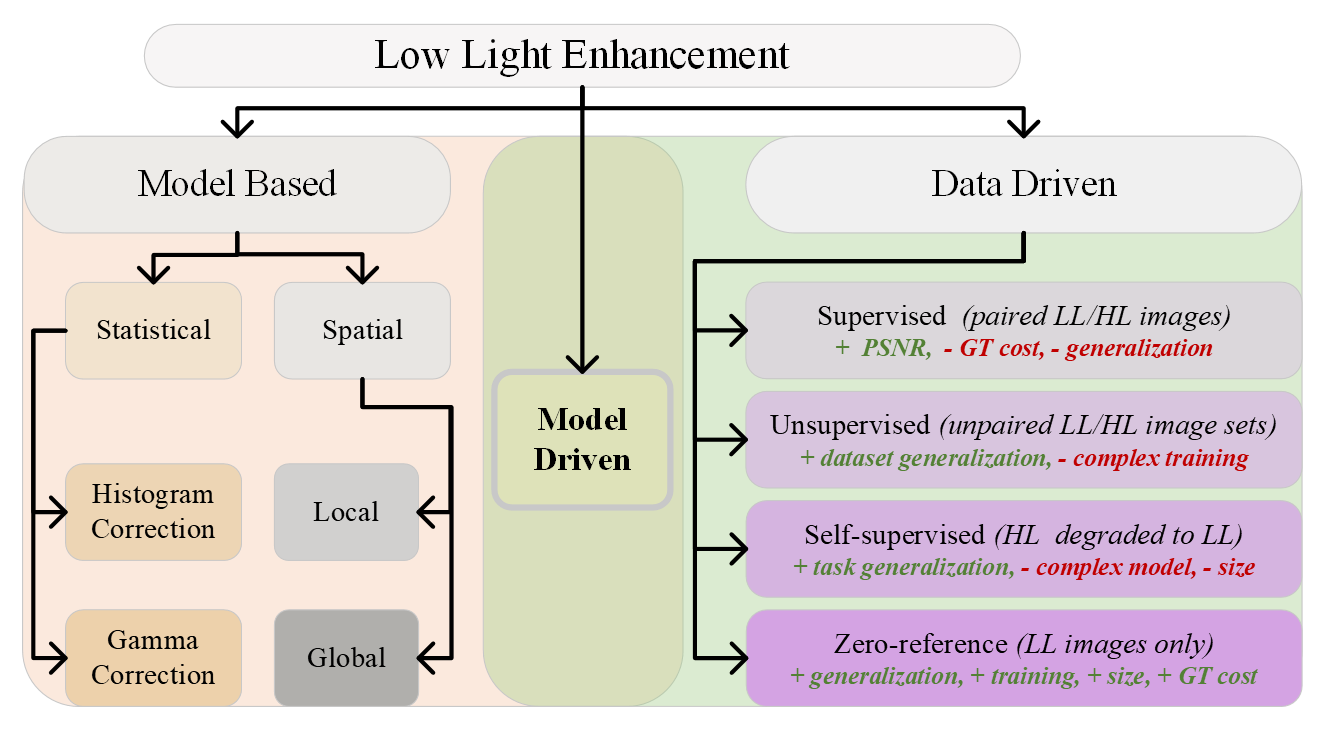}
    \caption{\textbf{LLE solutions categorization:} Data-driven methods are of mainly 4 types based on the type of input supervision available with each type having its \green{pros} and \red{cons} as listed above.}
    \label{fig:RWcategorization_supp}
\end{figure}

\begin{figure}[t]
    \centering
    \includegraphics[width=\linewidth, height=4.4cm]{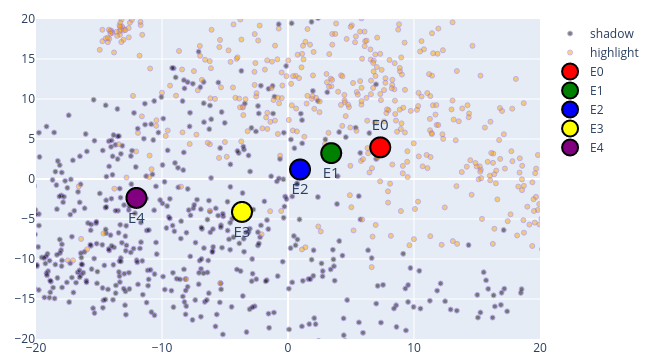} \\
    \includegraphics[width=\linewidth, height=4.4cm]{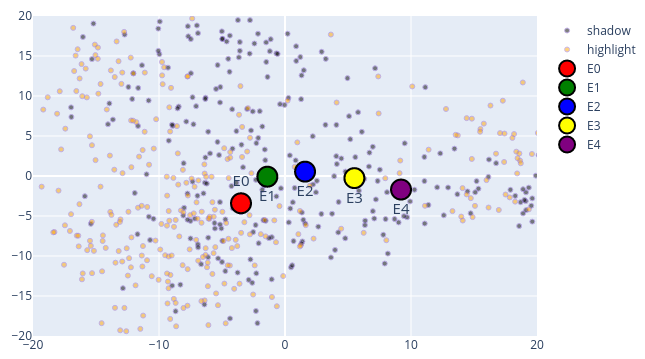} \\
    \includegraphics[width=\linewidth, height=4.4cm]{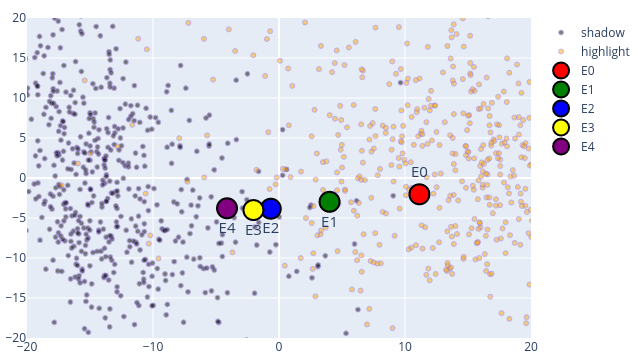} \\
    \includegraphics[width=\linewidth, height=4.4cm]{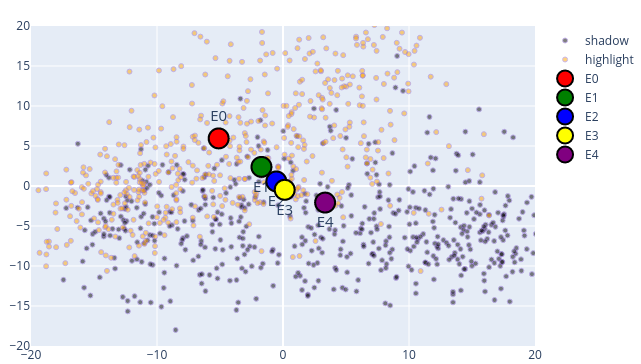} \\
    \caption{\textbf{Factor interpretability and analysis:} We perform factor distribution analysis \cref{fig:factExp} on four additional shadow datasets \cite{CHUK} (from top to bottom - shadow\_KITTI, shadow\_MAP, shadow\_USR and shadow\_WEB). Each plot represents features of shadow and non-shadow regions which forms the background and cluster centers of the five factors  feature distributions are plotted in the foreground. Note how in each case the series of factors is sequentially from bright to the dark region similar to \cref{fig:factExp} which provides more evidence to the validation done in \cref{sec:method}.}
    \label{fig:facAnalysis_supp}
\end{figure}


\Paragraph{Factorization Strategies:}
As mentioned in \cref{sec:intro,sec:relatedwork} and shown in \cref{tab:facRW}, various LLE solutions adopt different factorization strategies. We have provided a non-exhaustive list in the \cref{tab:facRW} but still others are possible.
The \textit{Frequency} strategy \cite{freqLLE} here refers to the low and high pass filtering of the input to extract coarse and fine image details, which are then processed separately.
On the other hand, \textit{spectral} strategy \cite{FEC} refers to decomposition into phase and amplitude using Fourier representation where phase is assumed to encode the entire structural information of the scene. 
\textit{Low rank} strategy based methods specifically exploit low rank structure of the reflectance component of the scene and are hence somewhat related to the Retinex division.
\cite{LR3M} focuses on hyper-spectral images, whereas \cite{QSEF} uses a complicated quaternion based robust PCA optimization strategy \cite{LRPCA} with no unrolled learning or generalization to other applications. 
\textit{Wavelets} and \textit{Multiscale} decompositions \cite{halfWave,afifiEC} build factors like image pyramids and can be considered to be an extension of the \textit{frequency} strategy. 
Decomposing input into extra glare or a shadow component \cite{shadingNet,Sharma2021NighttimeVE} along with the Retinex factorization has yielded better results and our method can be understood as the extreme case of such divisions. 
Similarities and differences with the often used \textit{intensity} based factorization strategy \cite{sef,sefIPOL} has already been discussed in the main paper.
Note that the global/local categorization here refers to whether the factors and the subsequent processing is limited to local image regions. 

\Paragraph{Training:} Training time of our RSFNet is quite fast. 
For any Lol dataset \cite{retinexNet2018,LOLv2,VELOL}, it takes approximately 30 minutes on a single 1080Ti GPU machine for the complete 50 epochs. 
We first train the factorization and fusion modules together for 25 epochs using \cref{eqn:L_fact} and then freeze the factorization parameters for next 25 epochs to train the fusion module with \cref{eqn:Loss}.
Initial versions of the system involved slow decay of factorization learning rate without abrupt freezing but the current setting was adopted to clearly ascertain the effect of each module training.
Hence we do not use any learning rate decay during our training but the reader is welcome to experiment with the same for their own datasets.

\Paragraph{Initialization:} During training instead of using any hard coded initialization value for thresholds, we allow per instance initialization.
Specifically, we use 0.9 ratio of learned threshold values and 0.1 fraction of the image mean for initialization with initial threshold values set to dataset mean.
This setting is also followed during inference and all the results reported in the main paper or supplementary are with this setting only.

Several optimization methods are sensitive to initialization conditions and when they are unrolled into model layers \cite{ALISTA,LRPCA}.
During implementation sources of randomness can be corrected by properly seeding the random number generators of the deep learning and the numerical algorithm libraries using:
\begin{quote}
\centering
   \texttt{np.random.seed($c$)} \\
    \texttt{torch.random.seed($c$)} \\
\end{quote}
where $c$ is some fixed integer constant. We use $c=2$ in our LLE experiments and the values of all the hyper-parameters will be provided with the final code in a config file.

\Paragraph{Testing:} For inference, we can edit the weights of the factors before concatenation and input into the fusion module to allow varying results. 
Although all results in the main paper are obtained without any weight manipulations (\ie all factors are equally important with each the weight vector corresponding to $E_0$ to $E_5$ set to [1,1,1,1,1,1]), better results are possible if dataset specific finetuning is allowed. 
If this is followed our scores on Lol-synthetic dataset in the main quantitative results table 
\cref{tab:quant_supp} can be updated to \cref{tab:weighted} by using $w=[1,4,4,4,4,4]$.
Yet another setting which can be configured is related to the bilateral filtering step which includes window size, color sigma and the spatial sigma in both of the horizontal directions. The values can be chosen based on the expected noise in the input datasets but we keep them constant as window size=5, color sigma=0.5 and spatial sigma=1 for all our experiments in \cref{tab:quant_supp}

\Paragraph{Datasets:} The details of five no-reference (\cref{tab:qual_quant}) and four Lol datasets \cref{tab:quant_supp} are given below:
\begin{itemize}
    \item Lolv1 \cite{retinexNet2018}: It contains 500 low light and well lit image pairs of real worl scenes with 485 for training and 15 for testing in the standard split. Each image is 400 $\times$ 600 in resolution with mean intensity = 0.05 (\ie very low light).
    \item Lolv2-real \cite{LOLv2}: It is an extension of Lolv1 dataset with 689 images in training and 100 in testing set. Mean intensity of images is 0.05 and resolution is same = 400 $\times$ 600. Note that majority of the images in the testing set of Lolv2 are present in the training set of Lolv1 and hence Lolv1 trained models should not be evaluated directly on Lolv2 testset.
    \item Lolv2-synthetic \cite{LOLv2}: As Lolv1 mostly contains only indoor scenes with heavy dark channel noise, Lolv2-synthetic presents a significant domain shift with mean intensity=0.2 and resolution= 384 $\times$ 384. The scenes are both indoors and outdoors and the supervision data is obtained by synthetically reducing the exposure by using the raw image data and natural image statistics.
    \item VE-Lol \cite{VELOL}: Vision Enhancement in LOw Level vision dataset (VE-LOL-L-Cap) consists of 1500 image pairs with 1400 \vs 100 training to test split. The trainset here consists of multiple under-exposed images of the same scene but the test set is similar to Lolv2-real. Dataset image resolution=400 $\times$ 600 and mean intensity=0.07. Multiple exposure settings here help ascertain model's robustness to input perturbations.\\
    \newline
 Other five datasets \cite{LLESurvey} are no-reference (\ie without any ground truth well lit image) and are used for perceptual quality evaluation and generalization assessment. Although varying number of images have been reported in the previous literature for a few of these datasets \cite{zeroDCE,afifiEC,LLESurvey}, we use the download links provided by \citet{LLESurvey} with the following brief description of each dataset:
    \item DICM \cite{DICM}: 69 images, mean=0.32, mixed exposure settings, variable resolutions, real scenes, varying scene including macros, landscapes, indoors, outdoors \etc.
    \item LIME \cite{LIME}: 10 images, mean=0.15, varying resolutions, real scenes, varying scene types.
    \item MEF \cite{MEF}: 17 images, mean=0.15, resolution=512 $\times$ 340, relatively darker images, varying scene types.
    \item NPE \cite{NPE}: 85 images, mean=0.31, varying resolution, both over and under exposed image regions, mostly outdoor scenes.
    \item VV \cite{VV}: 24 images, mean=0.26, resolution=2304 $\times$ 1728, large images, both over and under exposed image regions, both indoor/outdoor scene types.
\end{itemize}
These results are listed in \cref{tab:quant} \cref{tab:qual_quant} and \cref{tab:quant_supp}. As can be observed in the tables, our method achieves best score over all with best or second best performance on several benchmarks across multiple metrics.
\begin{table*}[t]
\small
\centering
    \begin{tabular}{@{}L{20mm}|C{15mm}C{15mm}C{15mm}C{15mm}R{15mm}R{15mm}@{}}
    \toprule
    \toprule
    $Type$ & $\text{PSNR}_y \uparrow$ & $\text{SSIM}_y \uparrow$ & $\text{PSNR}_c \uparrow$ & $\text{SSIM}_c \uparrow$ & $\text{NIQE} \downarrow$  & $\text{LPIPS} \downarrow$ \\
    \midrule
    $w/o$ weights  & 19.73   & 0.843  & 19.39  & 0.745 & 3.701 & 0.278 \\
    weighted       & 20.22   & 0.884  & 17.23  & 0.815 & 4.286 & 0.159 \\
    \bottomrule
    \bottomrule
    \end{tabular}
\caption{\textbf{Factor Weights:} Our updated results on Lol-synthetic dataset \cite{LOLv2} if we additionally allow the user to configure factor weights before concatenation and input to the fusion module. To be understood in the wider context of \cref{tab:quant} and \cref{tab:quant_supp}.}
\label{tab:weighted}
\end{table*}
\begin{table*}[t]
\centering
\small
    \begin{tabular}{@{}L{20mm}||C{15mm}C{15mm}|C{15mm}C{15mm}|| R{40mm}@{}}
    \toprule
    \toprule
       Configuration        & \multicolumn{2}{c|}{Factorization} & \multicolumn{2}{c||}{Fusion} & Experiment\\
                     & Trad.        & Deep      & Trad.         & Deep      &       \\
        \midrule
       $C_{11}$      &              & \ding{51} &               & \ding{51} &  RSFNet LLE \cref{fig:blockDiagram}     \\
       $C_{10}$      &              & \ding{51} & \ding{51}     &           &  Ablation ($w/o$ Fusion) \cref{tab:abla}     \\
       $C_{01}$      &  \ding{51}   &           &               & \ding{51} &  Extension Apps. \cref{fig:apps}     \\
       $C_{00}$      &  \ding{51}   &           & \ding{51}     &           &  User Apps. \cref{fig:userApps}     \\
    \bottomrule
    \bottomrule
    \end{tabular}
\caption{\textbf{System Configurations:} Various possible configurations of our proposed technique. Two central steps of our method, factorization and fusion, could each be either traditionally estimated with manual model-based optimization or using deep data-driven methods. This gives rises to four possible configurations all of which are used in one or the other experiment in the main paper}
\label{tab:configs}
\end{table*}

\Paragraph{Metrics:} Most frequently reported metric for LLE task is PSNR (Peak Signal to Noise Ratio). 
Although traditional usage of PSNR has been for denoising of grayscale images with only single channel but now it also has been extended to multichannel scenarios for various tasks. 
PSNR for a predicted enhanced output $\hat{y}$ is given as:
\begin{equation}
\label{eqn:psnr}
\small
    p = 10 \log \Bigg[\frac{ \frac{1}{N}\sum \limits_{i}^{N} (\hat{y}_i-y_i )^2}{M^2} \Bigg],
\end{equation}
where $N$ is total number of pixels and $M$ is the peak pixel value which depending upon the situation is either $1.0$ or $255$. 
\cref{eqn:psnr} is straightforward in case of single channel image but there is slight ambiguity in case of multichannel prediction.
Different results are obtained depending upon whether per channel mean is considered inside the logarithm or outside.
Correct way of multichannel PSNR definition is to consider it inside the logarithm \ie to take mean square error over all the channels simultaneously instead of individually and then averaging it as shown below:
\begin{equation}
\label{eqn:psnr_full}
\small
    p = 10 \log \Bigg[\frac{ \frac{1}{N*C} \sum \limits_{c}{} \sum \limits_{i}^{N} (\hat{y}_{i,c}-y_{i,c} )^2}{M^2} \Bigg].
\end{equation}

Yet another issue is during the YCbCr to rgb conversion for PSNR evaluation of Y only channel. Most of the codes directly use the in-built functions from the available libraries like opencv or PIL. The conversion involves applications of a transformation matrix which differs from library to library depending upon whether the input signal is assumed to be analog or digital \eg opencv applies the following transformation assuming analog input:
\begin{equation}
\label{eqn:rgb2ycbcr_opencv}
Y \leftarrow 0.299 \cdot R + 0.587 \cdot G + 0.114 \cdot B 
\end{equation}
whereas Matlab prefers the digital transformation as:
\begin{equation}
\label{eqn:rgb2ycbcr_matlab}
Y \leftarrow 0.2568 \cdot R + 0.5041 \cdot G + 0.0979 \cdot B 
\end{equation}
This leads to variability in results (approximately 1 PSNR difference) depending upon the conversion library chosen. In our opinion \cref{eqn:rgb2ycbcr_matlab} should be chosen and the PSNR tables should clearly highlight that it is a single Y channel evaluation.

\Paragraph{Configurations:} Our proposed method can be used for various applications in one of four possible configurations as shown in \cref{tab:configs}. 
This is dependent on whether the factorization and fusion steps are carried out via traditional model-based optimization or learned using data-driven deep networks. 
Model-based solutions are better generalizable but slower with lesser performance than data-driven solutions. 
In our main paper we have used all four configurations in one or the other experiment as listed in the \cref{tab:configs}.
For traditional factorization we use solution to the direct specularity estimation optimization equation \cref{eqn:sparsity} using \cref{eqn:ADMM}, whereas for deep solution we use the unrolled layers \cref{fig:blockDiagram} to learn the associated optimization thresholds using our Factorization Modules which are learned form the dataset in a data-driven fashion. Fusion is either task specific deep network or simply the running average as described in \cref{eqn:factOnly}.
This highlights the flexibility and versatile nature of our proposed technique which allows easy integration with pre-existing fusion methods with observed improvement in all scenarios.
\begin{figure*}[t]
    \centering
    \includegraphics[width=0.9\linewidth, height=5.8cm]{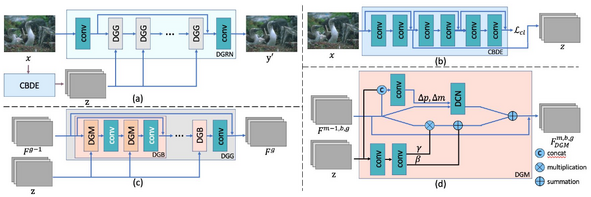}
    \caption{\textbf{AirNet:} (a) Block diagram from \cite{AirNet}. CBDE (b) refers to Contrastive-Based Degradation Encoder, DGG (c) means Degradation Guided Groups and DGM (d) is Degradation Guided Module. For complete details refers to \cite{AirNet}. For our usage, we alter first conv layer (first deep blue block on top-left (a)) and the first conv layer in CBDE (first deep blue block on top-right (b)).}
    \label{fig:AirNet_supp}
\end{figure*}
 \;\\
\begin{table*}[t]
    \centering
    \small
    \begin{tabular}{@{}L{35mm} || C{12mm}C{12mm} | C{12mm}C{12mm} | C{12mm}C{12mm} || C{12mm}C{12mm}@{}}
    \toprule
    \toprule
        \textbf{TASK} $\xrightarrow{}$ & \multicolumn{2}{c|}{\textbf{DEHAZE} \cite{dehazeDataset}} & \multicolumn{2}{c|}{\textbf{DERAIN} \cite{derainDataset}} & \multicolumn{2}{c||}{\textbf{DEBLUR} \cite{deblurDataset}} & \multicolumn{2}{c}{\textbf{Mean}} \\
    \midrule
        Method & $\text{PSNR}$  & $\text{SSIM}$  & $\text{PSNR}$  & $\text{SSIM}$  & $\text{PSNR}$  & $\text{SSIM}$ & $\text{PSNR}$  & $\text{SSIM}$  \\
    \midrule
        DL \cite{multitaskDL} & 20.54 & 0.826 & 21.96 & 0.762 & 19.86 & 0.672 & 20.78 & 0.753\\
        TransWeather \cite{TransWeather} & 21.32 & 0.885 & 29.43 & 0.905 & 25.12 & 0.757 & 25.29 & 0.849 \\
        TAPE \cite{TAPE} & 22.16 & 0.861 & 29.67 & 0.904 & 24.47 & 0.763 & 25.43 & 0.843 \\
    \midrule
        AirNet \cite{AirNet} (multi-task) & 21.04 & 0.884 & 32.98 & 0.951 & 24.35 & 0.781 & 26.12 & 0.872\\
        AirNet \cite{AirNet} (uni-task) & 23.18 & 0.900 & 34.90 & 0.966 & 26.42 & 0.801  & 28.17 & 0.889 \\
    \textbf{AirNet \cite{AirNet} + Ours} & \textbf{24.96} & \textbf{0.929} & \textbf{36.19} & \textbf{0.972} & \textbf{27.29} & \textbf{0.827} & \textbf{29.48} & \textbf{0.909} \\
    \green{\textbf{\% Improvement}} & \green{\textbf{$+7.68$}} & \green{\textbf{$+3.22$}} & \green{\textbf{$+3.70$}} & \green{\textbf{$+0.60$}} & \green{\textbf{$+3.29$}} & \green{\textbf{$+3.25$}} & \green{\textbf{$+4.65$}} & \green{\textbf{$+2.25$}} \\
    \bottomrule
    \bottomrule
    \end{tabular}
    \caption{\textbf{Prior Induction:} Our factors can induce structure prior in an existing base model \cite{AirNet} and improve performance for multiple enhancement tasks. Here we show extension of \cref{tab:app} in the main paper in the wider context of similar methods.}
    \label{tab:app_supp}
\end{table*}
\begin{table*}[h!]
    \centering
    \small
    \begin{tabular}{@{}p{20mm} || C{20mm}C{20mm} | C{20mm}C{20mm} || C{30mm}@{}}
    \toprule
    \toprule
        NIQE \quad $\downarrow$ & SNR \cite{SNR} & RFormer \cite{RFormer} & HEP \cite{HEP} & NeRCo \cite{NERCO} & RSFNet (Ours) \\
    \midrule
        DICM \cite{DICM}    & 3.622     &  3.076   &   4.064   &   3.553   &   3.230 \\
        LIME \cite{LIME}    & 3.752     &   3.910   &   3.981   &   3.422   &   3.800 \\    
        MEF  \cite{MEF}     & 3.917     &   3.135   &   3.648   &   3.152   &   3.000  \\
        NPE  \cite{NPE}     & 3.535     &   3.63*   &   2.986   &   3.241   &   3.310  \\
        VV   \cite{VV}      & 2.887     &   2.183   &   3.596   &   3.169   &  1.960   \\
    \midrule
        \textbf{Mean}       & 3.543     &   \underline{3.187}   &   3.655   &   3.307   &   \textbf{3.060} \\
    \bottomrule
    \bottomrule
    \end{tabular}
    \caption{\textbf{Generalized Performance:} Performance generalization comparison (\cref{tab:qual_quant} extension) of best ranking (\cref{tab:unsupQuant_supp}) two supervised LLE solutions (first two columns: SNR \cite{SNR}, RFormer \cite{RFormer}) and two unsupervised LLE solutions (last two columns: HEP \cite{HEP}, NeRCo \cite{NERCO}) \vs our zero-reference RSFNet method on five no-reference benchmarks namely: DICM \cite{DICM}, LIME \cite{LIME}, MEF \cite{MEF}, NPE \cite{NPE} and VV \cite{VV}. Our method is able to generalize better to unseen data compared to others as observed from the overall lowest NIQE scores \cite{NIQE} in the last row. (SNR, HEP and NeRCo results computed using provided pretrained weights with Lolsyn checkpoint where ever applicable and all images resized to 512x512 before processing to avoid dataloader errors. For RFormer, results downloaded from their official homepage. * refers to the incomplete NPE dataset results as available).}
    \label{tab:uslQual_supp}
\end{table*}
\Paragraph{Extensions:} In order to show the utility of our factors beyond the LLE task, we have shown the advantage of using them along with the pre-existing multi-task enhancement networks. Specifically, we use AirNet \cite{AirNet} (\cref{fig:AirNet_supp}) and alter the input tensor from a single 3 channel input to a tensor comprising of the concatenated input image and other factors by simply modifying the in-channels of the first convolutional layer in both the main AirNet backbone and the CBDE module. 
We train for 500 epochs for each task separately (with additional 50 epochs for initial warmup) and keep the default learning rate and decay parameters.
We found no significant difference in training from scratch or finetuning over the multi-task pre-trained checkpoint.
We also provide the extension of \cref{tab:app} in \cref{tab:app_supp} as the full comparison table using the values as provided by \cite{IDR} for various tasks in the multitask configuration. For uni-task configuration (\ie one task at a time), we report the values as provided in the main AirNet paper itself or compute them ourselves by retraining with default parameters (for deblurring). Note that the we have chosen AirNet over others due to its overall better performance than others (except IDR). IDR \cite{IDR} was not used as the public code is not available at the time of writing of this paper. 
As can be observed from the table, even straightforward introduction of our factors as priors without any loss or major architecture modifications can improve the existing performance consistently for all reported tasks.

\Paragraph{Visualizations:} We provide several visualizations of our results mentioned in the main paper. Specifically, we provide the following:
\begin{itemize}
    \item Visualization of our five extracted specular factors for the shadow\_ADE dataset \cite{CHUK} in \cref{fig:factViz_0_supp}.
    \item Visualization of our five extracted specular factors for the IIW dataset \cite{IIW} in \cref{fig:factViz_1_supp}.
    \item Visualization of our five extracted specular factors for extension applications using deraining \cite{derainDataset}, dehazing \cite{dehazeDataset} and deblurring datasets \cite{deblurDataset} in \cref{fig:factViz_2_supp}.
    \item Our qualitative results on low light image benchmarks in \cref{fig:lleRes_supp}.
    \item Qualitative comparison of our results with other zero-reference LLE solutions in \cref{fig:lleComp_supp}.
    \item Our results for the deraining application on the Rain100L dataset \cite{derainDataset} in \cref{fig:derain_supp}.
    \item Our results for the dehazing application on the RESIDE SOTS outdoor dataset \cite{dehazeDataset} in \cref{fig:dehaze_supp}.
    \item Our results for the deblurring application on the GoPro dataset \cite{deblurDataset} in \cref{fig:deblur_supp}.
    \item High resolution versions of the user controlled edited images (\cref{fig:userApps}) in GIMP \cite{GIMP} in \cref{fig:userApps_supp}.
    \item Extended quantitative comparison scores with contemporary traditional and zero-reference solutions (extension of \cref{tab:quant}) in \cref{tab:quant_supp}. 
    \item Quantitative comparison of our method with contemporary unsupervised LLE solutions on three Lol benchmarks in \cref{tab:unsupQuant_supp}.
\end{itemize}
\Paragraph{Generalization:}
Additionally, we also provide generalization performance comparison of various LLE solutions, including recent supervised and unsupervised methods, on the unseen data using images from standard no-reference LLE benchmarks (\ie without any ground truth) in \cref{tab:uslQual_supp} . We report NIQE scores \cite{NIQE} to assess the overall perceptual quality and the naturalness of the generated results. As can be seen from the \cref{tab:uslQual_supp}, our method, being a zero-reference solution, generalizes better due to low dependence on the training dataset compared to the supervised and the unsupervised counterparts. This generalization across unseen datasets, along with generalization to other applications like deraining, dehazing \etc, proves the advantage of zero-reference methods over other types of solutions.
\begin{figure*}[t]
    \centering
    \begin{tabular}{c c}
    \includegraphics[width=0.5\linewidth, height=6cm]{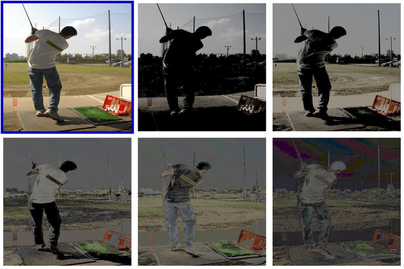}   &
    \includegraphics[width=0.5\linewidth, height=6cm]{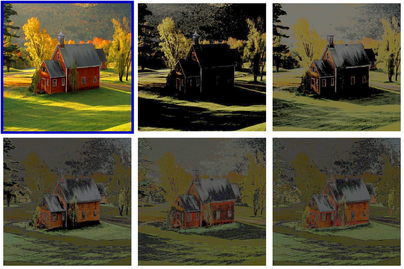}  \\
        & \\
    \includegraphics[width=0.5\linewidth, height=6cm]{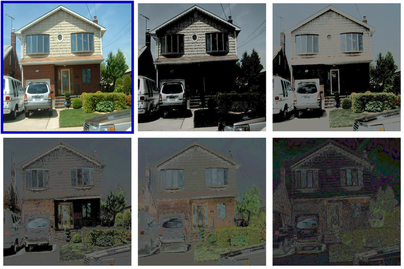}   &
    \includegraphics[width=0.5\linewidth, height=6cm]{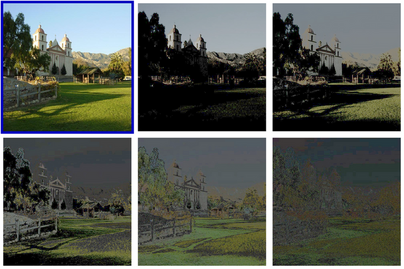}  \\
        & \\
    \includegraphics[width=0.5\linewidth, height=6cm]{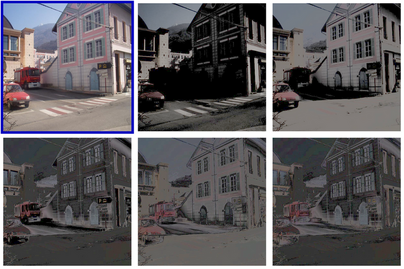}   &
    \includegraphics[width=0.5\linewidth, height=6cm]{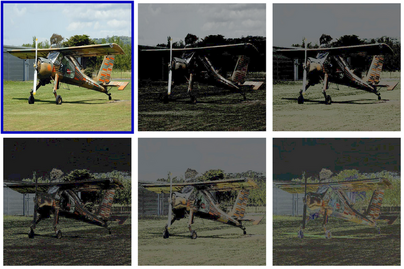}  \\
    \end{tabular}
    \caption{\textbf{Factor Visualizations (outdoors):} We show visualizations of our extracted five specular factors for various scenes. Input images (blue box) are taken from \cite{CHUK} dataset and factors are rescaled for visualization. Note how various regions are captured in the respective factors depending upon whether they are illuminated by directly, indirectly or in shadows.}
    \label{fig:factViz_0_supp}
\end{figure*}

\begin{figure*}[t]
    \centering
    \includegraphics[width=\linewidth, height=3cm]{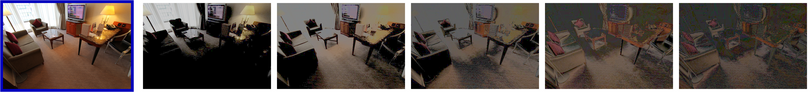} \\
    \includegraphics[width=\linewidth, height=3cm]{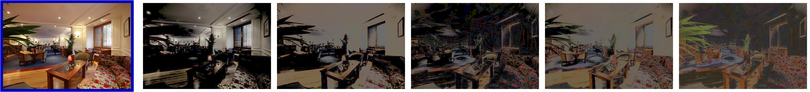} \\
    \includegraphics[width=\linewidth, height=3cm]{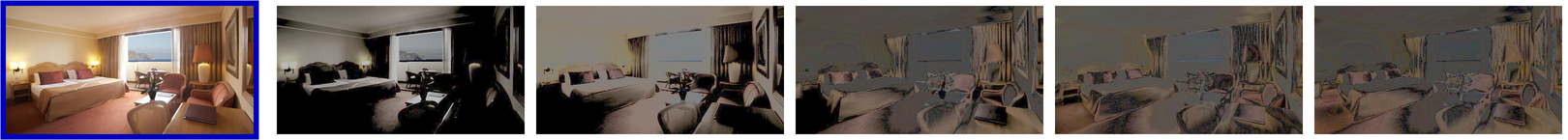} \\
    \includegraphics[width=\linewidth, height=3cm]{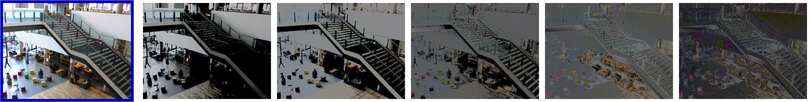} \\
    \includegraphics[width=\linewidth, height=3cm]{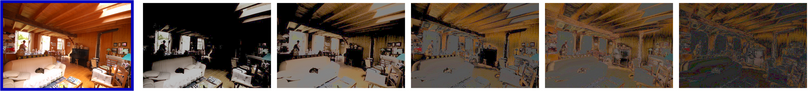} \\
    \includegraphics[width=\linewidth, height=3cm]{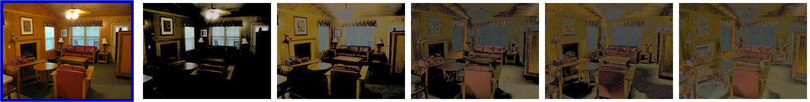} \\
    \includegraphics[width=\linewidth, height=3cm]{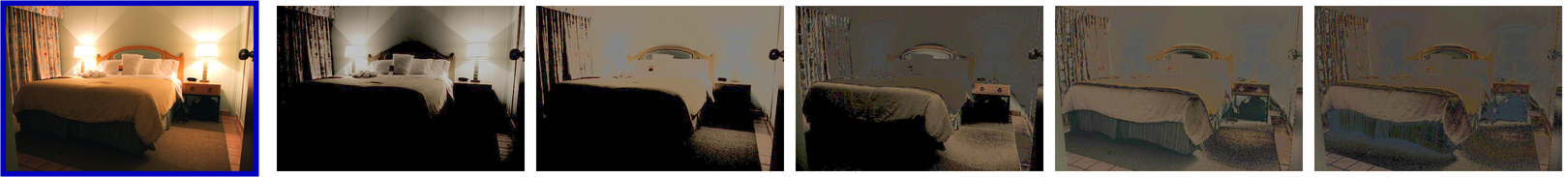} \\
    \caption{\textbf{Factor Visualizations (indoors):} We show visualizations of our extracted five specular factors for various scenes. Input images (blue box) are taken from \cite{IIW} dataset and factors are rescaled for visualization. Note how various regions are captured in the respective factors depending upon whether they are illuminated by directly, indirectly or in shadows.}
    \label{fig:factViz_1_supp}
\end{figure*}

\begin{figure*}[t]
    \centering
    \begin{tabular}{c}
    \includegraphics[width=\linewidth, height=3cm]{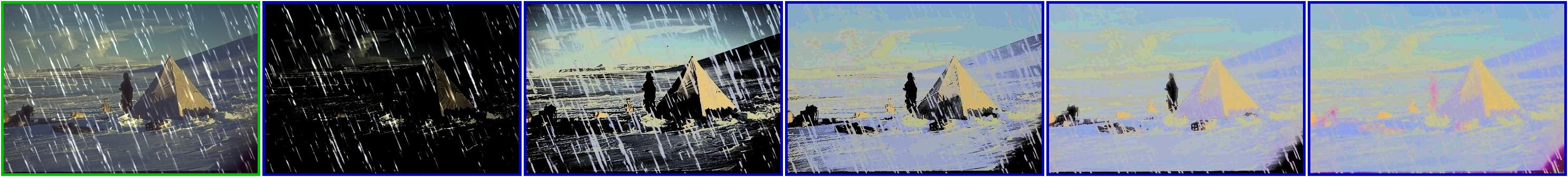}  \\
    \includegraphics[width=\linewidth, height=3cm]{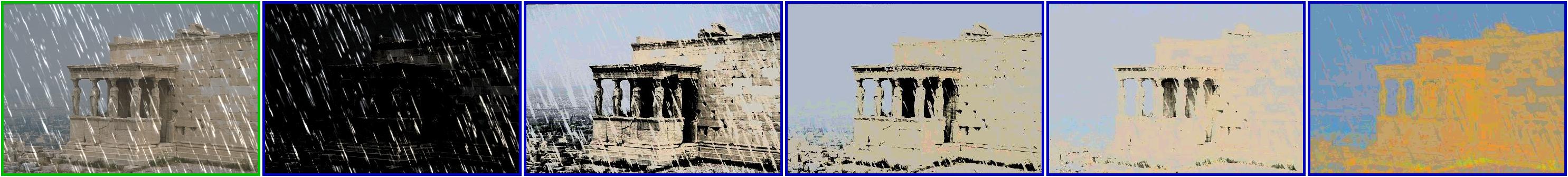}  \\
    \; \\
    \includegraphics[width=\linewidth, height=3cm]{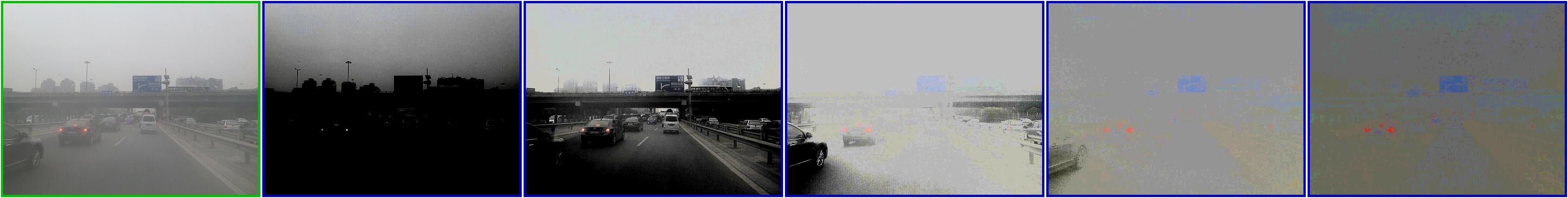}   \\
    \includegraphics[width=\linewidth, height=3cm]{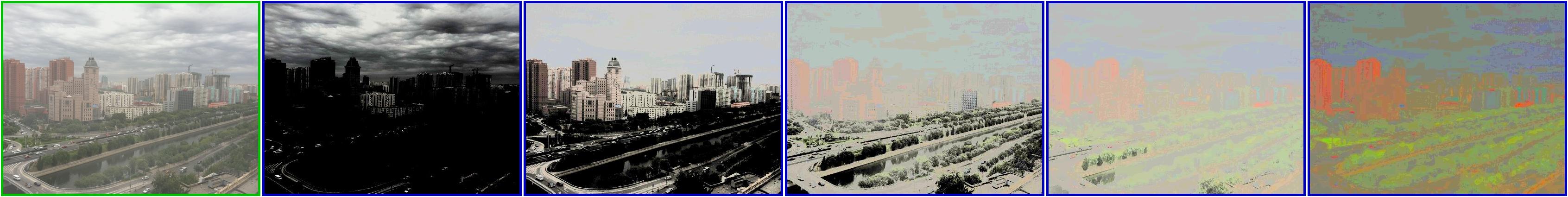}  \\
    \; \\
    \includegraphics[width=\linewidth, height=3cm]{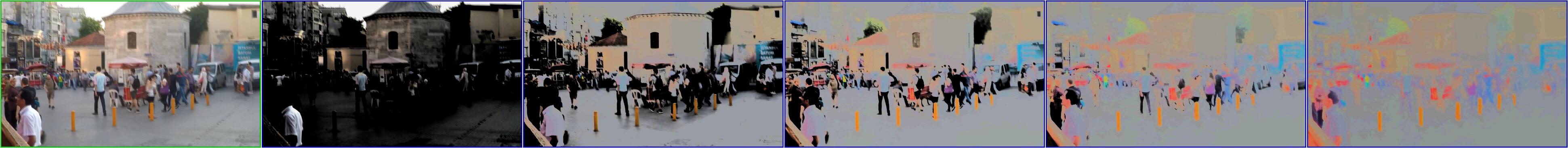}   \\
    \includegraphics[width=\linewidth, height=3cm]{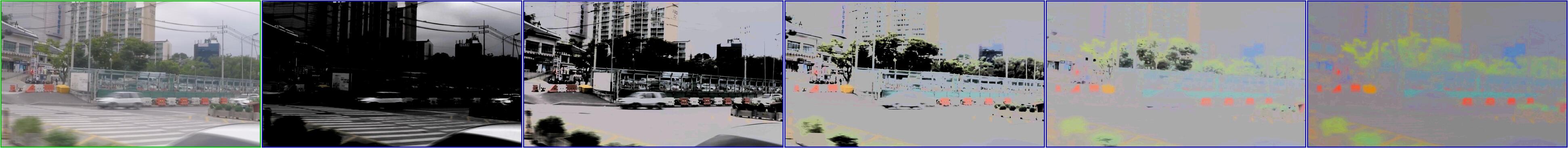}  \\
    \end{tabular}
    \caption{\textbf{Factor Visualizations (extensions):} We show visualizations of our extracted five specular factors for various scenes with different degradations. Input images (green box) are taken from 3 degraded images datasets \cite{dehazeDataset,derainDataset, deblurDataset} and factors (blue boxes) are rescaled for visualization. Note how specific degradation gets highlighted in different factors depending on the scene and the type of degradation.}
    \label{fig:factViz_2_supp}
\end{figure*}

\begin{figure*}[t]
    \centering
    \addtolength{\tabcolsep}{-0.4em}
    \begin{tabular}{ccc}
    \centering
    \includegraphics[width=0.5\linewidth, height=2.6cm]{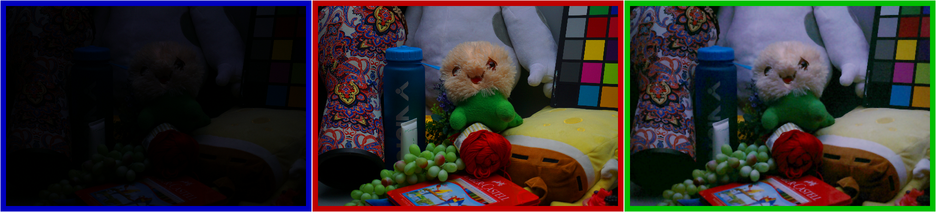} &
    \qquad &
    \includegraphics[width=0.5\linewidth, height=2.6cm]{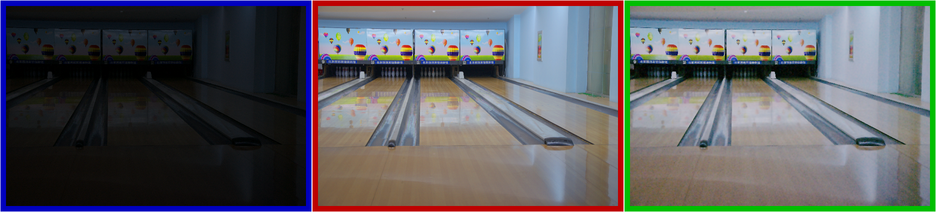} \\
    \includegraphics[width=0.5\linewidth, height=2.6cm]{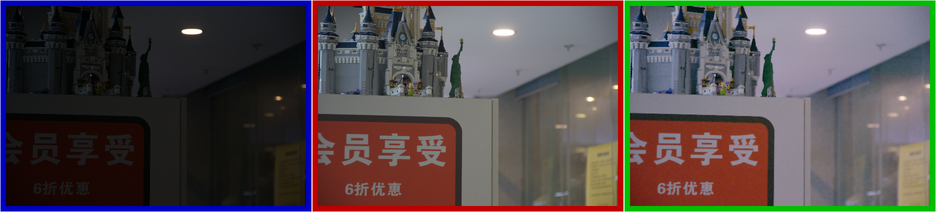} &
    \qquad &
    \includegraphics[width=0.5\linewidth, height=2.6cm]{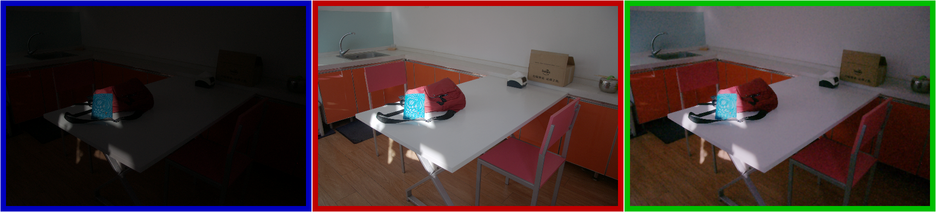} \\
    \includegraphics[width=0.5\linewidth, height=2.6cm]{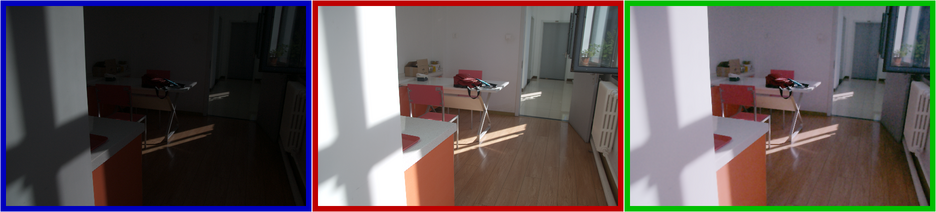} &
    \qquad &
    \includegraphics[width=0.5\linewidth, height=2.6cm]{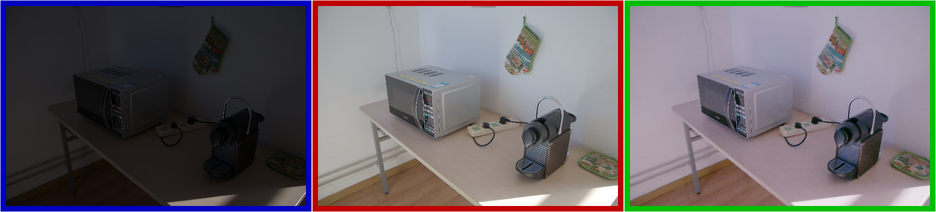} \\
    \includegraphics[width=0.5\linewidth, height=2.6cm]{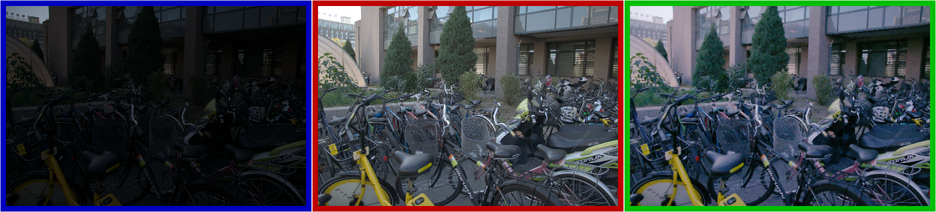} &
    \qquad &
    \includegraphics[width=0.5\linewidth, height=2.6cm]{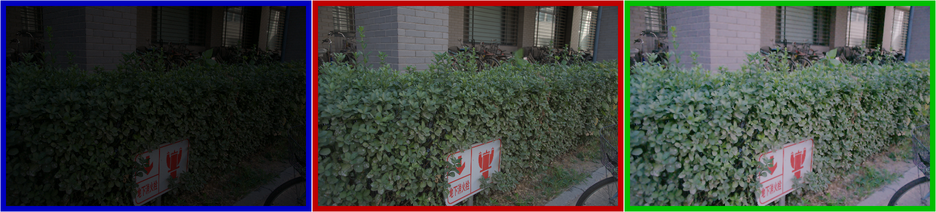} \\
    \includegraphics[width=0.5\linewidth, height=2.6cm]{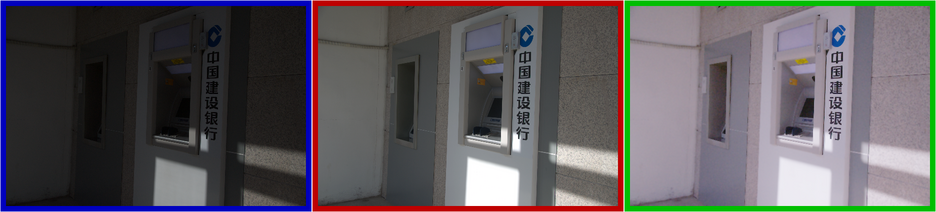} &
    \qquad &
    \includegraphics[width=0.5\linewidth, height=2.6cm]{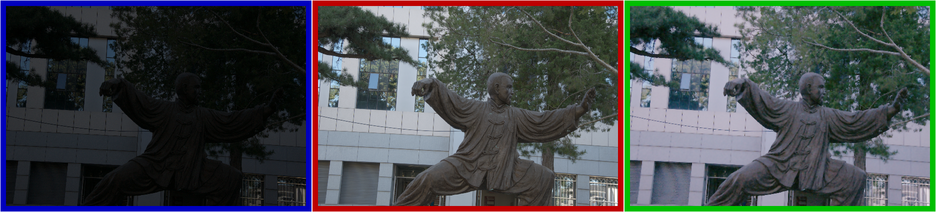} \\
    \includegraphics[width=0.5\linewidth, height=2.6cm]{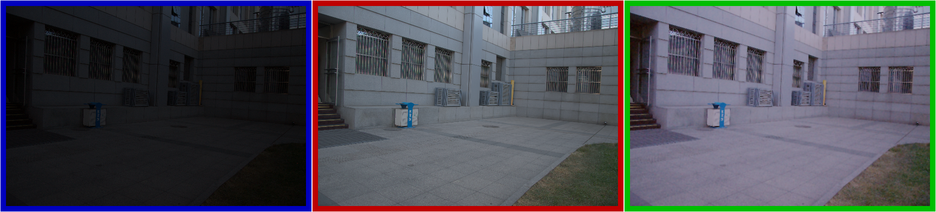} &
    \qquad &
    \includegraphics[width=0.5\linewidth, height=2.6cm]{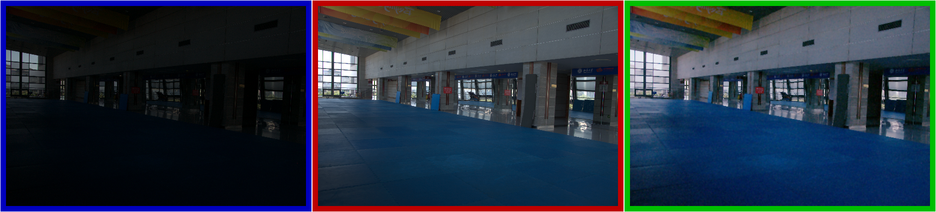} \\
    \includegraphics[width=0.5\linewidth, height=2.6cm]{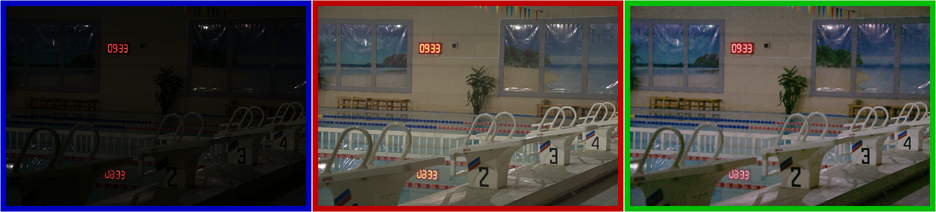} &
    \qquad &
    \includegraphics[width=0.5\linewidth, height=2.6cm]{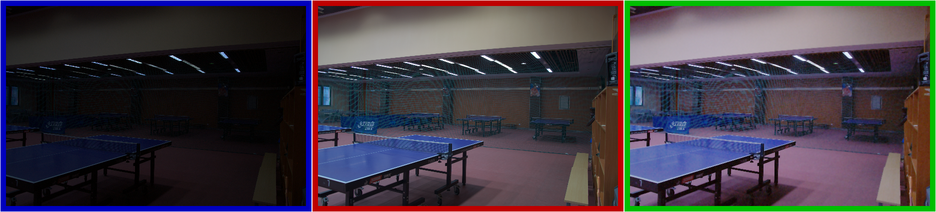} \\
    \includegraphics[width=0.5\linewidth, height=2.6cm]{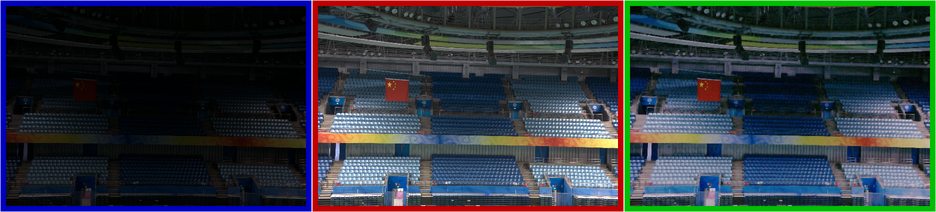} &
    \qquad &
    \includegraphics[width=0.5\linewidth, height=2.6cm]{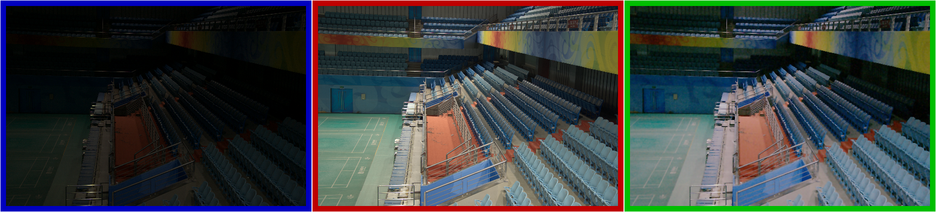} \\
    \end{tabular}
    \caption{\textbf{Our LLE Results}: Additional low light enhancement results from multiple Lol-x datasets \cite{retinexNet2018, LOLv2}. Each set contains input image (blue box), ground truth (red box) and our result (green box).}
    \label{fig:lleRes_supp}
\end{figure*}

\begin{figure*}[t]
    \centering
    \addtolength{\tabcolsep}{-0.4em}
    \begin{tabular}{ccc}
    \centering
    \includegraphics[width=0.5\linewidth, height=7cm]{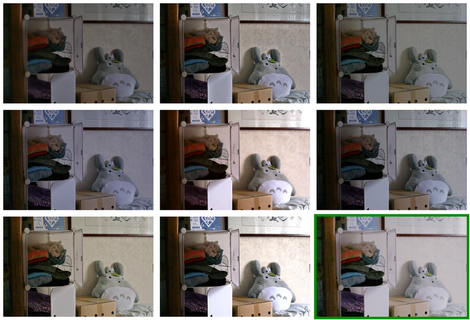} &
    \qquad &
    \includegraphics[width=0.5\linewidth, height=7cm]{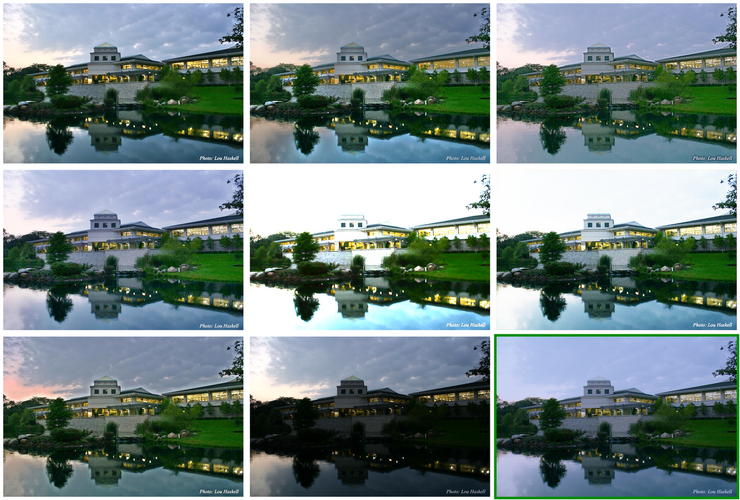} \\
    
    \includegraphics[width=0.5\linewidth, height=7cm]{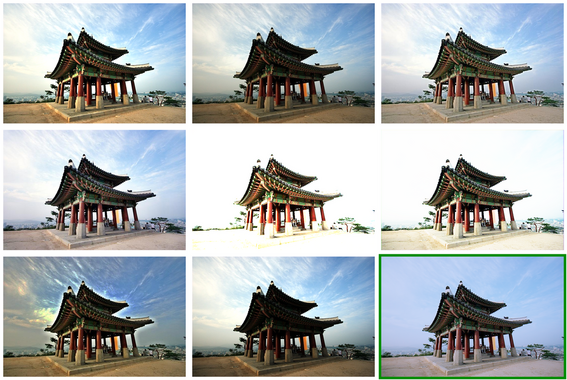} &
    \qquad &
    \includegraphics[width=0.5\linewidth, height=7cm]{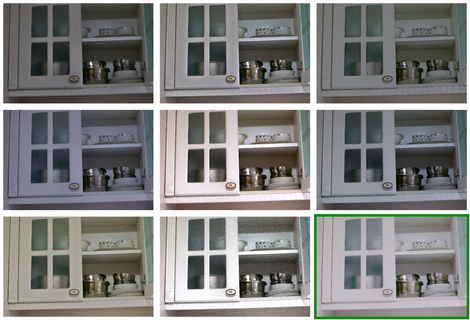} \\
    
    \includegraphics[width=0.5\linewidth, height=7cm]{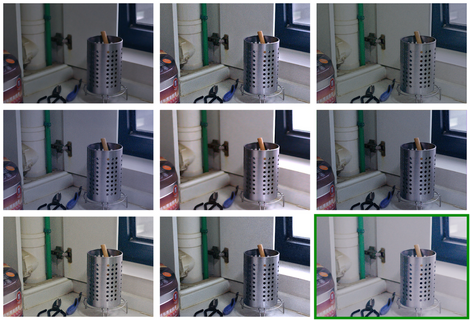} &
    \qquad &
    \includegraphics[width=0.5\linewidth, height=7cm]{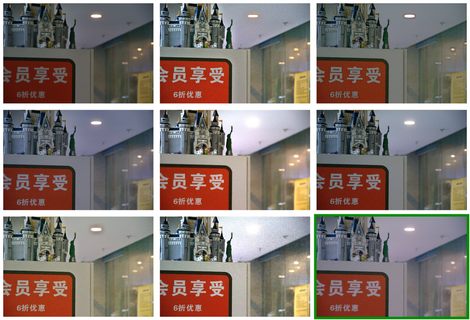} \\
     
    \end{tabular}
    \caption{\textbf{Qualitative Comparisons}: Additional low light enhancement comparisons (\cref{fig:qual} extension). Each set row in the grid contains results from: [SDD\cite{SDD},  ECNet\cite{ExcNet}, ZDCE\cite{zeroDCE}]; [ZD++\cite{zeroDCEPP}, RUAS\cite{RUAS}, SCI\cite{SCI}]; [PNet\cite{PSENet}, GDP\cite{GDP}, RSFNet(Ours, green box)]. Our results preserve the naturalness of the original scene without over/under exposing intensity or color saturation, which is also quantitatively supported by our overall better NIQE/LOE scores in  \cref{tab:qual_quant,fig:radar}.}
    \label{fig:lleComp_supp}
\end{figure*}

\begin{figure*}[t]
    \centering
    \addtolength{\tabcolsep}{-0.4em}
    \begin{tabular}{ccccc}
    \centering
    \includegraphics[width=0.25\linewidth, height=3cm]{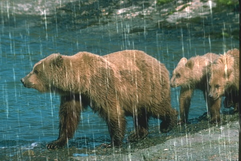} &
    \includegraphics[width=0.25\linewidth, height=3cm]{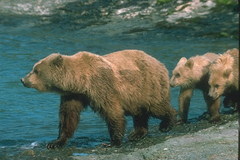} &
    \qquad &
    \includegraphics[width=0.25\linewidth, height=3cm]{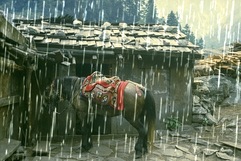} &
    \includegraphics[width=0.25\linewidth, height=3cm]{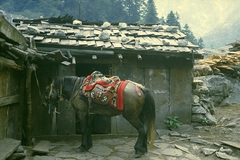} \\
    & \\
    \includegraphics[width=0.25\linewidth, height=3cm]{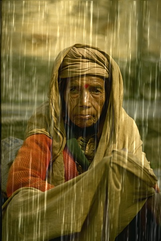} &
    \includegraphics[width=0.25\linewidth, height=3cm]{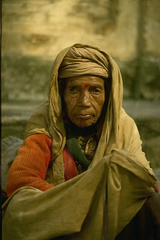} &
    \qquad &
    \includegraphics[width=0.25\linewidth, height=3cm]{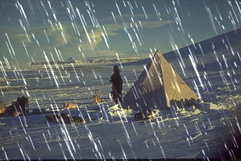} &
    \includegraphics[width=0.25\linewidth, height=3cm]{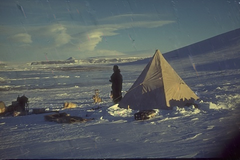} \\
    & \\
    \includegraphics[width=0.25\linewidth, height=3cm]{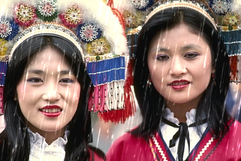} &
    \includegraphics[width=0.25\linewidth, height=3cm]{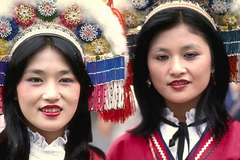} &
    \qquad &
    \includegraphics[width=0.25\linewidth, height=3cm]{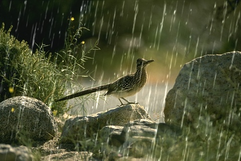} &
    \includegraphics[width=0.25\linewidth, height=3cm]{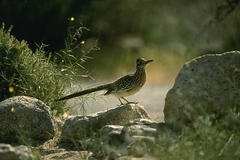} \\
    & \\
    \includegraphics[width=0.25\linewidth, height=3cm]{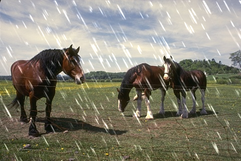} &
    \includegraphics[width=0.25\linewidth, height=3cm]{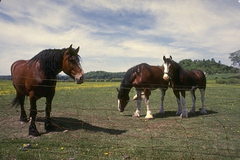} &
    \qquad &
    \includegraphics[width=0.25\linewidth, height=3cm]{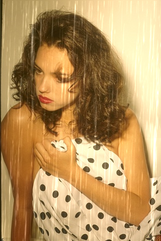} &
    \includegraphics[width=0.25\linewidth, height=3cm]{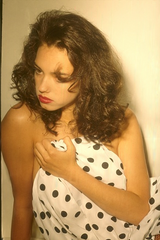} \\
    & \\
    \includegraphics[width=0.25\linewidth, height=3cm]{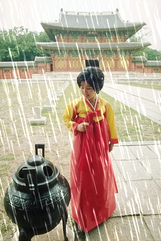} &
    \includegraphics[width=0.25\linewidth, height=3cm]{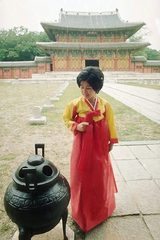} &
    \qquad &
    \includegraphics[width=0.25\linewidth, height=3cm]{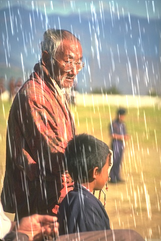} &
    \includegraphics[width=0.25\linewidth, height=3cm]{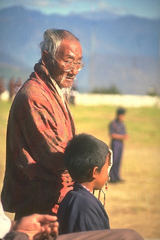} \\
    & \\
    \includegraphics[width=0.25\linewidth, height=3cm]{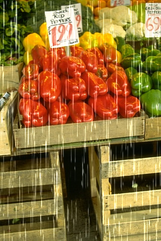} &
    \includegraphics[width=0.25\linewidth, height=3cm]{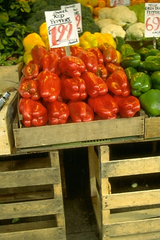} &
    \qquad &
    \includegraphics[width=0.25\linewidth, height=3cm]{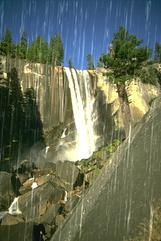} &
    \includegraphics[width=0.25\linewidth, height=3cm]{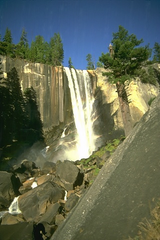} \\
    \end{tabular}
    \caption{\textbf{Our Deraining Results}: Additional results (\cref{fig:apps} extension) for the deraining application on the Rain100L dataset \cite{derainDataset}.}
    \label{fig:derain_supp}
\end{figure*}

\begin{figure*}[t]
    \centering
    \addtolength{\tabcolsep}{-0.4em}
    \begin{tabular}{ccccc}
    \centering
    \includegraphics[width=0.25\linewidth, height=3cm]{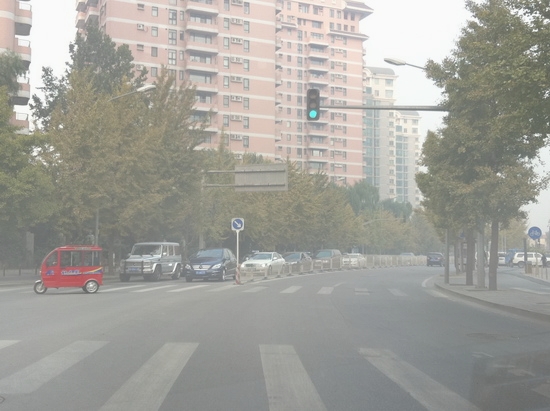} &
    \includegraphics[width=0.25\linewidth, height=3cm]{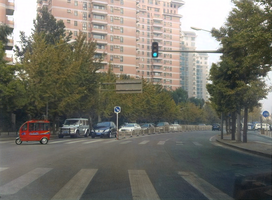} &
    \qquad &
    \includegraphics[width=0.25\linewidth, height=3cm]{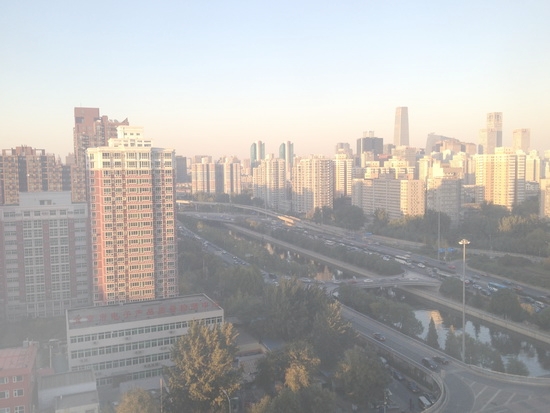} &
    \includegraphics[width=0.25\linewidth, height=3cm]{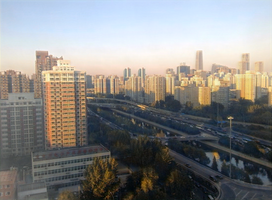} \\
    \\
    \includegraphics[width=0.25\linewidth, height=3cm]{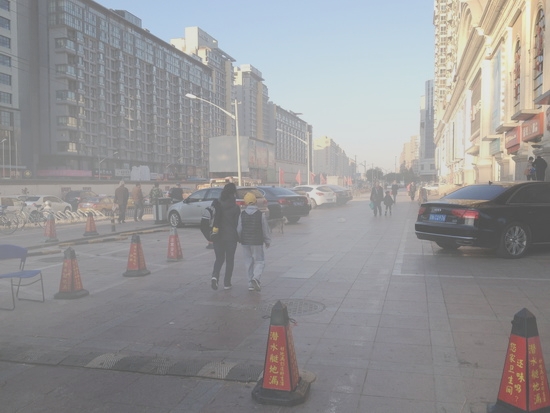} &
    \includegraphics[width=0.25\linewidth, height=3cm]{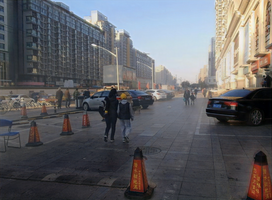} &
    \qquad &
    \includegraphics[width=0.25\linewidth, height=3cm]{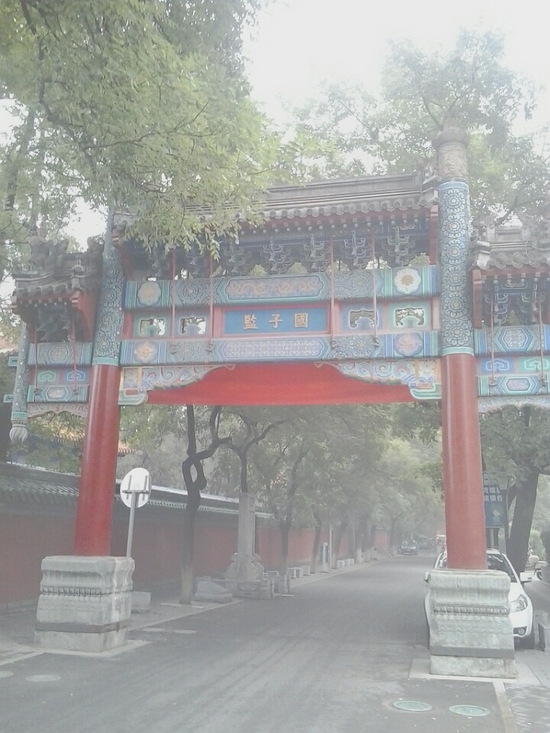} &
    \includegraphics[width=0.25\linewidth, height=3cm]{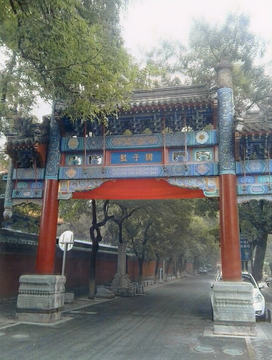} \\
    \\
    \includegraphics[width=0.25\linewidth, height=3cm]{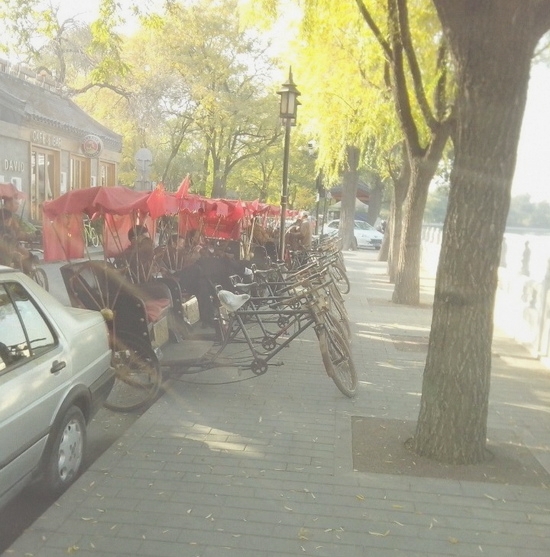} &
    \includegraphics[width=0.25\linewidth, height=3cm]{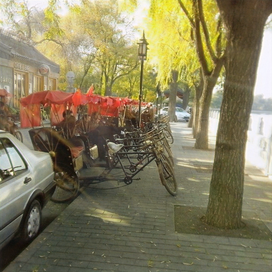} &
    \qquad &
    \includegraphics[width=0.25\linewidth, height=3cm]{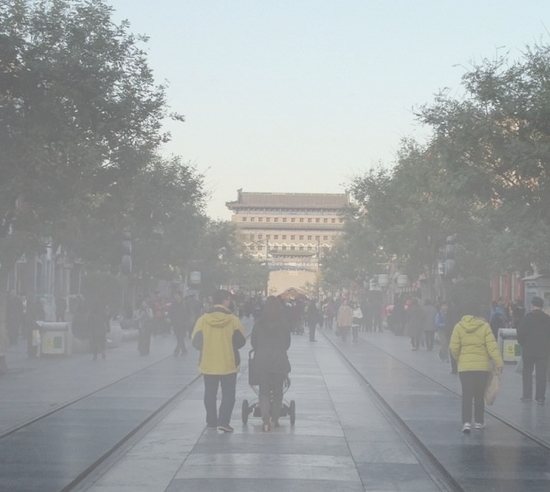} &
    \includegraphics[width=0.25\linewidth, height=3cm]{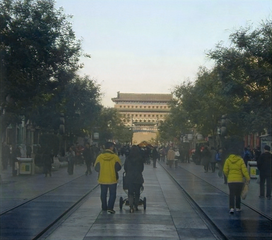} \\
    \\
    \includegraphics[width=0.25\linewidth, height=3cm]{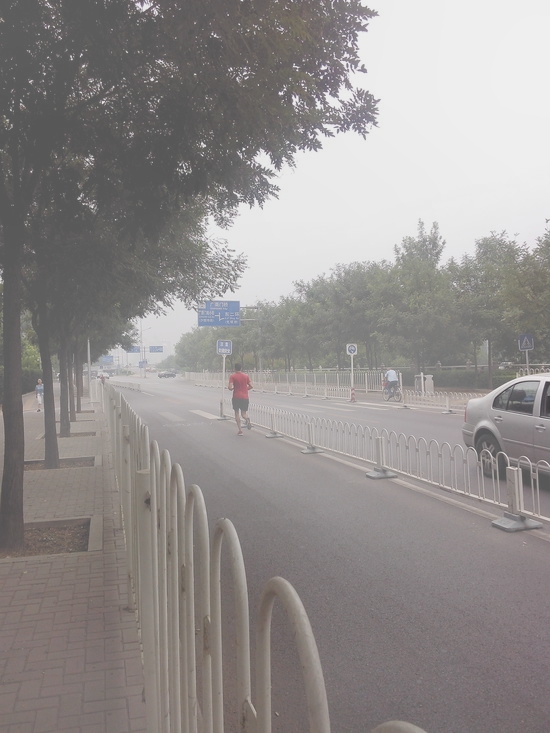} &
    \includegraphics[width=0.25\linewidth, height=3cm]{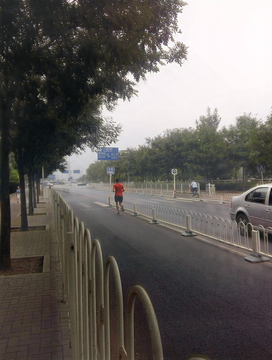} &
    \qquad &
    \includegraphics[width=0.25\linewidth, height=3cm]{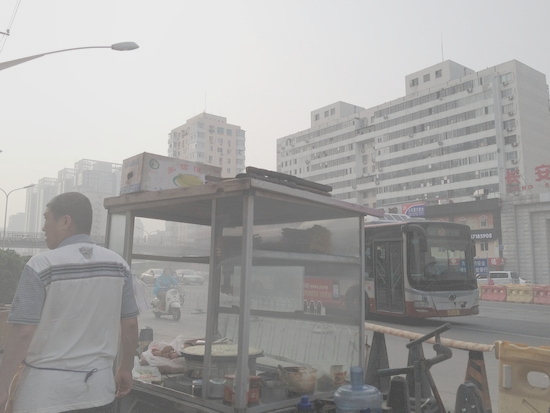} &
    \includegraphics[width=0.25\linewidth, height=3cm]{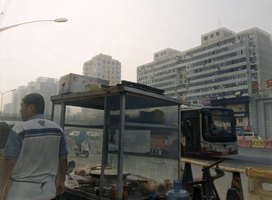} \\
    \\
    \includegraphics[width=0.25\linewidth, height=3cm]{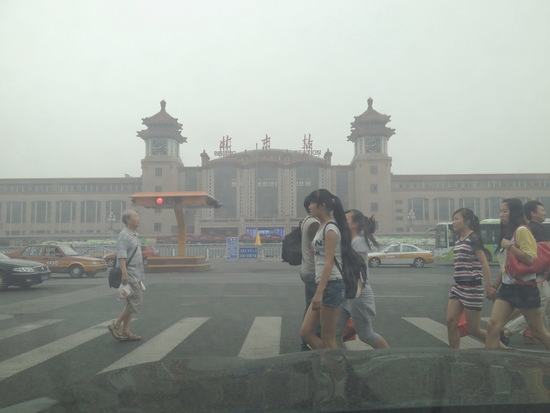} &
    \includegraphics[width=0.25\linewidth, height=3cm]{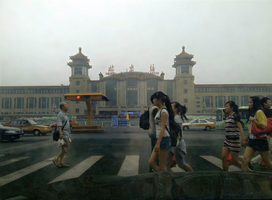} &
    \qquad &
    \includegraphics[width=0.25\linewidth, height=3cm]{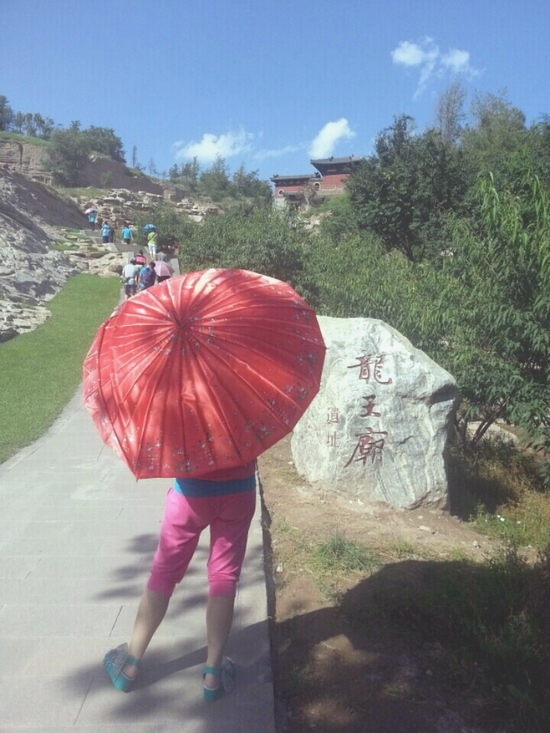} &
    \includegraphics[width=0.25\linewidth, height=3cm]{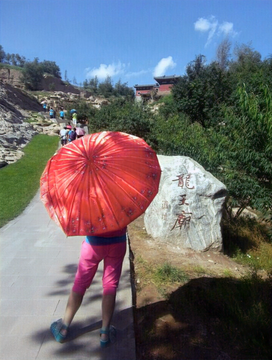} \\
    \\
    \includegraphics[width=0.25\linewidth, height=3cm]{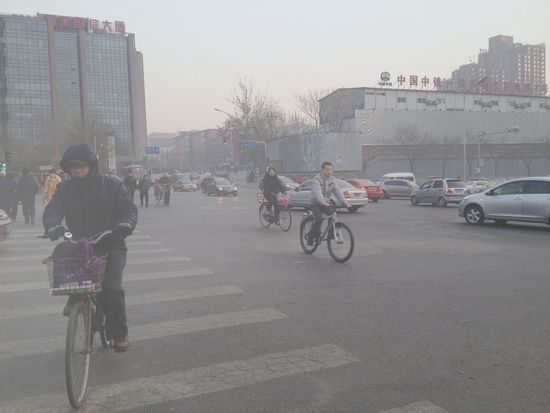} &
    \includegraphics[width=0.25\linewidth, height=3cm]{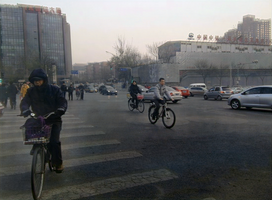} &
    \qquad &
    \includegraphics[width=0.25\linewidth, height=3cm]{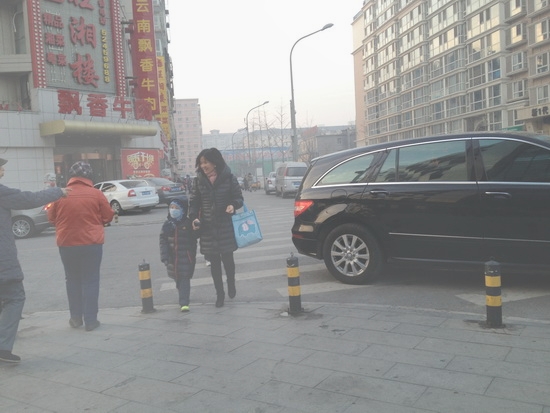} &
    \includegraphics[width=0.25\linewidth, height=3cm]{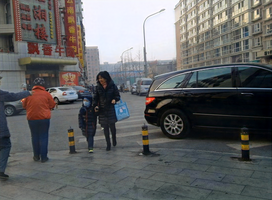} \\
    \end{tabular}
    \caption{\textbf{Our Dehazing Results}: Additional results (\cref{fig:apps} extension) for the dehazing application on the RESIDE dataset \cite{dehazeDataset}.}
    \label{fig:dehaze_supp}
\end{figure*}

\begin{figure*}[t]
    \centering
    \addtolength{\tabcolsep}{-0.4em}
    \begin{tabular}{ccccc}
    \centering
        \includegraphics[width=0.25\linewidth, height=3cm]{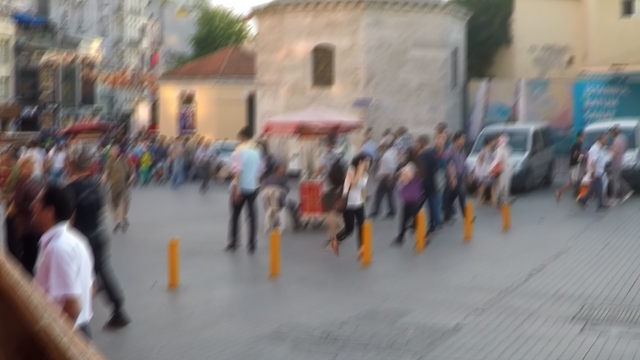} &
        \includegraphics[width=0.25\linewidth, height=3cm]{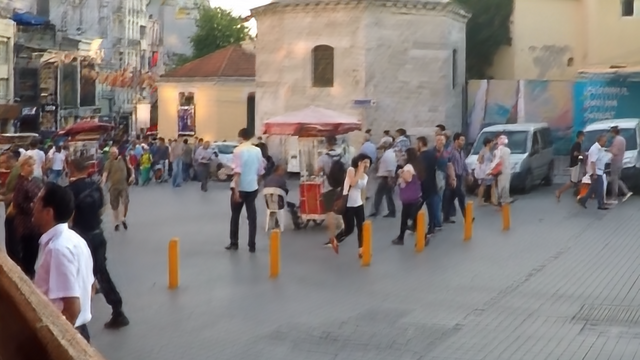} &
        \qquad &
        \includegraphics[width=0.25\linewidth, height=3cm]{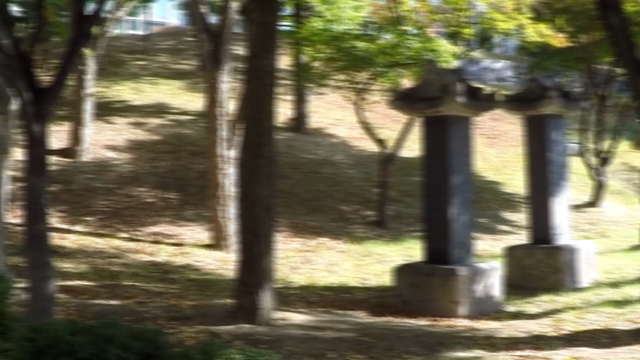} &
        \includegraphics[width=0.25\linewidth, height=3cm]{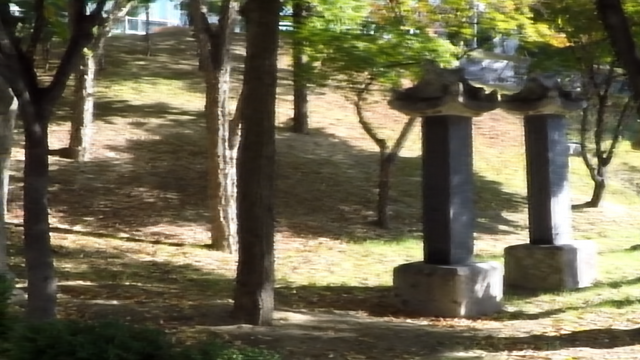} \\
        \\
        \includegraphics[width=0.25\linewidth, height=3cm]{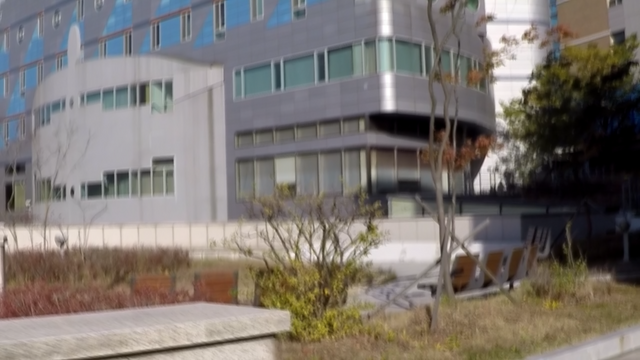} &
        \includegraphics[width=0.25\linewidth, height=3cm]{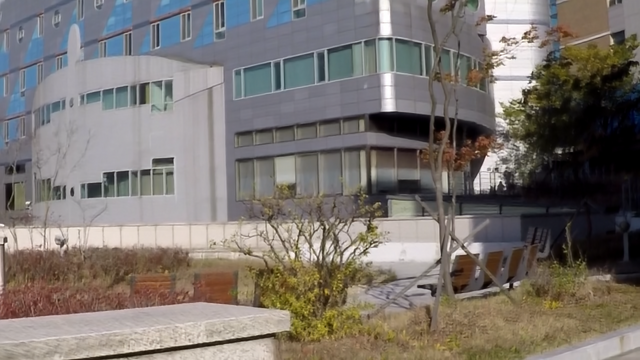} &
        \qquad &
        \includegraphics[width=0.25\linewidth, height=3cm]{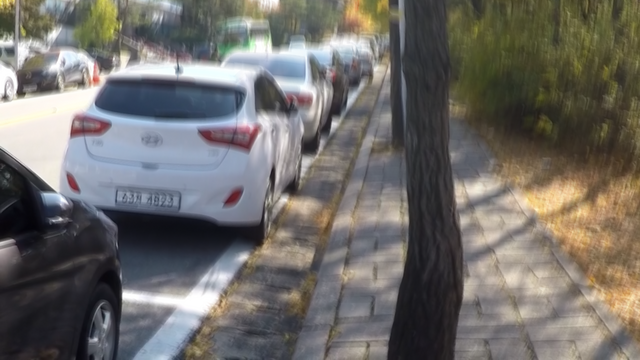} &
        \includegraphics[width=0.25\linewidth, height=3cm]{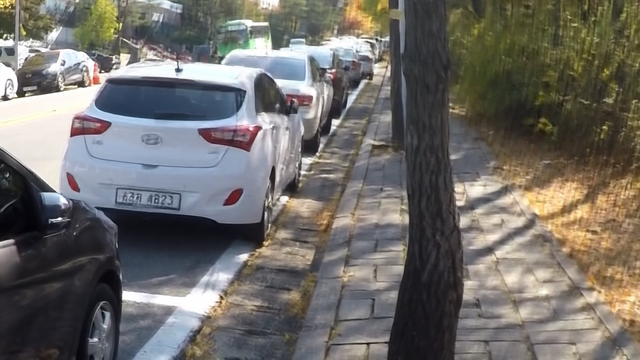} \\
        \\
        \includegraphics[width=0.25\linewidth, height=3cm]{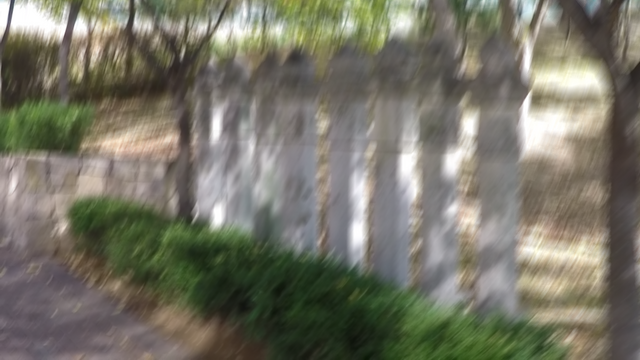} &
        \includegraphics[width=0.25\linewidth, height=3cm]{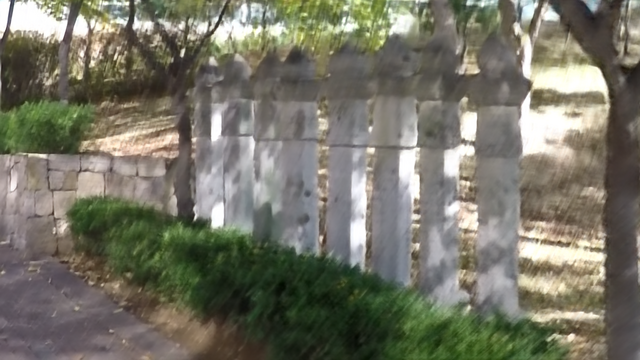} &
        \qquad &
        \includegraphics[width=0.25\linewidth, height=3cm]{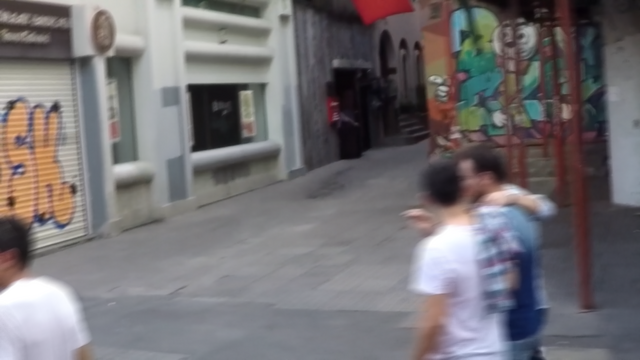} &
        \includegraphics[width=0.25\linewidth, height=3cm]{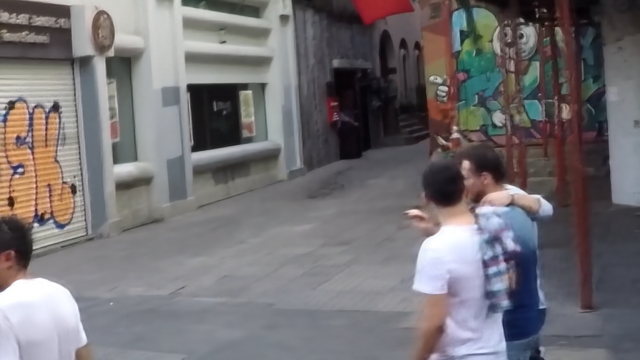} \\
        \\
        \includegraphics[width=0.25\linewidth, height=3cm]{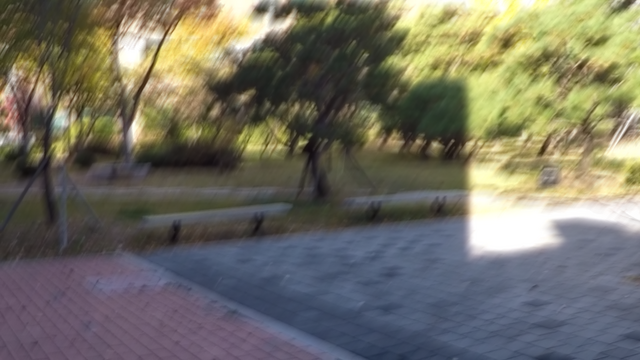} &
        \includegraphics[width=0.25\linewidth, height=3cm]{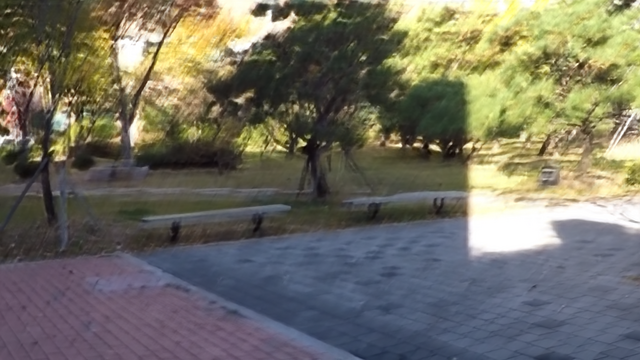} &
        \qquad &
        \includegraphics[width=0.25\linewidth, height=3cm]{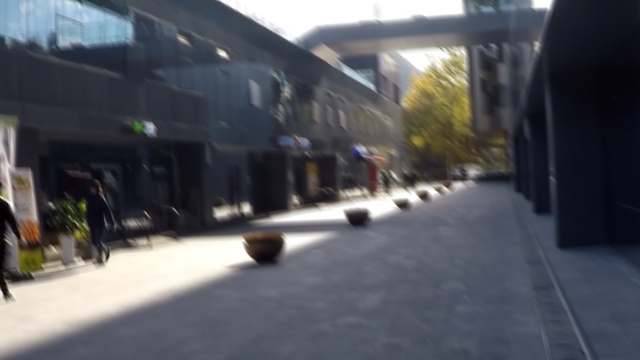} &
        \includegraphics[width=0.25\linewidth, height=3cm]{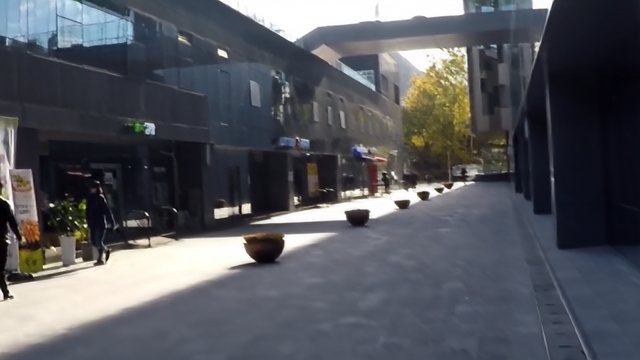} \\
        \\
        \includegraphics[width=0.25\linewidth, height=3cm]{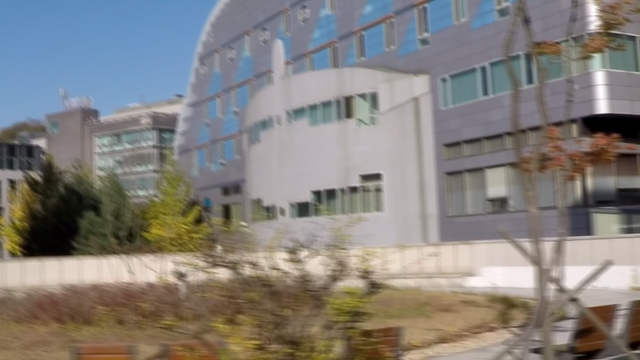} &
        \includegraphics[width=0.25\linewidth, height=3cm]{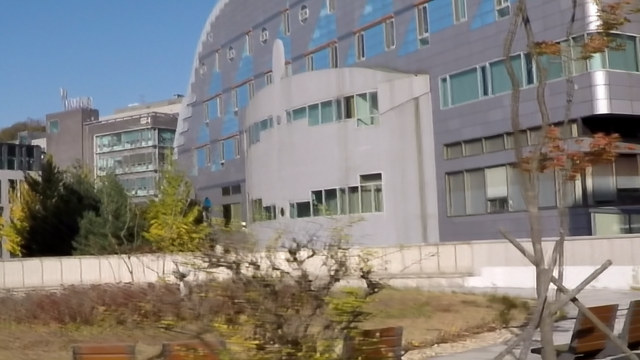} &
        \qquad &
        \includegraphics[width=0.25\linewidth, height=3cm]{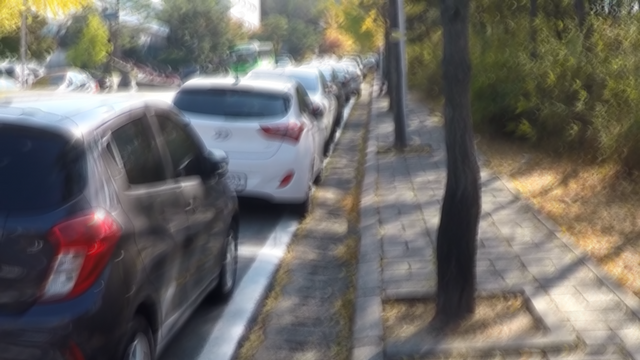} &
        \includegraphics[width=0.25\linewidth, height=3cm]{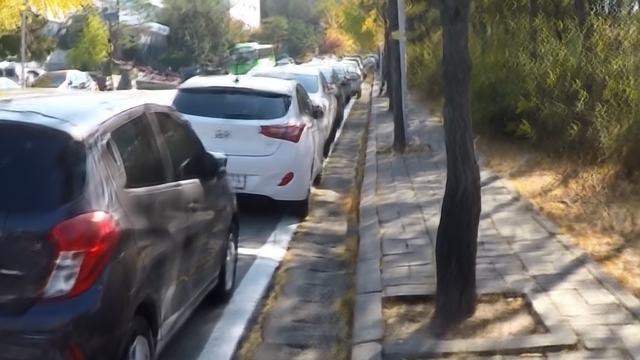} \\
        \\
        \includegraphics[width=0.25\linewidth, height=3cm]{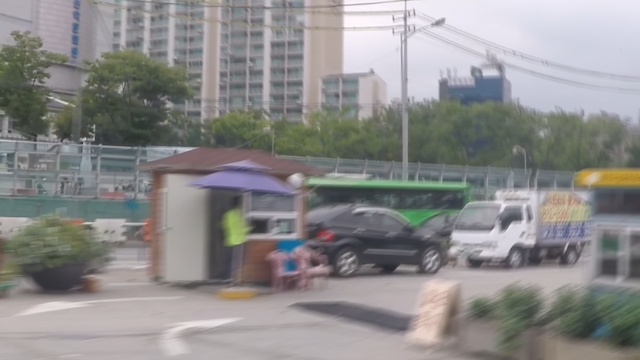} &
        \includegraphics[width=0.25\linewidth, height=3cm]{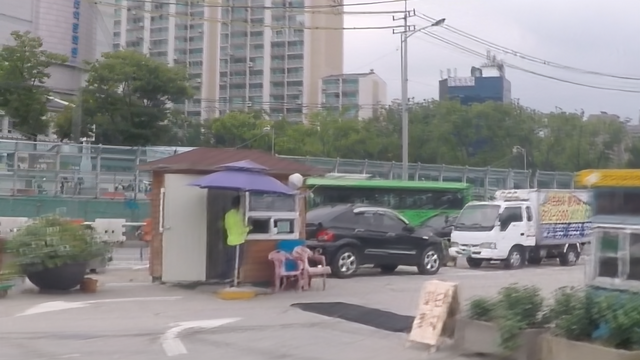} &
        \qquad &
        \includegraphics[width=0.25\linewidth, height=3cm]{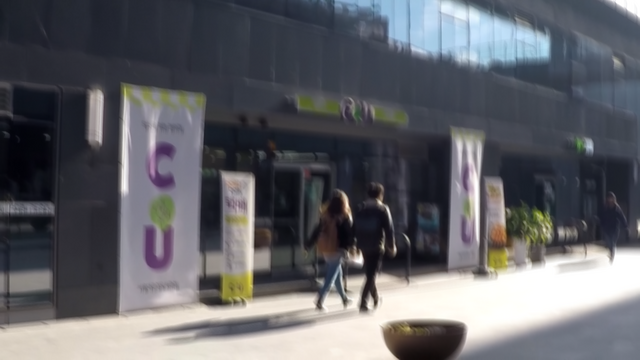} &
        \includegraphics[width=0.25\linewidth, height=3cm]{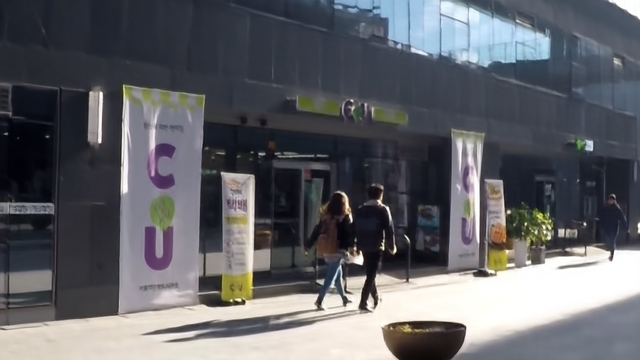} \\
        \\
    \end{tabular}
    \caption{\textbf{Our Deblurring Results}: Additional results (\cref{fig:apps} extension) for the deblurring application on the GoPro dataset \cite{deblurDataset}.}
    \label{fig:deblur_supp}
\end{figure*}

\begin{figure*}[b]
    \centering
    \begin{tabular}{c c}
         \includegraphics[width=0.5\linewidth, height=5.6cm]{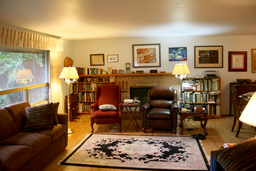} &
         \includegraphics[width=0.5\linewidth, height=5.6cm]{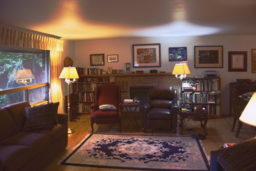} \\
         \vspace{1\baselineskip} \\
          \includegraphics[width=0.5\linewidth, height=5.6cm]{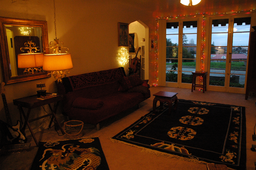} &
         \includegraphics[width=0.5\linewidth, height=5.6cm]{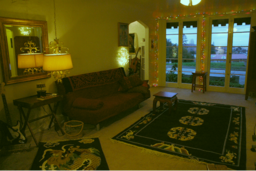} \\
         \vspace{1\baselineskip} \\
          \includegraphics[width=0.5\linewidth, height=5.6cm]{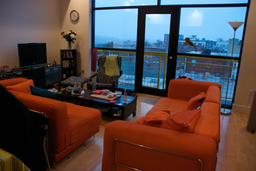} &
         \includegraphics[width=0.5\linewidth, height=5.6cm]{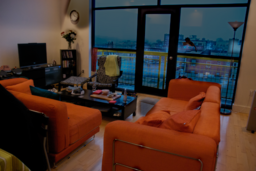} \\
    \end{tabular}
    \caption{\textbf{User-controlled Edits:} Here we show high resolution version of our results in \cref{fig:userApps}. For three scene from top to bottom we show modification of illumination specularity, indoor lighting color and outdoor lighting intensity respectively. All edits were carried out in GIMP \cite{GIMP} using our factors as layers and only global layer operations like curve adjustments, blurring, layer blending \etc were used without any local selection or modifications. Notice how our factors seamlessly merge to render such edits preserving the naturalness of the original image and without any additional artifacts. Note that these are only three representative applications and several other edits are possible with appropriate masking, color adjustments and even cross image layers harmonization.}
    \label{fig:userApps_supp}
\end{figure*}
\clearpage


\begin{table*}[t]
\centering
\resizebox{\textwidth}{!}{%
\footnotesize
\begin{tabular}{L{15mm} || C{10mm}C{10mm}C{10mm} | C{10mm}C{10mm}C{10mm}C{10mm}C{10mm}C{10mm}C{10mm} | R{12mm}}
\toprule
\toprule
\rowcolor{lightgray!40} \multicolumn{1}{c||}{} & \multicolumn{3}{c|}{} & \multicolumn{8}{c}{} \\
\rowcolor{lightgray!40} \multicolumn{1}{c||}{Paradigm} & \multicolumn{3}{c|}{\textbf{Traditional Model Based}} & \multicolumn{8}{c}{\textbf{Zero-reference}} \\
\rowcolor{lightgray!40} \multicolumn{1}{c||}{} & \multicolumn{3}{c|}{} & \multicolumn{8}{c}{} \\
\midrule
Method & LIME \cite{LIME} & DUAL \cite{DUAL} & SDD \;\cite{SDD} & ECNet \cite{ExcNet} & ZDCE \cite{zeroDCE} & ZD++ \cite{zeroDCEPP} & RUAS \cite{RUAS} & SCI \quad \cite{SCI} & PNet \; \cite{PSENet} & GDP \;\cite{GDP} & \textbf{RSFNet} \;(Ours)\\
\text{\#Params} & - & - & - & $16.5$M & $79.42$K & $10.56$K & $3.43$K & $\textbf{0.26}$K & $15.25$K & $552$K & $\underline{2.11}$K \\
\midrule
\multicolumn{12}{c}{}\\
\multicolumn{12}{c}{\textbf{Lolv1} \cite{retinexNet2018} \qquad (dataset split: 485/15, mean$\approx 0.05$, resolution: $400 \times 600$)}\\
\multicolumn{12}{c}{}\\
\midrule
$\text{PSNR}_y$ $\uparrow$  & $16.20$ & $15.97$ & $15.14$ & $18.01$ & $16.76$ & $16.38$ & $18.45$ & $16.45$ & $\underline{19.85}$ & $17.68$ & $\textbf{22.17}$\\
$\text{SSIM}_y$ $\uparrow$  & $0.695$ & $0.692$ & $0.754$ & $0.644$ & $0.734$ & $0.645$ & $\underline{0.766}$ & $0.709$ & $0.718$ & $0.678$ &  $\textbf{0.860}$\\
$\text{PSNR}_c$ $\uparrow$  & $14.22$ & $14.02$ & $13.34$ & $15.81$ & $14.86$ & $14.74$ & $16.40$ & $14.78$ & $\underline{17.50}$ & $15.80$ &  $\textbf{19.39}$\\
$\text{SSIM}_c$ $\uparrow$  & $0.521$ & $0.519$ & $\underline{0.634}$ & $0.469$ & $0.562$ & $0.496$ & $0.503$ & $0.525$ & $0.550$ & $0.539$ &  $\textbf{0.755}$\\
$\text{NIQE}$ $\downarrow$  & $8.583$ & $8.611$ & $\underline{3.706}$ & $8.844$ & $8.223$ & $8.195$ & $5.927$ & $8.374$ & $8.629$ & $6.437$ &  $\textbf{3.129}$\\
$\text{LPIPS}$$\downarrow$  & $0.344$ & $0.346$ & $\underline{0.278}$ & $0.358$ & $0.331$ & $0.346$ & $0.303$ & $0.327$ & $0.340$ & $0.375$ &  $\textbf{0.265}$\\
\midrule
\multicolumn{12}{c}{}\\
\multicolumn{12}{c}{\textbf{Lolv2-real} \cite{LOLv2} \qquad (dataset split: 689/100, mean$\approx 0.05$, resolution: $400 \times 600$)} \\
\multicolumn{12}{c}{}\\
\midrule
$\text{PSNR}_y$ $\uparrow$  & $19.31$ & $19.10$ & $18.47$ & $18.86$ & $\underline{20.31}$ & $19.36$ & $17.49$ & $19.37$ & $20.08$ & $15.83$ &  $\textbf{21.46}$\\
$\text{SSIM}_y$ $\uparrow$  & $0.705$ & $0.704$ & $\underline{0.792}$ & $0.613$ & $0.745$ & $0.585$ & $0.742$ & $0.722$ & $0.691$ & $0.627$ &  $\textbf{0.836}$\\
$\text{PSNR}_c$ $\uparrow$  & $17.14$ & $16.95$ & $16.64$ & $16.27$ & $\underline{18.06}$ & $17.36$ & $15.33$ & $17.30$ & $17.63$ & $14.05$ &  $\textbf{19.27}$\\
$\text{SSIM}_c$ $\uparrow$  & $0.537$ & $0.535$ & $\underline{0.678}$ & $0.459$ & $0.580$ & $0.442$ & $0.493$ & $0.540$ & $0.539$ & $0.502$ &  $\textbf{0.738}$\\
$\text{NIQE}$ $\downarrow$  & $9.076$ & $9.083$ & $\underline{4.191}$ & $9.475$ & $\underline{4.191}$ & $8.709$ & $6.172$ & $8.739$ & $9.152$ & $6.867$ &  $\textbf{3.769}$\\
$\text{LPIPS}$$\downarrow$  & $0.322$ & $0.324$ & $\textbf{0.280}$ & $0.360$ & $0.310$ & $0.340$ & $0.325$ & $\underline{0.294}$ & $0.340$ & $0.390$ &  $\textbf{0.280}$\\
\midrule
\multicolumn{12}{c}{}\\
\multicolumn{12}{c}{\textbf{Lolv2-synthetic} \cite{LOLv2} (dataset split: 900/100, mean$\approx 0.2$, resolution: $384 \times 384$)} \\
\multicolumn{12}{c}{}\\
\midrule
$\text{PSNR}_y$ $\uparrow$  & $19.16$ & $17.16$ & $17.93$ & $18.21$ & $19.65$ & $\underline{19.81}$ & $14.91$ & $17.09$ & $18.29$ & $13.26$ & $\textbf{19.82}$ \\
$\text{SSIM}_y$ $\uparrow$  & $0.843$ & $0.812$ & $0.787$ & $0.842$ & $\underline{0.884}$ & $0.882$ & $0.720$ & $0.825$ & $0.849$ & $0.602$ & $\textbf{0.893}$ \\
$\text{PSNR}_c$ $\uparrow$  & $\underline{17.63}$ & $15.61$ & $16.47$ & $16.75$ & $\textbf{17.76}$ & $17.58$ & $13.40$ & $15.43$ & $16.62$ & $11.97$ & $17.13$ \\
$\text{SSIM}_c$ $\uparrow$  & $0.787$ & $0.742$ & $0.725$ & $0.769$ & $\underline{0.814}$ & $0.811$ & $0.640$ & $0.744$ & $0.773$ & $0.481$ & $\textbf{0.816}$ \\
$\text{NIQE}$ $\downarrow$  & $4.685$ & $4.741$ & $4.335$ & $4.311$ & $4.357$ & $\textbf{4.257}$ & $5.092$ & $4.652$ & $\underline{4.308}$ & $-$ & $4.404$ \\
$\text{LPIPS}$$\downarrow$  & $0.174$ & $0.194$ & $0.235$ & $0.178$ & $\textbf{0.142}$ & $\underline{0.157}$ & $0.365$ & $0.203$ & $0.160$ & $0.311$ & $\underline{0.157}$ \\
\midrule
\multicolumn{12}{c}{}\\
\multicolumn{12}{c}{\textbf{VE-Lol} \cite{VELOL} (dataset split: 1400/100, mean$\approx 0.07$, resolution: $400 \times 600$)} \\
\multicolumn{12}{c}{}\\
\midrule
$\text{PSNR}_y$ $\uparrow$  & $19.31$ & $19.10$ & $18.47$ & $18.72$ & $20.31$ & $19.36$ & $17.49$ & $19.37$ & $\underline{20.39}$ & $16.29$ & $\textbf{21.18}$ \\
$\text{SSIM}_y$ $\uparrow$  & $0.705$ & $0.704$ & $\underline{0.792}$ & $0.610$ & $0.745$ & $0.585$ & $0.742$ & $0.722$ & $0.715$ & $0.628$ & $\textbf{0.817}$ \\
$\text{PSNR}_c$ $\uparrow$  & $17.14$ & $16.95$ & $16.64$ & $16.15$ & $\textbf{18.06}$ & $17.36$ & $15.33$ & $17.30$ & $\underline{17.64}$ & $14.42$ & $\textbf{18.06}$ \\
$\text{SSIM}_c$ $\uparrow$  & $0.537$ & $0.535$ & $\underline{0.678}$ & $0.457$ & $0.580$ & $0.442$ & $0.493$ & $0.540$ & $0.557$ & $0.498$ & $\textbf{0.714}$ \\
$\text{NIQE}$ $\downarrow$  & $9.076$ & $9.083$ & $\underline{4.191}$ & $9.482$ & $8.767$ & $8.709$ & $6.172$ & $8.739$ & $9.073$ & $7.027$ & $\textbf{3.782}$ \\
$\text{LPIPS}$$\downarrow$  & $0.322$ & $0.324$ & $\textbf{0.275}$ & $0.418$ & $\underline{0.310}$ & $0.340$ & $0.390$ & $0.355$ & $0.368$ & $0.444$ & $0.397$ \\
\midrule
\rowcolor{lightgray!40} \multicolumn{12}{c}{}\\
\rowcolor{lightgray!40} \multicolumn{12}{c}{\underline{\textbf{Mean Scores}} \qquad (\textbf{Lolv1} \cite{retinexNet2018}, \textbf{Lolv2-real} \cite{LOLv2}, \textbf{Lolv2-syn} \cite{LOLv2} and \textbf{VE-Lol} \cite{VELOL}) } \\
\rowcolor{lightgray!40} \multicolumn{12}{c}{}\\
\midrule
$\text{PSNR}_y$ $\uparrow$ & $18.50$  &  $17.83$  &  $17.50$  &  $18.45$  &  $19.26$  &  $18.73$  &  $17.09$  &  $18.07$  &  $\underline{19.65}$  &  $15.88$  &  $\textbf{21.16}$ \\
$\text{SSIM}_y$ $\uparrow$ & $0.737$  &  $0.728$  &  $\underline{0.781}$  &  $0.677$  &  $0.777$  &  $0.674$  &  $0.743$  &  $0.745$  &  $0.743$  &  $0.634$  &  $\textbf{0.854}$ \\
$\text{PSNR}_c$ $\uparrow$ & $16.53$  &  $15.88$  &  $15.77$  &  $16.25$  &  $17.19$  &  $16.76$  &  $15.12$  &  $16.20$  &  $\underline{17.35}$  &  $14.15$  &  $\textbf{18.45}$ \\
$\text{SSIM}_c$ $\uparrow$ & $0.596$  &  $0.583$  &  $\underline{0.679}$  &  $0.538$  &  $0.634$  &  $0.548$  &  $0.532$  &  $0.587$  &  $0.605$  &  $0.504$  &  $\textbf{0.758}$ \\
$\text{NIQE}$ $\downarrow$ & $7.855$  &  $7.880$  &  $\underline{4.106}$  &  $8.028$  &  $6.385$  &  $7.468$  &  $5.841$  &  $7.626$  &  $7.791$  &  $6.826$  &  $\textbf{3.763}$ \\
$\text{LPIPS}$$\downarrow$ & $0.291$  &  $0.297$  &  $\textbf{0.266}$  &  $0.329$  &  $\underline{0.273}$  &  $0.296$  &  $0.346$  &  $0.295$  &  $0.302$  &  $0.379$  &  $0.276$ \\
\bottomrule
\bottomrule
\end{tabular}
}
\caption{\textbf{Quantitative comparison} of our method RSFNet with other \textbf{traditional and zero-reference} solutions on multiple lowlight benchmarks and six evaluation metrics. Shown here are scores for two datasets Lolv1 \cite{retinexNet2018} and Lolv2-real \cite{LOLv2} with mean value across all datasets in the last sub-table. Our scores here are same as the ones reported in last sub-table in \cref{tab:quant} in the main paper (key: $\uparrow$ higher better; $\downarrow$ lower better; \textbf{bold}: best; \underline{underline}: second best; '-': NaN error computing value).}
\label{tab:quant_supp}
\end{table*}
\begin{table*}[t]
\centering
\resizebox{\textwidth}{!}{%
\small
\begin{tabular}{L{15mm} || C{8mm}C{8mm}C{8mm}C{10mm} || C{14mm}C{14mm}C{14mm}C{14mm}C{14mm} | R{8mm}}
\toprule
\toprule
\rowcolor{lightgray!50}\multicolumn{1}{c||}{} & \multicolumn{4}{c||}{} & \multicolumn{5}{c|}{} & \multicolumn{1}{c}{} \\
\rowcolor{lightgray!50} \multicolumn{1}{l||}{Paradigm} & \multicolumn{4}{c||}{\textit{\textbf{Supervised LLE}}} & \multicolumn{5}{c|}{\textbf{Unsupervised LLE}} & \multicolumn{1}{c}{\textbf{Zero}} \\
\rowcolor{lightgray!50}\multicolumn{1}{c||}{} & \multicolumn{4}{c||}{} & \multicolumn{5}{c|}{} & \multicolumn{1}{c}{\textbf{Reference}} \\
\midrule
Method & \textit{URetinex} \cite{UretinexNet} & \textit{CUE} \cite{CUE} & \textit{SNR} \cite{SNR} & \textit{RFormer} \cite{RFormer} & EGAN \cite{jiang2021enlightengan} & HEP \cite{HEP} & PairLIE* \cite{PairLLE} & CLIP-LIT \cite{CLIP-LIT} & NeRCo* \cite{NERCO} & \textbf{RSFNet} \;(Ours)\\
\midrule
\multicolumn{11}{c}{}\\
\multicolumn{11}{c}{\textbf{Lolv1} \cite{retinexNet2018} \qquad (dataset split: 485/15, mean$\approx 0.05$, resolution: $400 \times 600$)}\\
\multicolumn{11}{c}{}\\
\midrule
$\text{PSNR}_y$ $\uparrow$  & $\textit{22.16}$ & $\textit{24.57}$ & $\textit{28.33}$ & $\textit{28.81}$ & $19.69$ & $20.82$ & $20.51$ & $14.13$ & $25.53$ & $22.15$\\
$\text{SSIM}_y$ $\uparrow$  & $\textit{0.900}$ & $\textit{0.852}$ & $\textit{0.910}$ & $\textit{0.914}$ & $0.764$ & $0.874$ & $0.840$ & $0.659$ & $0.860$ & $0.860$\\
$\text{PSNR}_c$ $\uparrow$  & $\textit{19.84}$ & $\textit{21.67}$ & $\textit{24.16}$ & $\textit{25.15}$ & $17.48$ & $20.23$ & $18.47$ & $12.39$ & $22.95$ & $19.35$\\
$\text{SSIM}_c$ $\uparrow$  & $\textit{0.824}$ & $\textit{0.769}$ & $\textit{0.840}$ & $\textit{0.843}$ & $0.652$ & $0.790$ & $0.743$ & $0.493$ & $0.784$ & $0.755$\\
$\text{NIQE}$  $\downarrow$   & $\textit{3.541}$ & $\textit{3.198}$ & $\textit{4.016}$ & $\textit{2.972}$ & $4.889$ & $3.295$ & $4.038$ & $8.797$ & $3.538$ & $3.146$\\
$\text{LPIPS}$ $\downarrow$   & $\textit{0.168}$ & $\textit{0.277}$ & $\textit{0.207}$ & $\textit{0.193}$ & $0.327$ & $0.223$ & $0.290$ & $0.359$ & $0.243$ & $0.265$\\
\midrule
\multicolumn{11}{c}{}\\
\multicolumn{11}{c}{\textbf{Lolv2-real} \cite{LOLv2} \qquad (dataset split: 689/100, mean$\approx 0.05$, resolution: $400 \times 600$)} \\
\multicolumn{11}{c}{}\\
\midrule
$\text{PSNR}_y$ $\uparrow$  & $\textit{22.97}$ & $\textit{24.48}$ & $\textit{23.20}$ & $\textit{24.80}$ & $21.27$ & $20.87$ & $-$ & $17.03$ & $-$ & $21.59$\\
$\text{SSIM}_y$ $\uparrow$  & $\textit{0.900}$ & $\textit{0.848}$ & $\textit{0.893}$ & $\textit{0.888}$ & $0.770$ & $0.860$ & $-$ & $0.696$ & $-$ & $0.843$\\
$\text{PSNR}_c$ $\uparrow$  & $\textit{21.09}$ & $\textit{22.56}$ & $\textit{21.48}$ & $\textit{22.79}$ & $18.64$ & $18.97$ & $-$ & $15.18$ & $-$ & $19.39$\\
$\text{SSIM}_c$ $\uparrow$  & $\textit{0.858}$ & $\textit{0.799}$ & $\textit{0.848}$ & $\textit{0.839}$ & $0.677$ & $0.808$ & $-$ & $0.533$ & $-$ & $0.745$\\
$\text{NIQE}$  $\downarrow$   & $\textit{4.010}$ & $\textit{3.709}$ & $\textit{4.141}$ & $\textit{3.594}$ & $5.503$ & $3.618$ & $-$ & $9.220$ & $-$ & $3.701$\\
$\text{LPIPS}$ $\downarrow$   & $\textit{0.147}$ & $\textit{0.270}$ & $\textit{0.199}$ & $\textit{0.228}$ & $0.321$ & $0.218$ & $-$ & $0.328$ & $-$ & $0.278$\\
\midrule
\multicolumn{11}{c}{}\\
\multicolumn{11}{c}{\textbf{Lolv2-synthetic} \cite{LOLv2} (dataset split: 900/100, mean$\approx 0.2$, resolution: $384 \times 384$)} \\
\multicolumn{11}{c}{}\\
\midrule
$\text{PSNR}_y$ $\uparrow$  & $\textit{20.35}$ & $\textit{18.48}$ & $\textit{25.89}$ & $\textit{27.66}$ & $18.18$ & $17.69$ & $21.13$ & $17.65$ & $18.55$ & $20.15$\\
$\text{SSIM}_y$ $\uparrow$  & $\textit{0.888}$ & $\textit{0.803}$ & $\textit{0.957}$ & $\textit{0.962}$ & $0.843$ & $0.828$ & $0.866$ & $0.840$ & $0.745$ & $0.895$\\
$\text{PSNR}_c$ $\uparrow$  & $\textit{18.25}$ & $\textit{16.49}$ & $\textit{24.14}$ & $\textit{25.67}$ & $16.57$ & $15.62$ & $19.07$ & $16.19$ & $16.07$ & $17.18$\\
$\text{SSIM}_c$ $\uparrow$  & $\textit{0.821}$ & $\textit{0.734}$ & $\textit{0.927}$ & $\textit{0.928}$ & $0.772$ & $0.752$ & $0.794$ & $0.772$ & $0.673$ & $0.817$\\
$\text{NIQE}$  $\downarrow$   & $\textit{4.338}$ & $\textit{4.165}$ & $\textit{3.969}$ & $\textit{3.939}$ & $3.831$ & $4.692$ & $4.946$ & $4.690$ & $3.735$ & $4.404$\\
$\text{LPIPS}$ $\downarrow$   & $\textit{0.195}$ & $\textit{0.283}$ & $\textit{0.065}$ & $\textit{0.076}$ & $0.188$ & $0.283$ & $0.224$ & $0.177$ & $0.378$ & $0.159$\\
\midrule
\rowcolor{lightgray!40} \multicolumn{11}{c}{}\\
\rowcolor{lightgray!40} \multicolumn{11}{c}{\underline{\textbf{Mean Scores}} \qquad (\textbf{Lolv1} \cite{retinexNet2018}, \textbf{Lolv2-real} \cite{LOLv2}, \textbf{Lolv2-syn} \cite{LOLv2}) } \\
\rowcolor{lightgray!40} \multicolumn{11}{c}{}\\
\midrule
$\text{PSNR}_y$ $\uparrow$  & $\textit{21.83}$ & $\textit{22.51}$ & $\textit{25.81}$ & $\textit{\textbf{27.09}}$ & $19.71$ & $20.46$ & $20.82$ & $16.27$ & $\textbf{22.04}$ & $\textbf{21.30}$\\
$\text{SSIM}_y$ $\uparrow$  & $\textit{0.896}$ & $\textit{0.834}$ & $\textit{0.920}$ & $\textit{\textbf{0.921}}$ & $0.792$ & $\textbf{0.854}$ & $0.853$ & $0.732$ & $0.803$ & $\textbf{0.866}$\\
$\text{PSNR}_c$ $\uparrow$  & $\textit{19.73}$ & $\textit{20.24}$ & $\textit{23.41}$ & $\textit{\textbf{24.54}}$ & $17.56$ & $18.27$ & $18.77$ & $14.59$ & $\textbf{19.51}$ & $\textbf{18.64}$\\
$\text{SSIM}_c$ $\uparrow$  & $\textit{0.834}$ & $\textit{0.767}$ & $\textit{\textbf{0.872}}$ & $\textit{0.870}$ & $0.700$ & $\textbf{0.783}$ & $0.769$ & $0.599$ & $0.729$ & $\textbf{0.772}$\\
$\text{NIQE}$  $\downarrow$   & $\textit{3.963}$ & $\textit{3.691}$ & $\textit{4.042}$ & $\textit{\textbf{3.502}}$ & $4.741$ & $3.868$ & $4.492$ & $7.569$ & $\textbf{3.637}$ & $\textbf{3.424}$\\
$\text{LPIPS}$ $\downarrow$   & $\textit{0.170}$ & $\textit{0.277}$ & $\textit{\textbf{0.157}}$ & $\textit{0.166}$ & $0.279$ & $\textbf{0.241}$ & $0.257$ & $0.288$ & $0.311$ & $\textbf{0.234}$\\
\bottomrule
\bottomrule
\end{tabular}
}
\caption{\textbf{Quantitative comparison} of our method RSFNet with five other \textbf{Unsupervised LLE} solutions \cite{jiang2021enlightengan,HEP,PairLLE,CLIP-LIT,NERCO} and four recent Supervised LLE solutions \cite{UretinexNet,SNR,CUE,RFormer} for reference. Note that the latter two categories require both low-light and well-lit images, either unpaired or paired, for supervision during training. The final average scores are presented in the last sub-table. (* For PairLIE \cite{PairLLE} and NeRCo \cite{NERCO}, training set includes Lolv2 test images, hence the results are not estimated for Lolv2 and average computed using other two scores. Even with zero-reference training requirements, our method (last column) is able to perform competitively against all unsupervised solutions. For \cite{NERCO} and \cite{HEP}, our method beats both of them separately on 4/6 and 5/6 metrics. Note that supervised solutions require significantly more supervision information during training and can not be compared directly with other categories. Here they are shown only for reference (Best score in each category here is in \textbf{bold} in the last sub-table. Our method in the last column gives the best mean results among Zero-Reference methods as shown elsewhere.).
}
\label{tab:unsupQuant_supp}
\end{table*}


}
\end{document}